\documentclass[10pt,twocolumn,letterpaper]{article}

\usepackage{cvpr}
\usepackage{times}
\usepackage{epsfig}
\usepackage{graphicx}
\usepackage{amsmath}
\usepackage{amssymb}

\usepackage{subcaption}
\usepackage[table]{xcolor}
\usepackage{array}
\usepackage{multirow}

\usepackage{enumitem}

\newcommand{\juju}[1]{{\color{orange} [J: #1]}}
\DeclareMathOperator*{\argmin}{arg\,min}

\usepackage[pagebackref=true,breaklinks=true,letterpaper=true,colorlinks,bookmarks=false]{hyperref}

\cvprfinalcopy 

\begin{document}
\title{Detecting Overfitting of Deep Generative Networks \emph{via} Latent Recovery}

\author{Ryan Webster, Julien Rabin, Lo\"{i}c Simon and Fr\'{e}d\'{e}ric Jurie\\
Normandie Univ, UNICAEN, ENSICAEN, CNRS — UMR GREYC   \\                %
{\tt\small ryan.webster@unicaen.fr}                                                       
}                                                                   
\maketitle

\begin{abstract} State of the art deep generative networks are capable of producing images with such incredible realism that they can be suspected of memorizing training images.  
It is why it is not uncommon to include visualizations of training set nearest neighbors, to suggest generated images are not simply memorized. We demonstrate this is not sufficient and motivates the need to study memorization/overfitting of deep generators with more scrutiny. This paper addresses this question by i) showing how simple losses are highly effective at reconstructing images for deep generators ii) analyzing the statistics of reconstruction errors when reconstructing training and validation images, which is the standard way to analyze overfitting in machine learning. Using this methodology, this paper shows that overfitting is not detectable in the pure GAN models proposed in the literature, in contrast with those using hybrid adversarial losses, which are amongst the most widely applied generative methods. The paper also shows that standard GAN evaluation metrics fail to capture memorization for some deep generators. Finally, the paper also shows how off-the-shelf GAN generators can be successfully applied to face inpainting and face super-resolution using the proposed reconstruction method, without hybrid adversarial losses.  

\end{abstract}

\iffalse
\begin{figure*}[!htb]
\centering
\begin{tabular}{cccc}
	\multicolumn{4}{c}{\em Registered target images $y$ at $1024 \times 1024$ resolution} \\
    \includegraphics[width=.2\linewidth]{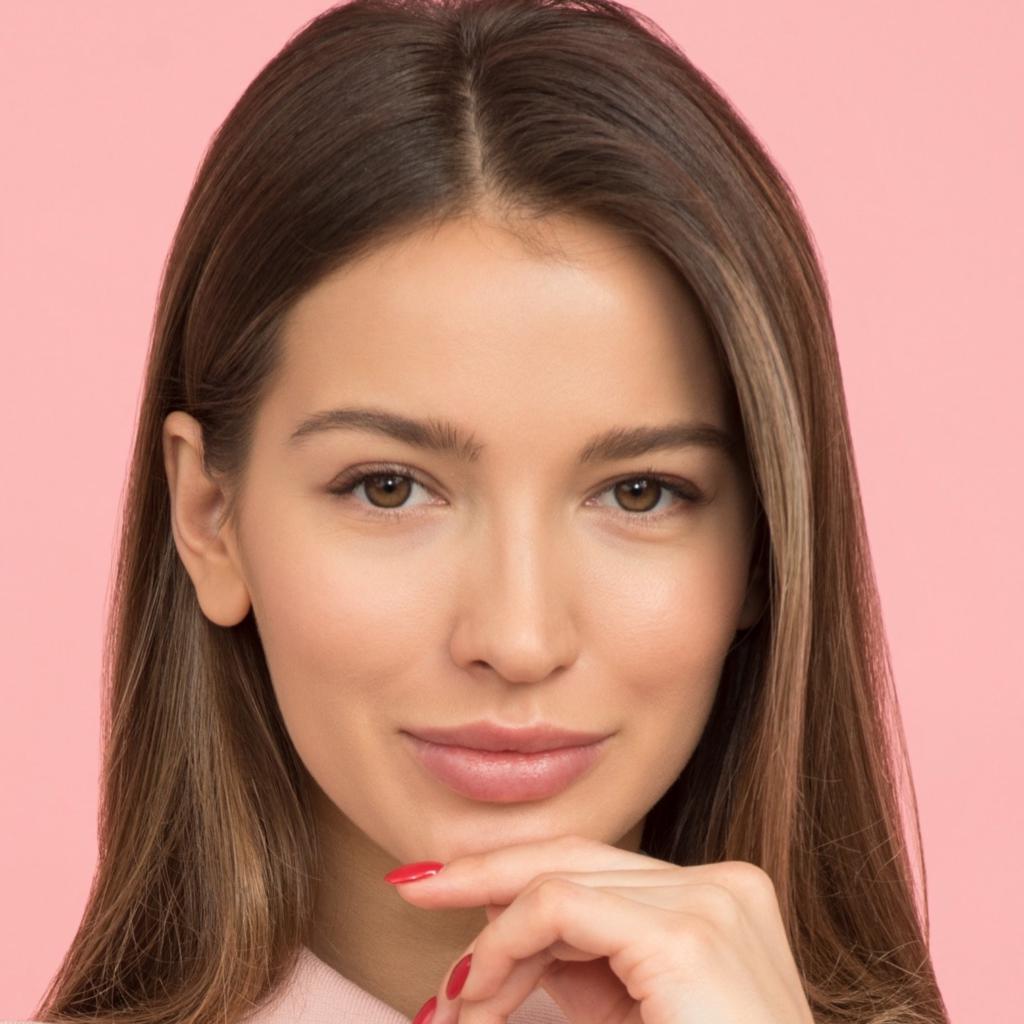} &
    \includegraphics[width=.2\linewidth]{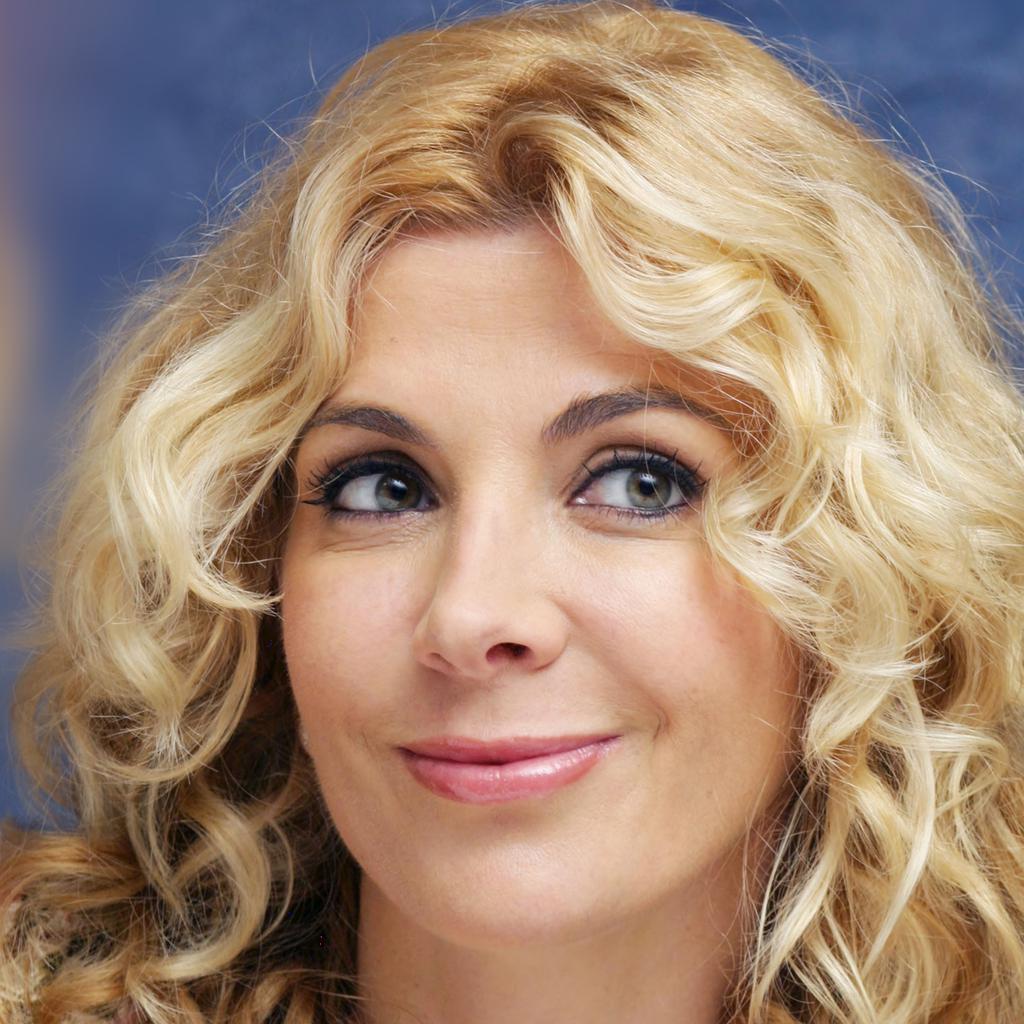} &
    \includegraphics[width=.2\linewidth]{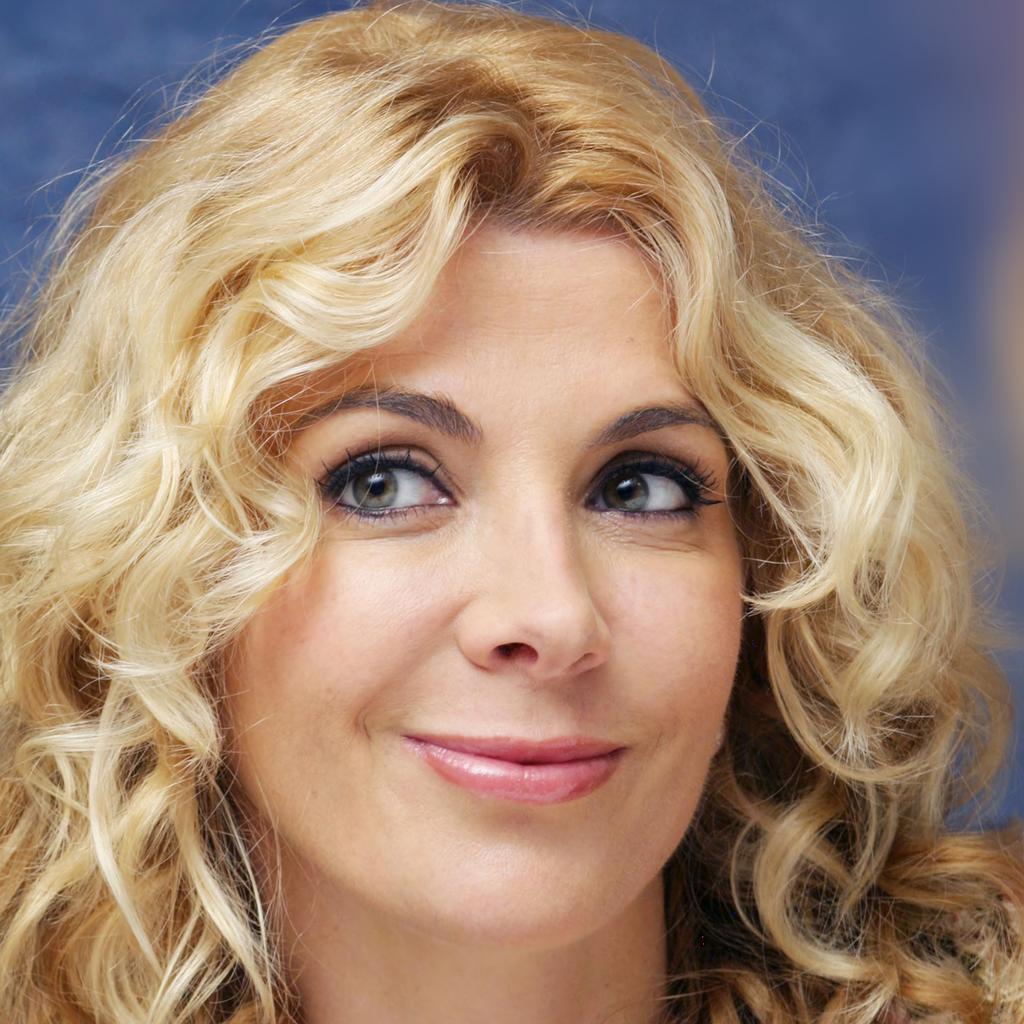} &
    \includegraphics[width=.2\linewidth]{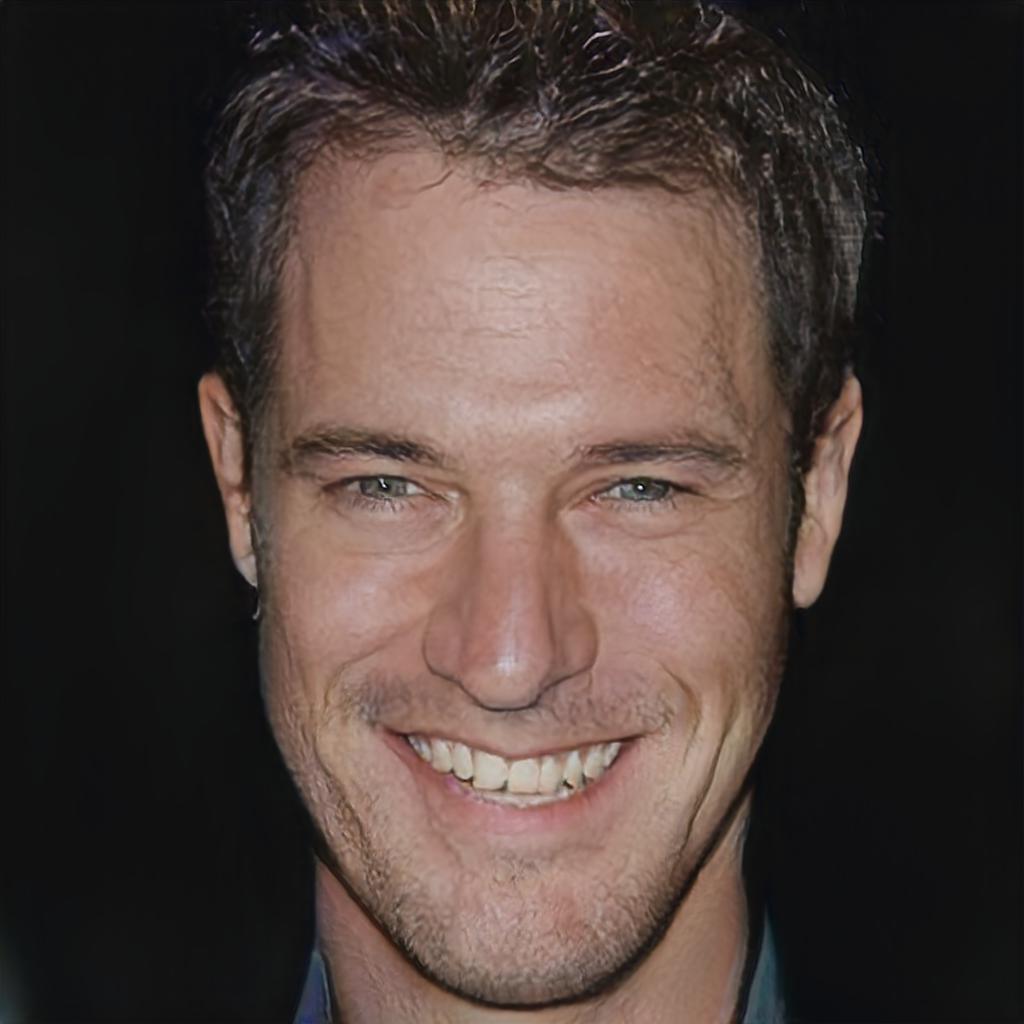} 
    \\
    $y \not\in \cal D$ &
    $y \in \cal D$&
    (Flipped) &
    $G(z)$
    \\[2mm]
    
	\multicolumn{4}{c}{\em NN recovery in CelebA-HQ} $\argmin_{y \in D} \| y - y \|$ \\
    \includegraphics[width=.2\linewidth]{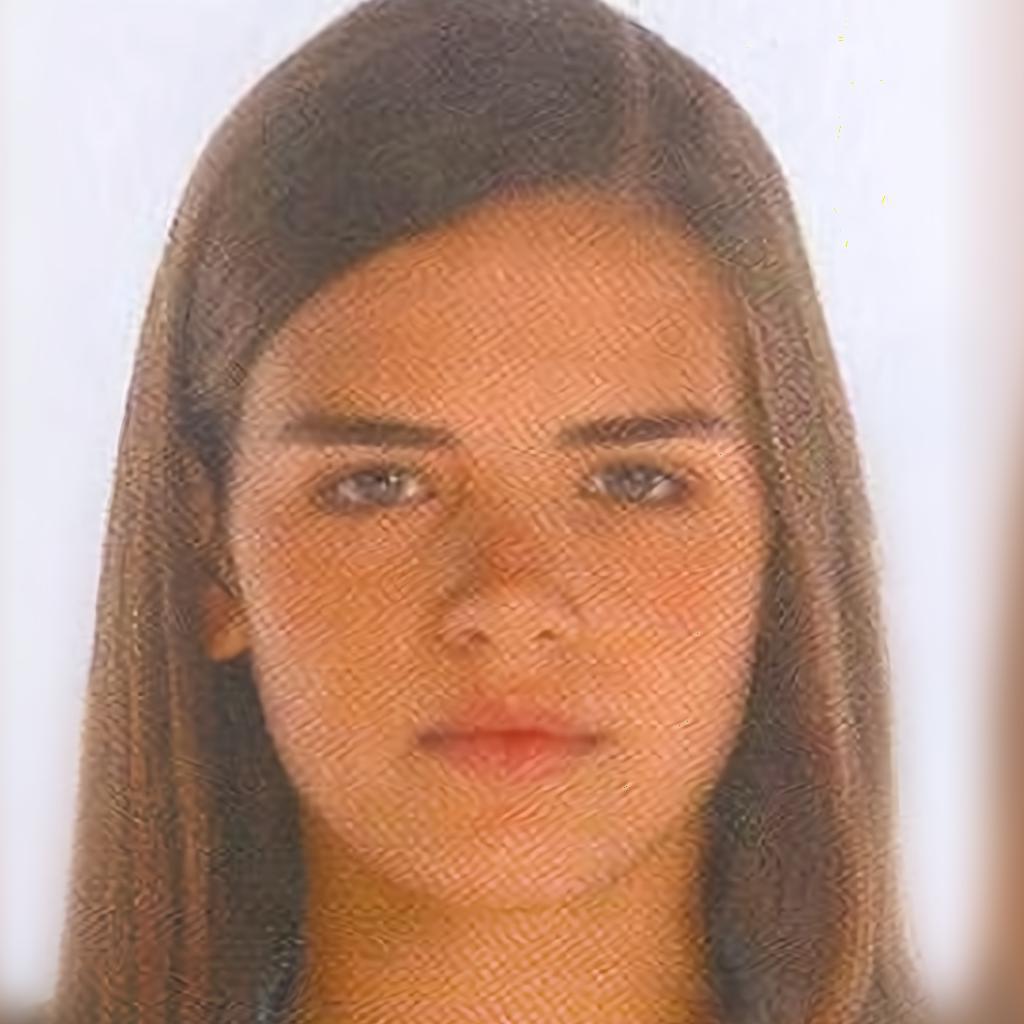} &
    \includegraphics[width=.2\linewidth]{fig_teaser/02014_target.jpg} &
    \includegraphics[width=.2\linewidth]{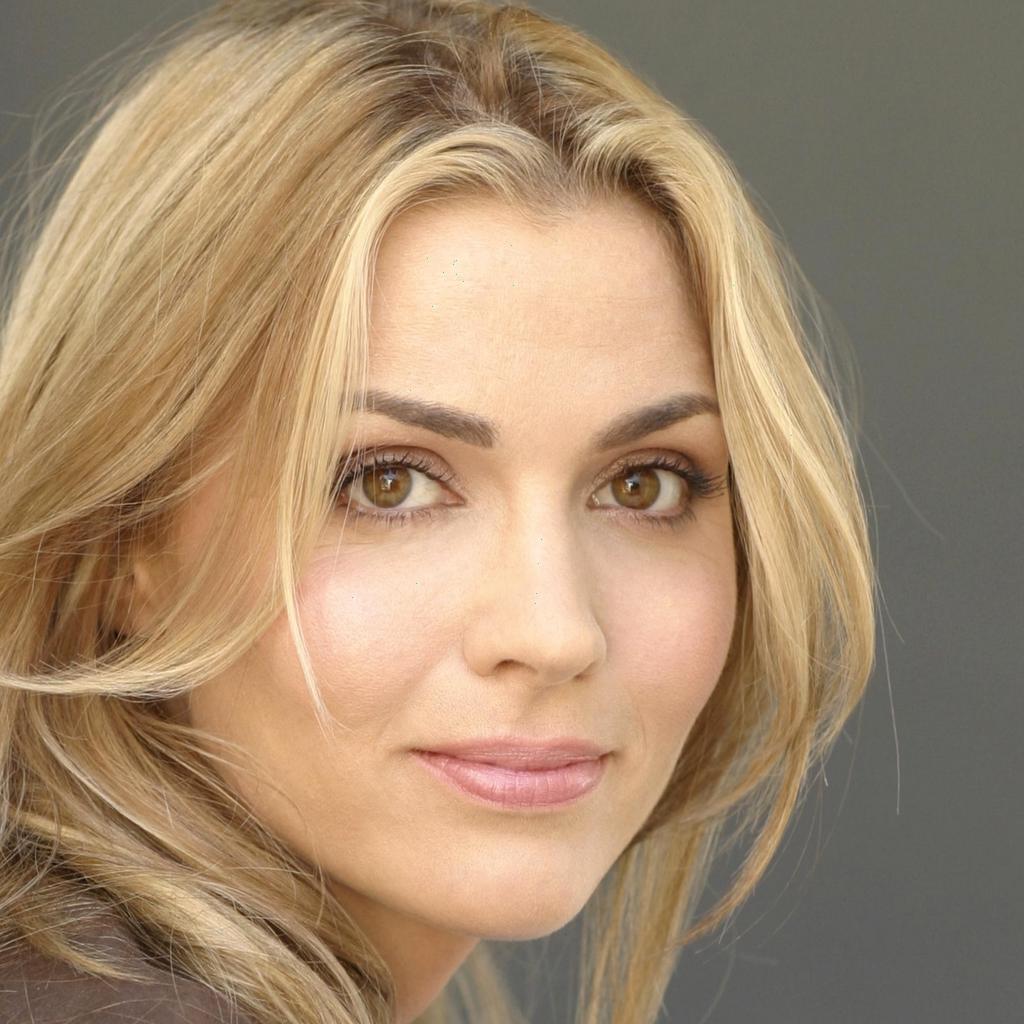}&
    \includegraphics[width=.2\linewidth]{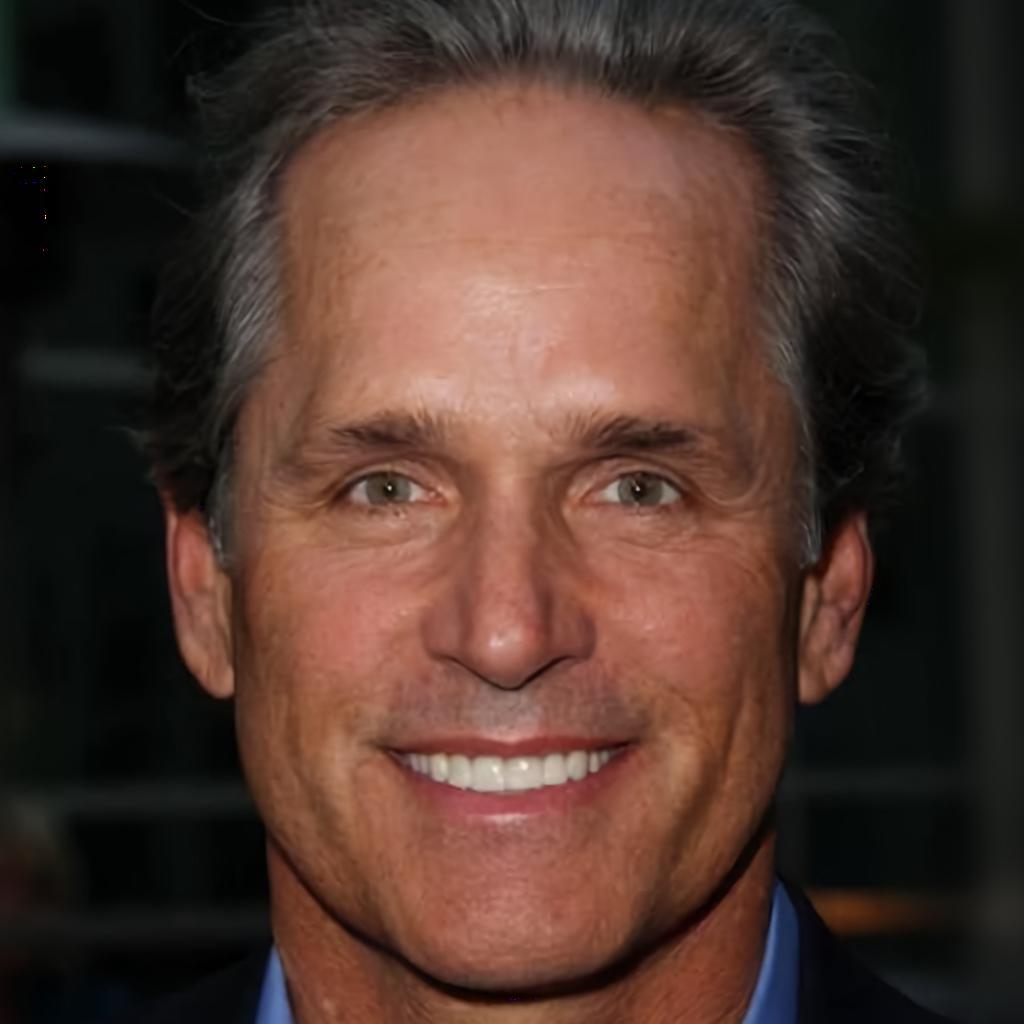}
    \\[2mm]
    
	\multicolumn{4}{c}{\em Latent recovery from the manifold} $\argmin_{G(z)} \| G(z) - y \|$\\
    \includegraphics[width=.2\linewidth]{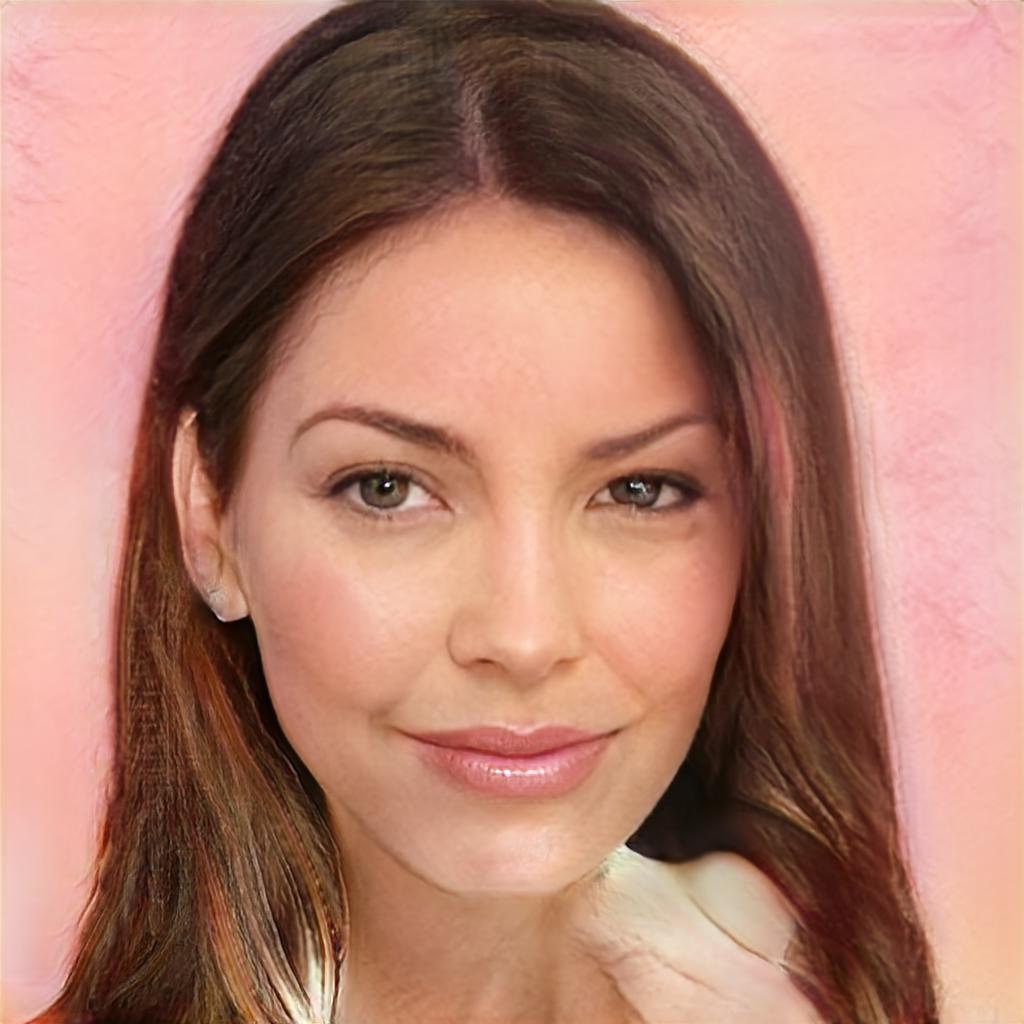} &
    \includegraphics[width=.2\linewidth]{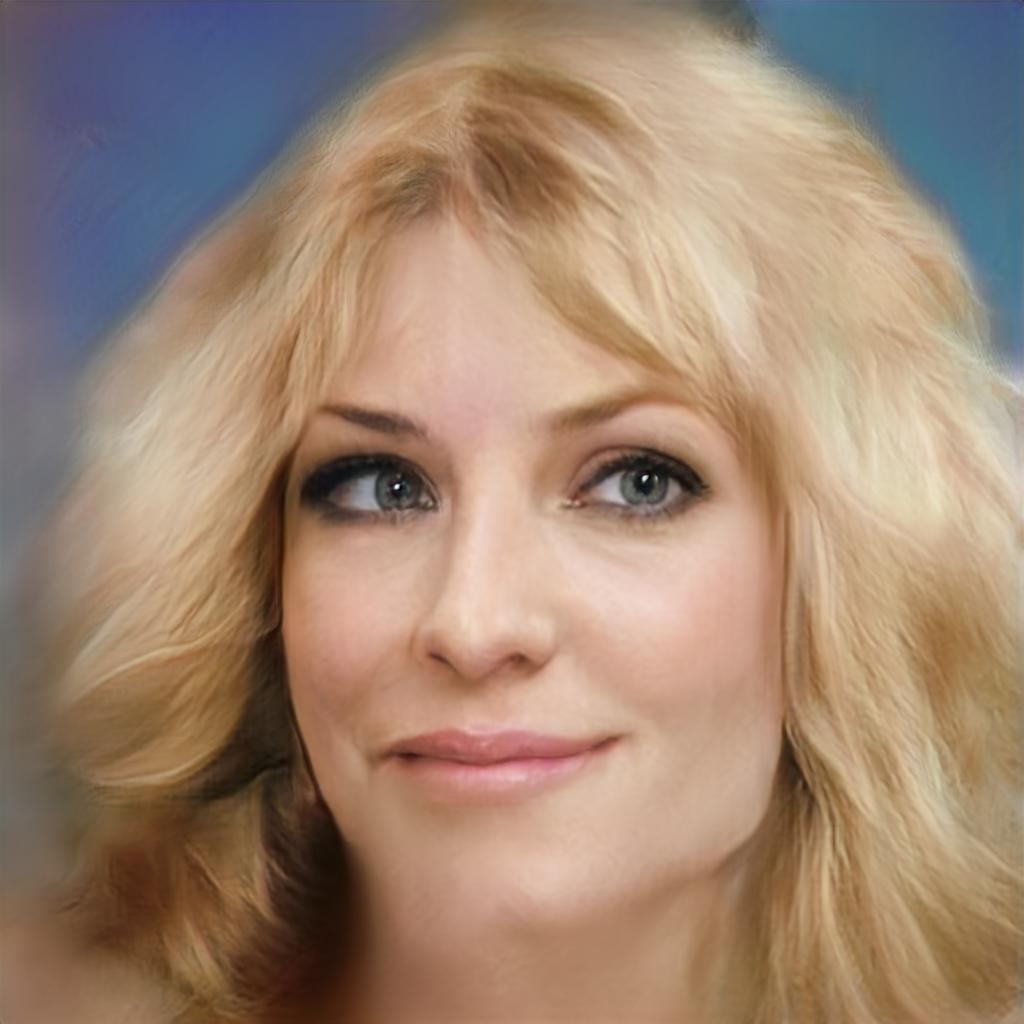} &
    \includegraphics[width=.2\linewidth]{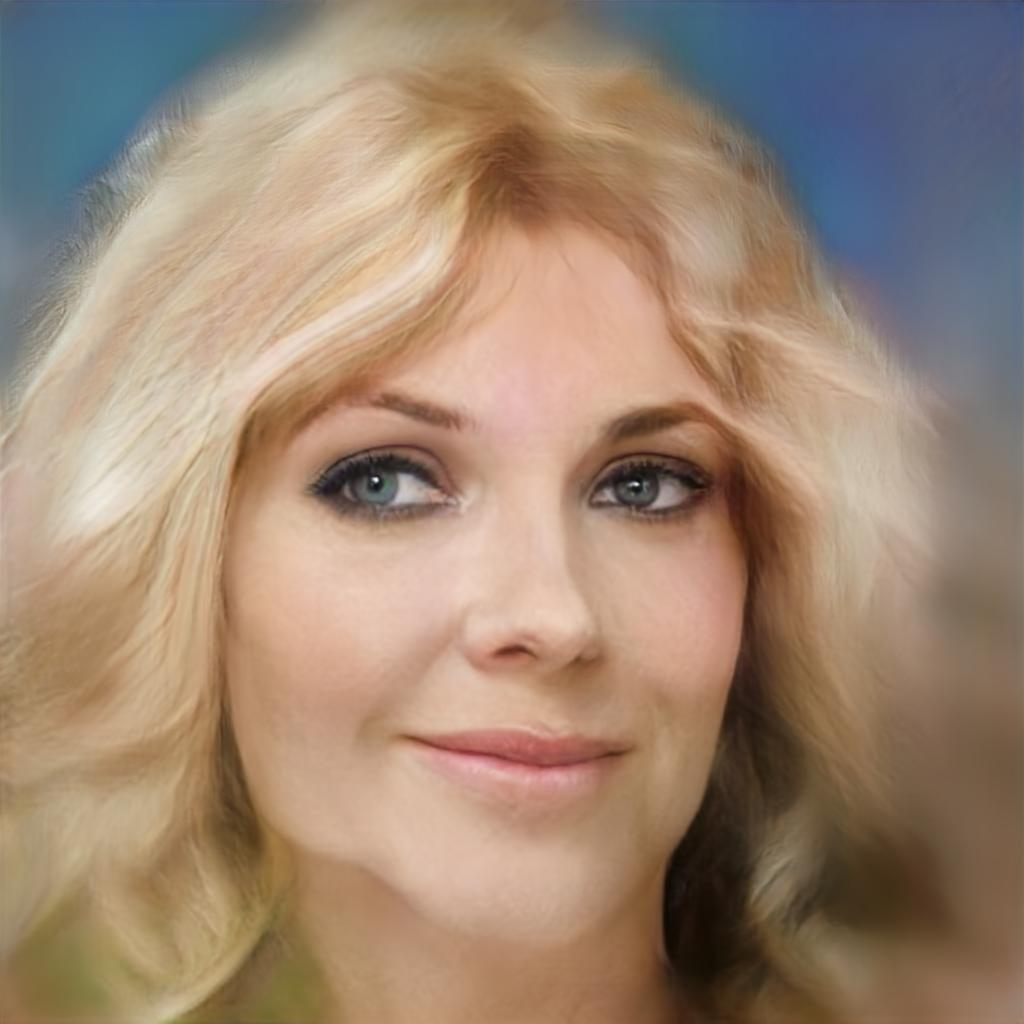} &
    \includegraphics[width=.2\linewidth]{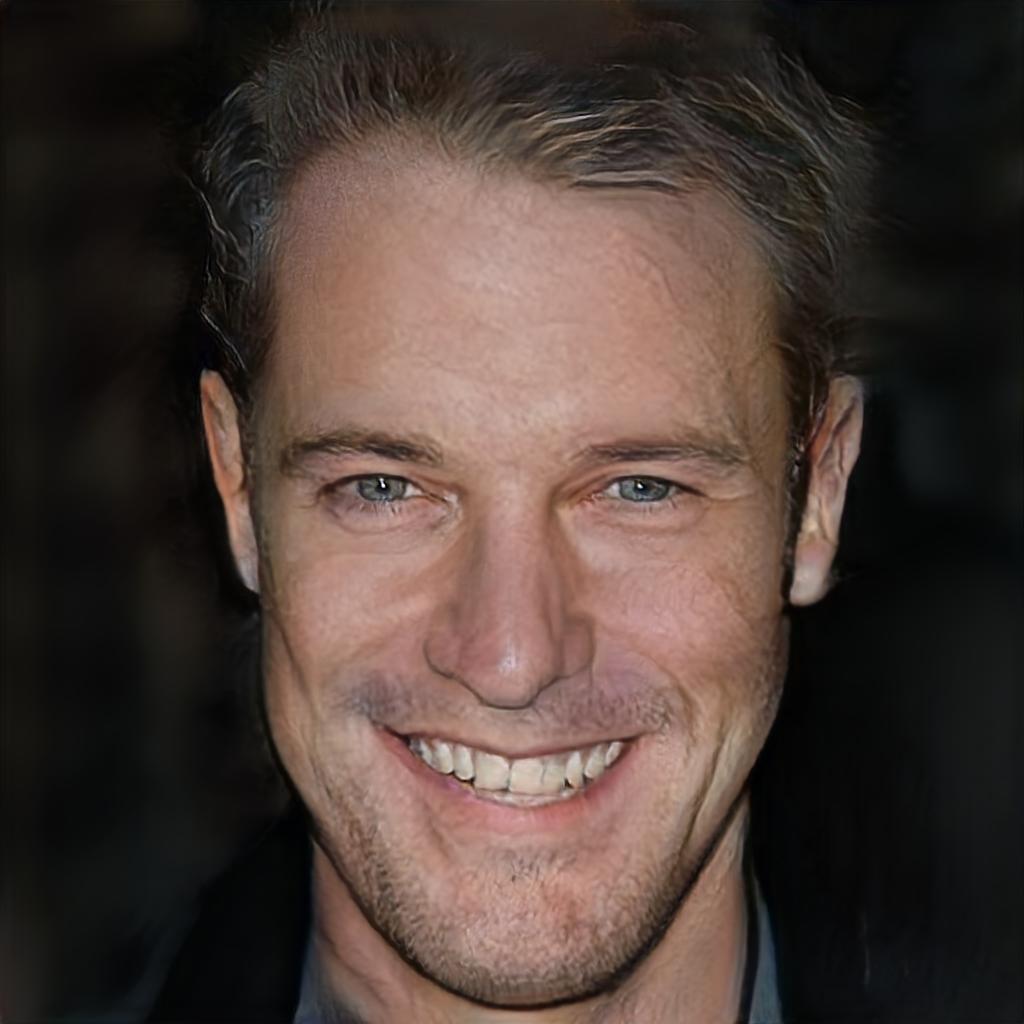} 
\end{tabular}

	\caption{\em \small Rather than inspecting the most similar pictures (NN) in the training dataset $\cal D$ for a few sampled generated images $G(z)$, as usually done for evaluating GANs, we consider in this work the opposite: finding the most similar image in the manifold $\mathcal G = G(\mathcal Z)$ of generated images and statistically test the discrepancy with the reference images. 
    Indeed, while the Euclidean distance is surprisingly good at recovering pose with registered face images, it is yet very sensitive to simple perturbations, such as flipping. Exploring the latent space makes it much more robust to such perturbation, as demonstrated in this work. Analysis of the discrepancy between NN from image in the train set $\cal D$ versus not in $\cal D$ makes it possible to detect overfitting by the generator $G$.}
    \label{fig:teaser}
\end{figure*}
\else
\newcommand{\STAB}[1]{\begin{tabular}{@{}c@{}}#1\end{tabular}}
\begin{figure}[!htb]
\centering
\begin{tabular}{c c c c}
	&
    \textbf{target $y$} &
    \textbf{$\text{NN}_{\cal G}(y)$} & 
	\textbf{$\text{NN}_{\cal D}(y)$} 
    \\[1mm]

    
    {\rotatebox[origin=c]{90}{$y \not\in \cal D$}}   &
    \raisebox{-0.5\height}{\includegraphics[width=.25\linewidth]{fig_teaser/woman_test4_query.jpg}} &
    \raisebox{-0.5\height}{\includegraphics[width=.25\linewidth]{fig_teaser/woman_test4_recovery_query.jpg}} &
    \raisebox{-0.5\height}{\includegraphics[width=.25\linewidth]{fig_teaser/woman_test4_query_NN_MSE.jpg}}
    \\[10mm]
    
    {\rotatebox[origin=c]{90}{$\phi(y)$ (pool)}}   &
    \raisebox{-0.5\height}{\includegraphics[width=.25\linewidth]{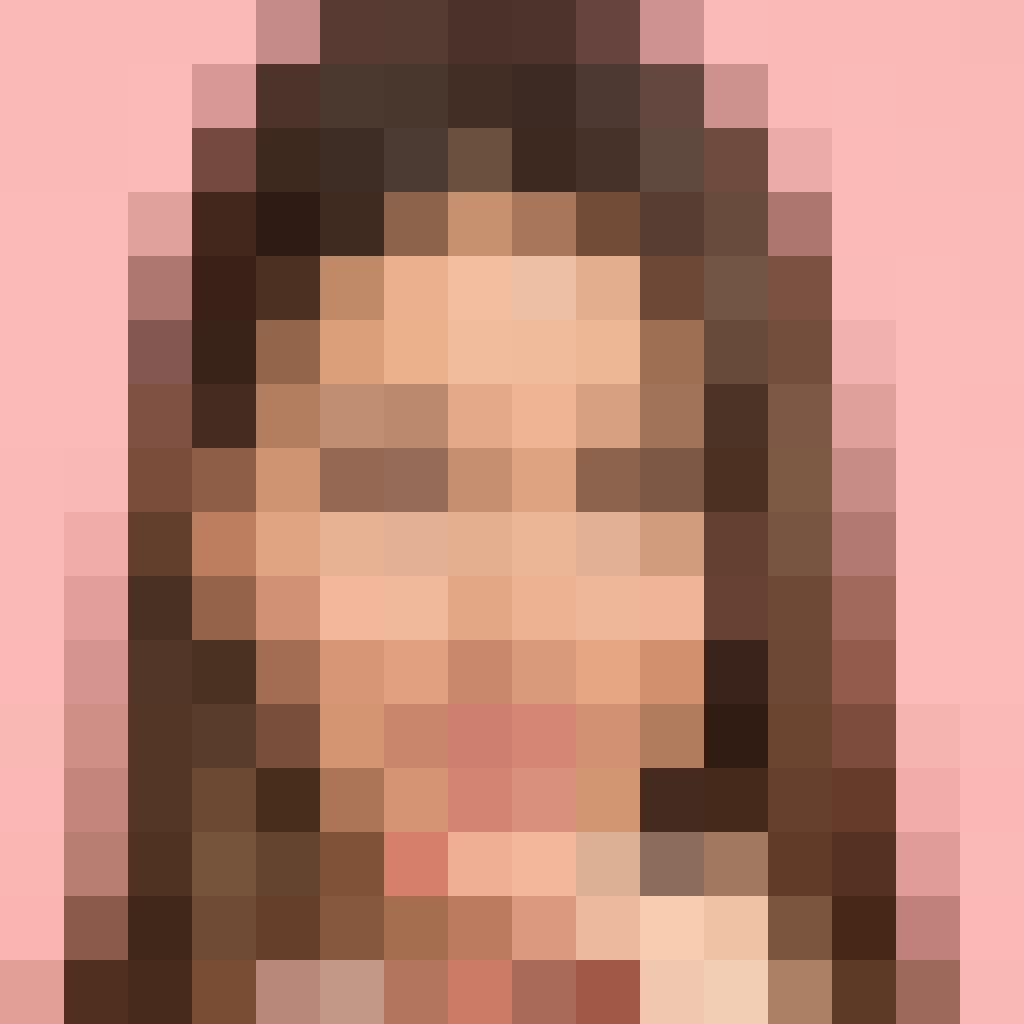}}  &
    \raisebox{-0.5\height}{\includegraphics[width=.25\linewidth]{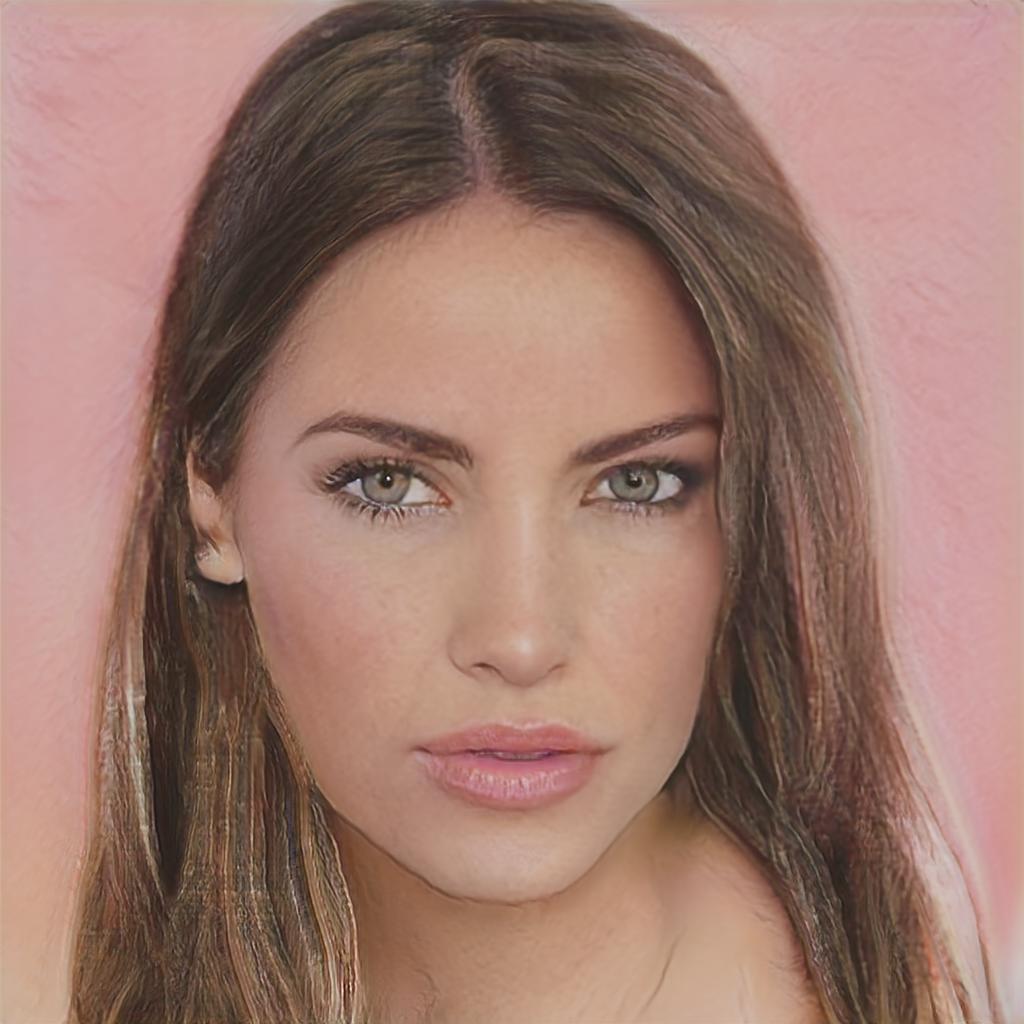}}  &
    \raisebox{-0.5\height}{\includegraphics[width=.25\linewidth]{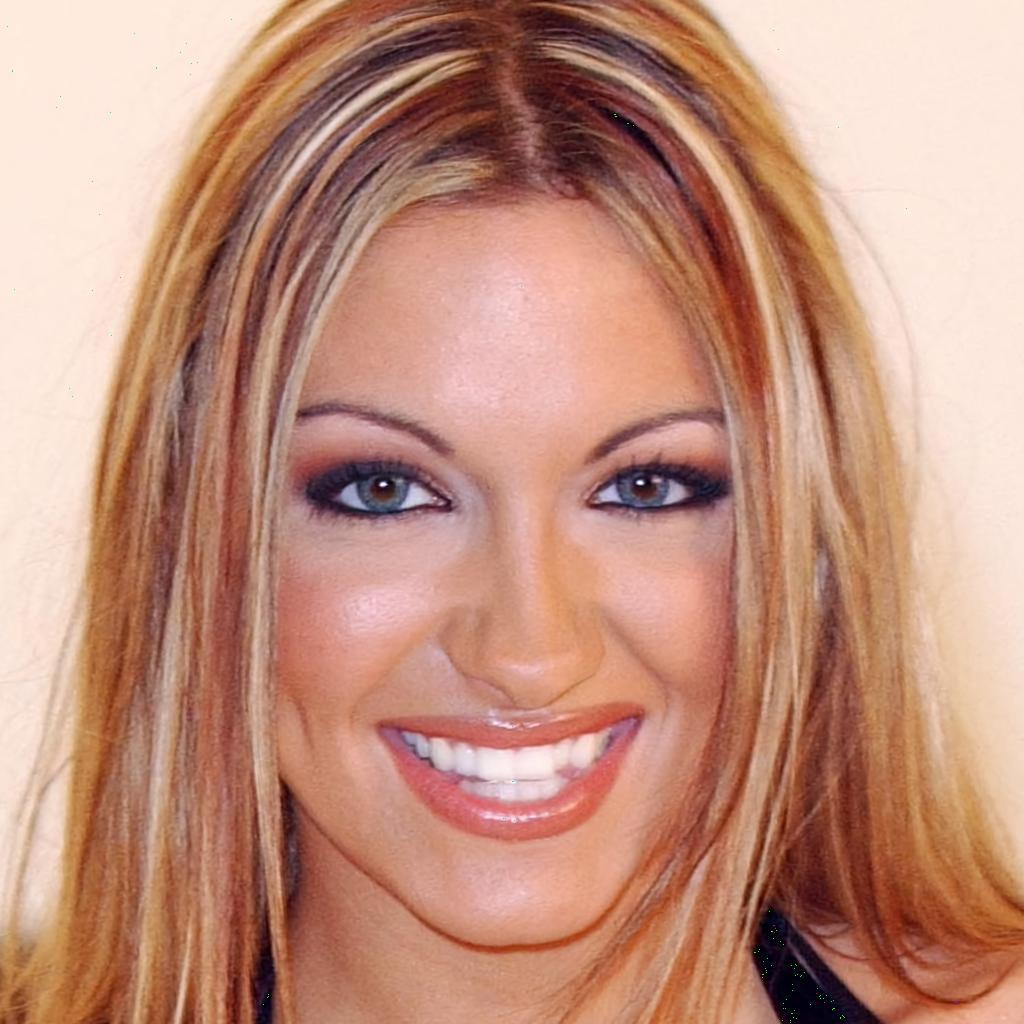}} 
    \\[10mm]
    
    {\rotatebox[origin=c]{90}{$G(z)$}}  &
    \raisebox{-0.5\height}{\includegraphics[width=.25\linewidth]{fig_teaser/fake0_target.jpg}} &
    \raisebox{-0.5\height}{\includegraphics[width=.25\linewidth]{fig_teaser/fake0_recovery_target.jpg}} &
    \raisebox{-0.5\height}{\includegraphics[width=.25\linewidth]{fig_teaser/fake0_target_NN_MSE.jpg}}
    \\[10mm]
    
    {\rotatebox[origin=c]{90}{$y \in \cal D$}} &
    \raisebox{-0.5\height}{\includegraphics[width=.25\linewidth]{fig_teaser/02014_target.jpg}} &
    \raisebox{-0.5\height}{\includegraphics[width=.25\linewidth]{fig_teaser/02014_recovery_target.jpg}} &
    \raisebox{-0.5\height}{\includegraphics[width=.25\linewidth]{fig_teaser/02014_target.jpg}}
    \\[10mm]
    
    {\rotatebox[origin=c]{90}{$\phi(y)$ (flip)}} &
    \raisebox{-0.5\height}{\includegraphics[width=.25\linewidth]{fig_teaser/02014_query.jpg}} &
    \raisebox{-0.5\height}{\includegraphics[width=.25\linewidth]{fig_teaser/02014_recovery_query.jpg}} &
    \raisebox{-0.5\height}{\includegraphics[width=.25\linewidth]{fig_teaser/02014_query_NN_MSE.jpg}}
\end{tabular}

	\caption{\em Rather than inspecting the nearest neighbor (column \textbf{$\text{NN}_{\cal D}(y)$}) in the training dataset $\cal D$ (here CelebA-HQ) for a few sampled generated images $y=G(z)$, as usually done for evaluating GANs, we consider in this work the opposite: finding the most similar image (\textbf{$\text{NN}_{\cal G}(y)$}) in the manifold $\cal G$ of generated images and measure its discrepancy with the reference image $y \in \cal D$. 
    We demonstrate that analyzing the discrepancy for \textbf{$\text{NN}_{\cal G}$} with image $y \in \cal D$ versus $y \not\in\cal D$ makes it possible to detect training images that are memorized by the generator $G$.
    }
    \label{fig:teaser} 
\end{figure}
\fi

\section{Introduction and related work} 

In just a few short years, image generation with deep networks has gone from niche to the center piece of machine learning. This was largely initiated by Generative Adversarial Networks (GANs)~\cite{goodfellow2014generative}. Incredible progress has been made, from deep convolutional GAN (DCGAN)~\cite{radford2015unsupervised} producing clearly computer generated faces, to PGGANs~\cite{karras2017progressive} producing faces which are virtually indistinguishable from real ones even to human observers and at high resolution (see Fig.~\ref{fig:teaser}). 
Images have such realism in fact that it is somewhat standard to include nearest neighbors for generated images to suggest that training examples weren't memorized ~\cite{brock2018large,karras2017progressive}. Memorization\footnote{As in ~\cite{radford2015unsupervised}, the terms 'memorization' and 'overfitting' are used interchangeably.} is not simply a curiosity for generative models; in fact, it is easy to envision a setting where training data is private or copyrighted and should not appear in generated images. 
%
%

Fig.~\ref{fig:teaser} (last column) illustrates the nearest neighbor (NN) test, where $\text{NN}_{\cal D}(y)$ is the training dataset NN of a few images $y$. While NN with the Euclidean distance is surprisingly good at recovering pose with registered face images (last column in Fig.~\ref{fig:teaser}), it is very sensitive to simple perturbations, such as flipping (see the last two rows of Fig.~\ref{fig:teaser}). Consequently, giving the NN in the train set in not enough to guaranty that the GAN is not generating memorized images.

In contrast, we suggest to rely on the opposite methodology by optimizing the latent code $z \sim \mathcal Z$ to find the nearest neighbors $\text{NN}_{\cal G}(y)$ in the manifold of generated faces $\mathcal G = \{G(z)\}_{z \sim Z}$ of images from the training ($y \in \cal D$) and a validation set ($y \not\in \cal D$).
As shown in the second column of Fig.~\ref{fig:teaser}, exploring the latent space makes NN search much more robust to such perturbation. 
%
Using this framework that we refer to as \emph{latent recovery}, two main contributions are proposed:
\begin{itemize}
\item A comprehensive study of Latent Recovery for generative networks. 
In Section~\ref{sec:recovery}, recovery errors reveal insight into the specificity of generators. 
We then successfully apply latent recovery to perform inpainting and face super-resolution in Section~\ref{sec:inpainting}.
\item Section~\ref{sec:memorization} introduces a simple yet novel way to detect memorization/overfitting in any deep generator via statistics of latent recovery errors on test and train sets. Overfitting is undetectable for GANs, which is corroborated visually in Fig.~\ref{fig:recovery} and statistically in Table~\ref{table:p_values}. Overfitting is however detectable in hybrid adversarial losses such as CycleGAN~\cite{zhu2017unpaired}, and easily detectable in non-adversarial generators such as GLO~\cite{bojanowski2017optimizing}. 
Additionally, we show that standard evaluation metrics do not detect overfitting in some models.
\end{itemize}

\subsection{Related Work}

Adversarial like losses have seen successful applications in a myriad of settings: unpaired image to image translation in the hybrid adversarial loss in CycleGAN~\cite{zhu2017unpaired}, face attribute modification in StarGAN~\cite{choi2018stargan} and various image inpainting techniques~\cite{iizuka2017globally,yeh2017inpainting} to name a few. This progress has created a huge need to evaluate generated image quality, which to some degree has not been fully answered~\cite{borji2018pros}.

\paragraph{GAN Evaluation Metrics}
The Fr\'echet Inception Distance (FID)~\cite{heusel2017gans} has been adopted as a popular and effective way to evaluate generative models which can be sampled. 
In the wide scale study \cite{lucic2017gans}, FID was used to compare a huge variety of GANs, wherein it was shown auxiliary factors such as hyperparameter tuning can obfuscate true differences between GANs. 
In \cite{sajjadi2018assessing}, notions of precision and recall are introduced for generated images, to help characterize model failure rather than providing a scalar in image quality. 

\paragraph{Overfitting}
For image classification, overfitting is defined as the discrepancy between the performance of a classifier on the training set  and a hold out set of images. As a testament to the memorization ability of deep networks, Zhang et al.\cite{zhang2016understanding}, demonstrated arbitrary labels can be perfectly memorized for enormous sets of training images. 

Notions of generalization for generative models are less standardized. 
In DCGAN ~\cite{radford2015unsupervised}, questions about overfitting/memorization are already raised.
In \cite{arora2017generalization}, the authors defined generalization for GANs in a largely theoretical setting. The formulation, however, was used to suggest a new GAN training protocol rather than provide an evaluation technique. In \cite{arora2017gans}, the support of a GAN generator, in terms of the number of face identities it could produce, was estimated using the birthday paradox heuristic. While crude, it suggested the support of faces could be quite large with respect to the size of the training set. 

\paragraph{Memorization and Privacy}
Beyond these aspects of memorization and practical evaluation of generators lies the important and debated issue of privacy: 
How to ensure that the data used for training cannot leak by some reverse engineering, such as CNN inversion~\cite{mahendran2015understanding} or reconstruction from feature~\cite{Weinzaepfel_CVPR11} ?
Because GANs have seen such widespread real world application for image editing, it is imperative that we have better evaluations tools to assess how much these networks have overfit the training data. 
For example, if a user is using a neural net to inpaint faces as in \cite{li2017generative} or to perform super-resolution enhancing~\cite{dahl2017pixel}, it seems necessary to ensure verbatim copies of training images do not appear, due to privacy or even copyright concerns. 
Indeed, several  attacks against machine learning systems have been exposed in the literature \cite{papernot2016attack}.
For instance, authors in \cite{fredrikson2015inversion} designed an inversion attack to coarsely recover faces used during the training of a white box facial recognition neural network.
More recently, \cite{shokri2017membership} designed a membership attack allowing to determine if a given image was part of the training set of a black box machine learning system. In essence, this membership attack exploits the irregularities on how the classifier has overfit the training data.
\section{Reconstruction by Latent Code Recovery}\label{sec:recovery}
This section proposes a methodology for reconstructing the most similar images to target images with an existing generator.

Inversion of deep representations has been already addressed in the literature.  \cite{mahendran2015understanding}  used a simple optimization procedure to maximize an output class of the VGG-like network. This demonstrated a key insight, namely that such deep nets represent progressively more abstract and large-scale image features. 
In the seminal works of \cite{nguyen2015deep,szegedy2013intriguing}, a similar inversion of deep nets unveiled adversarial examples, which shows that somehow deep networks can be easily fooled. In \cite{lucic2017gans}, \textit{generative} networks are inverted to study recall, which is the ability of the network to reproduce all images in the dataset. Similarly, \cite{metz2016unrolled} used latent recovery of a GAN generator to evaluate its quality, with the hypothesis that a high quality generator also should recover data samples with high fidelity. 


Other works tackle recovering latent codes directly by training an encoder network to send images back from image space to latent space, such as the BEGAN model \cite{berthelot2017began} or Adversarially Learned Inference (AGI) \cite{dumoulin2016adversarially}. In generative latent optimization (GLO) \cite{bojanowski2017optimizing}, a generative model is trained along with a fixed-size set of latent codes, so that they are known explicitly when training finishes.

In this paper, we will proceed by recovering latent codes via optimization, following \cite{metz2016unrolled,bora2017compressed,lipton2017precise,lucic2017gans}. In contrast with \cite{metz2016unrolled}, we will be concerned with recovering images the generator has not seen, such as validation images or deliberately distorted images in order to gain insight into the generator.

\begin{figure}[!tb]
\centering
\if 0 
  \begin{subfigure}{.33\linewidth}
      \centering
      \includegraphics[width=1\linewidth]{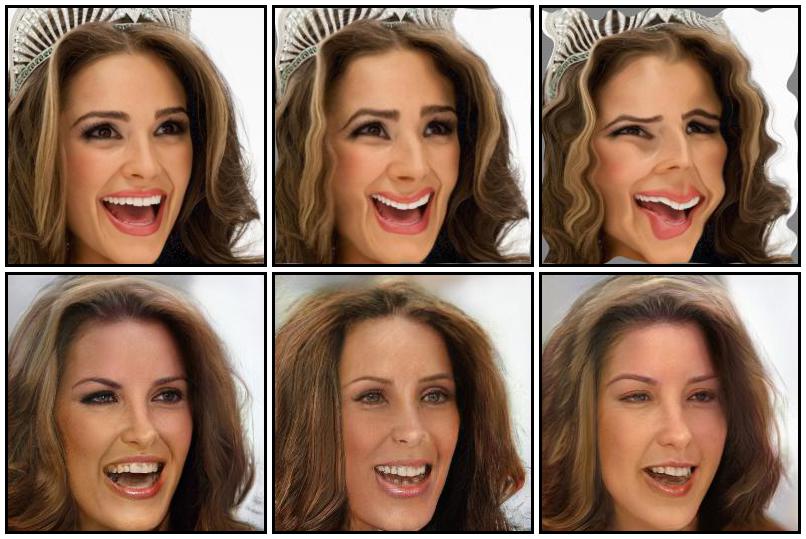}
  \end{subfigure}
  \begin{subfigure}{.33\linewidth}
      \centering
      \includegraphics[width=1\linewidth]{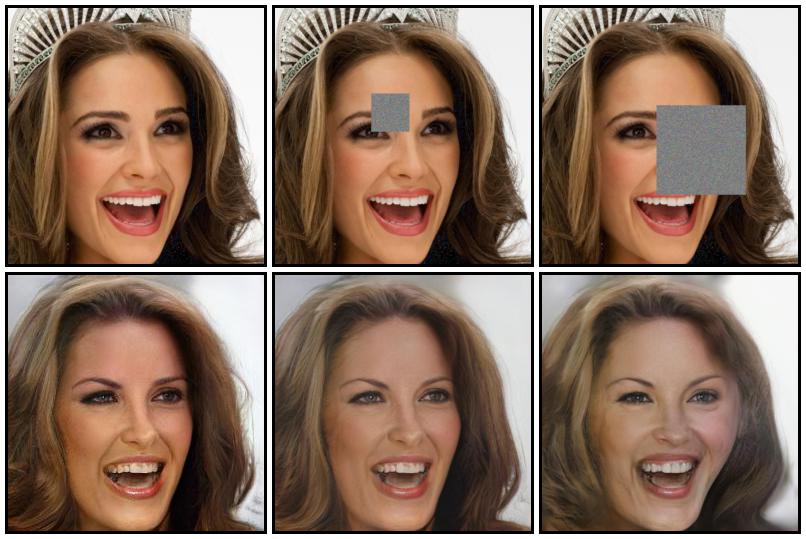}
  \end{subfigure}
  \begin{subfigure}{.33\linewidth}
      \centering
      \includegraphics[width=1\linewidth]{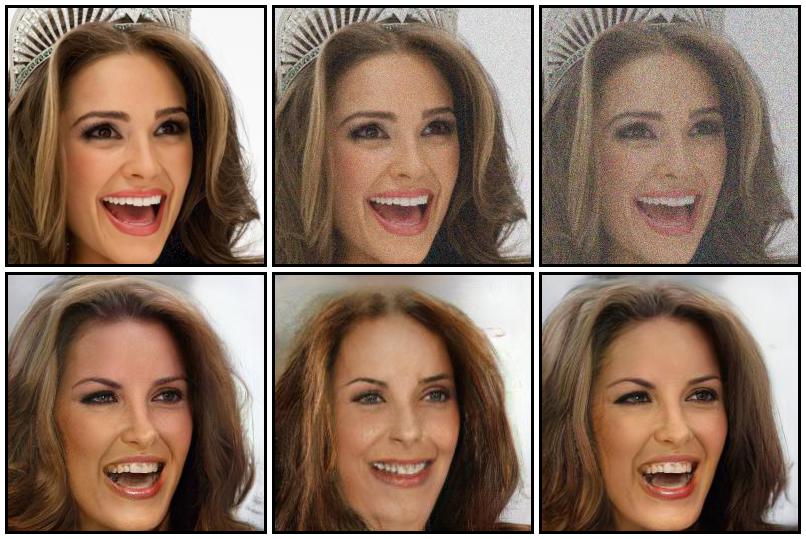}
  \end{subfigure}

  \centering
  \begin{subfigure}{.33\linewidth}
      \centering
      \includegraphics[width=1\linewidth]{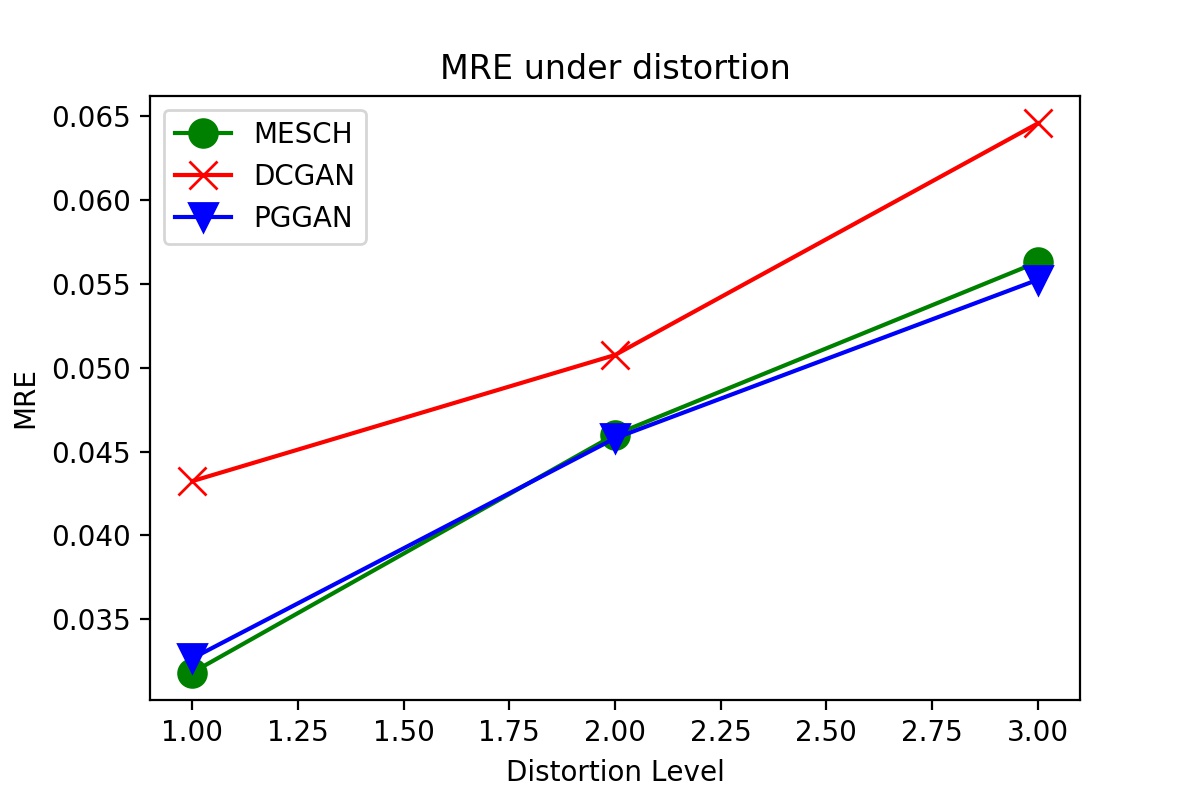}
      \caption{Warping}\label{sub:warp}
  \end{subfigure}
  \begin{subfigure}{.33\linewidth}
      \centering
      \includegraphics[width=1\linewidth]{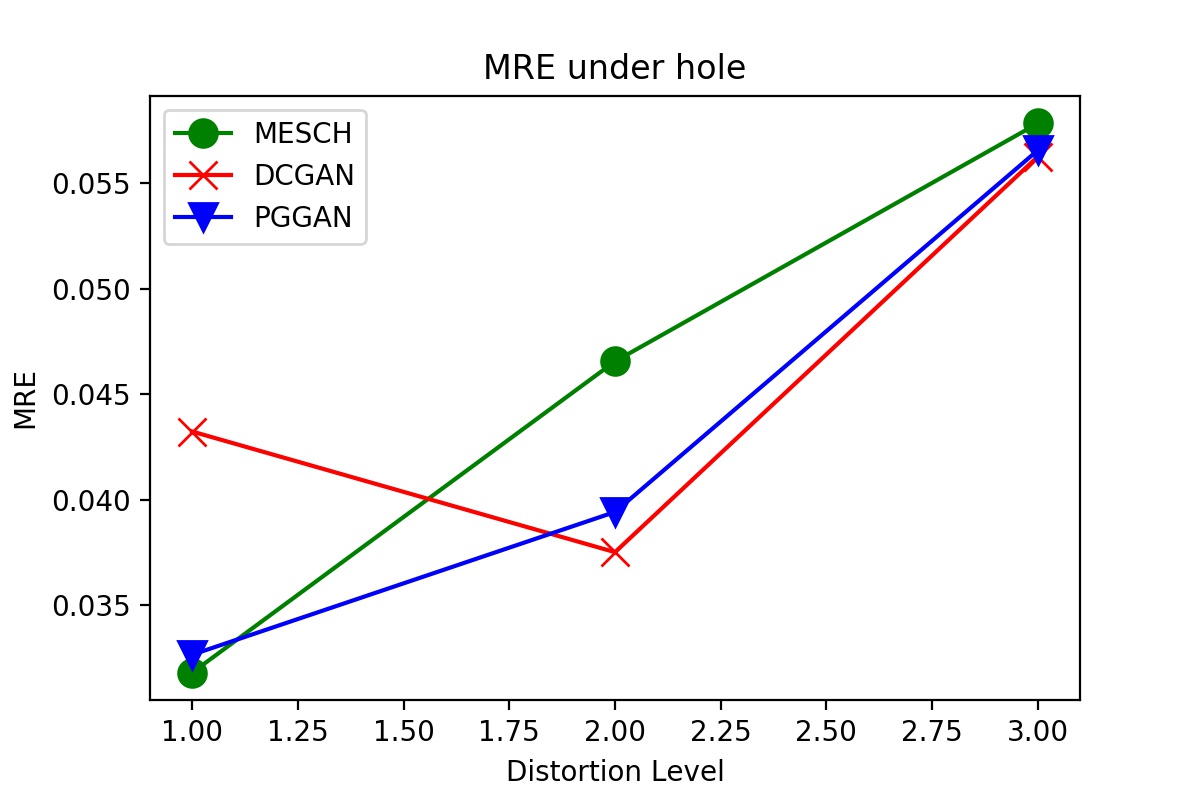}
      \caption{Hole}\label{sub:hole}
  \end{subfigure}
  \begin{subfigure}{.33\linewidth}
      \centering
      \includegraphics[width=1\linewidth]{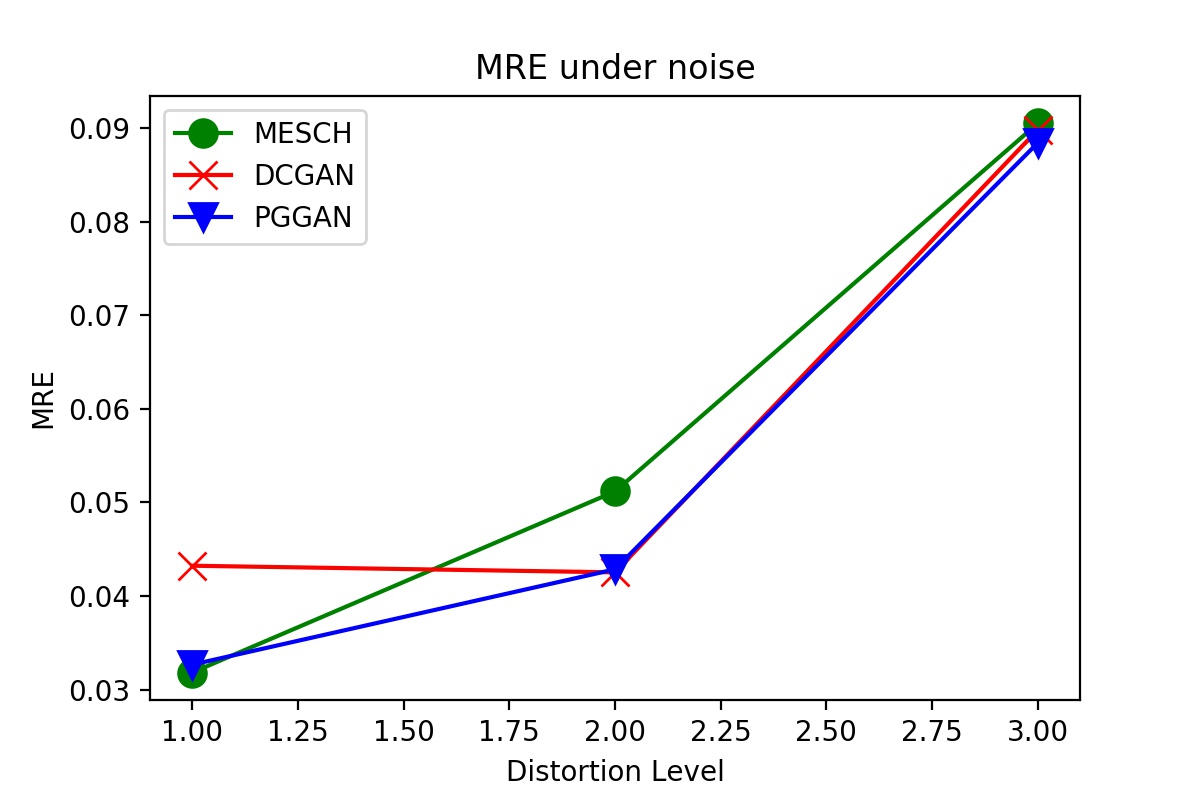}
      \caption{Noise}\label{sub:noise}
  \end{subfigure}

\else  

  \begin{subfigure}{\linewidth}
      \centering
      \includegraphics[width=.45\linewidth]{fig_distort/fig_distort.jpg}
      \includegraphics[width=.45\linewidth]{fig_distort/MRE_under_distortion.jpg}
      \caption{deformation by smooth diffeomorphism (warping)}\label{sub:warp}
  \end{subfigure}
  
  \centering
  \begin{subfigure}{\linewidth}
      \centering
      \includegraphics[width=.45\linewidth]{fig_distort/fig_hole.jpg}
      \includegraphics[width=.45\linewidth]{fig_distort/MRE_under_hole.jpg}
      \caption{Unsupervised inpainting (face completion)}\label{sub:hole}
  \end{subfigure}

  \centering
  \begin{subfigure}{1\linewidth}
      \centering
      \includegraphics[width=.45\linewidth]{fig_distort/fig_noise.jpg}
      \includegraphics[width=.45\linewidth]{fig_distort/MRE_under_noise.jpg}
      \caption{Additive white noise}\label{sub:noise}
  \end{subfigure}
\fi
\caption{\em Median recovery error (MRE, see Eq.~\eqref{def:MRE}) for 1800 test 
images on various GAN generators (PGGAN~\cite{karras2017progressive}, MESCH~\cite{mescheder2018training} and DCGAN~\cite{radford2015unsupervised}) against to various distortions $\phi$ in latent recovery optimization~\eqref{eq:recovery_L2} (see text for details). 
}
   \label{fig:recovery distortion}
\end{figure}

\subsection{Latent Code Recovery with Euclidean Loss}
As demonstrated in this section, the simple Euclidean loss is effective at recovering latent codes for a variety of GAN methods.  This is in line with 
many previous works which demonstrate the efficacy of Euclidean losses used at the output of a generative network. 
In~\cite{ulyanov2018deep}, Euclidean losses were shown to be sufficient for a variety of image processing tasks, when a deep generator helps automatically regularize the loss.  
Here, we consider the following \emph{latent recovery} optimization problem
%
\begin{equation}\label{eq:recovery_L2}
    z^\star(y) \in \argmin_{z} {\| \phi(G(z)) - \phi(y)\|}^{2}_{2}
    \tag{$\text{NN}_{\mathcal G}$}
\end{equation}
where $G$ is a deep generative network, $z$ is the input latent vector and $y$ is the target image. 
Using a solution $z^*$ of Problem~\eqref{eq:recovery_L2}, 
we denote by $\text{NN}_{\mathcal G}(y) = G(z^*)$ the Nearest Neighbor recovery of a given image $y$ in the set of generated images, as opposed to the usual NN search in a dataset $\cal D$: $\text{NN}_{\mathcal D}(y) = \argmin_{x \in \cal D} \|x-y\|$.
In this work, we consider mostly $\phi$ as the identity, but other operators are discussed in the next paragraph and in Section~\ref{sec:inpainting}.
Fig.~\ref{fig:teaser} illustrates the difference between the 2 NN search on a few examples.

\begin{figure*}[!htb]
\vspace*{-5mm}
  \centering
  \begin{subfigure}{.05\linewidth}
  \begin{tabular}{c}
  {\rotatebox[origin=t]{90}{target $y$}} 
  \\[10mm]
  {\rotatebox[origin=t]{90}{PGGAN}} 
  \\[10mm]
  {\rotatebox[origin=t]{90}{MESCH}} 
  \\[10mm]
  {\rotatebox[origin=t]{90}{GLO}} 
  \\[10mm]
  {\rotatebox[origin=t]{90}{AEGAN}} 
  \end{tabular}
  \end{subfigure}
  \begin{subfigure}{.85\linewidth}
        \begin{tabular}{ >{\centering}m{.5\linewidth} >{\centering}m{.5\linewidth} }
          $y \in \cal D$ (train)
          &
          $y \in \cal T$ (validation) 
        \end{tabular}
  
  		\includegraphics[width=1\linewidth]{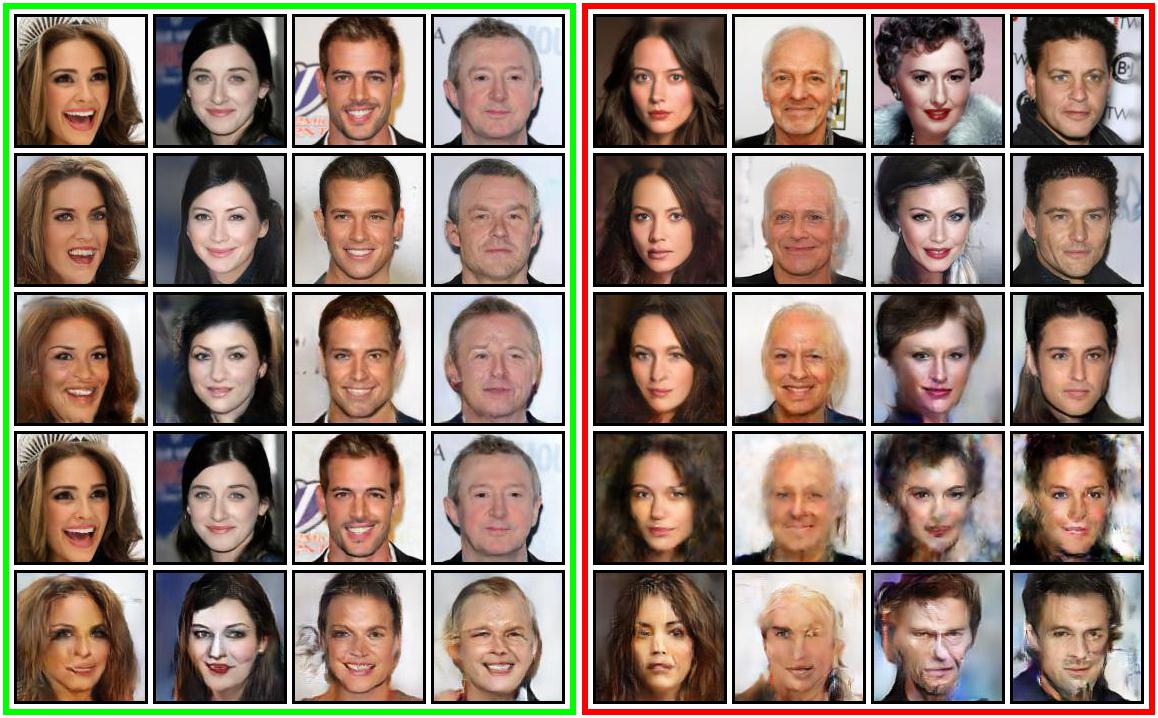}
  \end{subfigure}      
        
  \caption{\em Latent recovery of training images from $\cal D$ (left, green frame) and test images from $\cal T$ (right, red frame) for 128x128 images of Celeba-HQ~\cite{karras2017progressive}. From top row to bottom are first
  \textbf{target} images, and then recovery from
  Progressive GANs~\cite{karras2017progressive} (\textbf{PGGAN}), 
  0-GP resnet GAN~\cite{mescheder2018training} (\textbf{MESCH}), 
  a \textbf{GLO} network \cite{bojanowski2017optimizing},
  and finally 
  an {AutoEncoder-GAN} network \cite{zhu2017unpaired} (\textbf{AE-GAN}).
  While GLO obviously shows some memorization of training examples, it is hard to visually assess when overfitting happens for other methods, as discussed in Section~\ref{sec:memorization} (with additional details on architectures and training).
  \vspace*{-3mm}
  }
   \label{fig:recovery}
\end{figure*}

\paragraph{Experimental validation}
In every experiment, we employ LBFGS and noted it converges roughly 10x faster than SGD (successful recovery requiring approximately 50 iterations as opposed to 500 in \cite{bora2017compressed,lipton2017precise}). 
Although Problem.~\eqref{eq:recovery_L2} is highly non-convex, the proposed latent recovery optimization is works surprisingly well, as shown in Fig.~\ref{fig:teaser} and Fig.~\ref{fig:recovery}.
In particular for generated images $y=G(z)$, where $\text{NN}_{\mathcal G}(y) = z$, a global minimum (verbatim copy) is consistently achieved (see third row of Fig.~\ref{fig:teaser} and Fig.~\ref{fig:recovery distortion}). Visual inspection revealed that every network analyzed in this document could verbatim recover generated images, an observation also noted by \cite{lipton2017precise}, exemplified by the tight distribution of errors near zero in Fig.~\ref{fig:glo_memorization}.
Note that we also considered the widely used perceptual loss~\cite{johnson2016perceptual} by taking $\phi$ to be VGG features, with either no improvement or even degradation of visual results.

\subsection{Latent Code Recovery Under Distortion}
The Fr\'echet Inception Distance (FID), introduced in \cite{heusel2017gans}, is the \emph{de facto} standard for evaluating the quality of generated GAN images. The FID is computed by modeling the probability distribution of vision relevant features at the output of the Inception network \cite{salimans2016improved} and was shown to be in line with human inspection. As an important demonstration, the FID was shown to be sensitive to a variety of image transformations and corruptions, while other metrics, such as the inception distance \cite{salimans2016improved}, were not sensitive. 
In a similar vein, we wish to demonstrate that Eq.~\eqref{eq:recovery_L2} is meaningful, by showing that it responds to various distortions $\phi$. 
We choose $\phi$ to be one of the three distortions that are illustrated in Fig.~\ref{fig:recovery distortion}:
\smallskip
\begin{itemize}[nosep]
\setlength{\itemindent}{0em}
\item \textbf{Smooth Vector Field Warp (Fig.~\ref{sub:warp})} 
Following~\cite{heusel2017gans,zhang2018unreasonable} we warp training images by bilinear interpolation with a smooth 2D vector field $V_{\sigma_d} = V*g$, which is obtained from the Gaussian smoothing $g$ of a Gaussian random vector field $V(x) \sim \mathcal{N}(0,\,\sigma_{d}^{2})$; 
%

\item \textbf{Corruption Noise Patches (Fig.~\ref{sub:hole})} 
As~\cite{li2017generative}, we corrupt training images by replacing patches of various sizes with fixed Gaussian noise with variance $\sigma_{d}^{2}$; 

\item \textbf{Additive Noise (Fig.~\ref{sub:noise})} 
We add noise to each training image with $X_{n} = X + W_{d}$, where $W_{d}$ is sampled from a Gaussian distribution $W_{d}\sim \mathcal{N}(0,\,\sigma_{d}^{2})$. 
\end{itemize}
\smallskip

\paragraph{Experimental Validation} Fig. \ref{fig:recovery distortion} demonstrates a number of insights into latent recovery in Eq.~\eqref{eq:recovery_L2}. 
First of all, visual inspection reveals it is powerful enough to recover a face near the ground truth even if the image has been heavily distorted. 
It also demonstrates the specificity of the network, for example the three networks highlighted will reject images only slightly outside the manifold. In Table~\ref{table:p_values}, we can see that not all networks share the same specificity. For example, the GLO networks can recover distorted images with similar MRE's to training images, which means the networks are less precise. This is coupled with a lower FID of the network, for example see Fig.~\ref{fig:fid_vs_mre}. 

\section{Using Latent Recovery to Assess Overfitting}
\label{sec:memorization}

In the context of classification, overfitting pertains to the discrepancy between classifier performance on images for which it was trained and validation images not seen during training. This methodology is far less common with generative models, in part because there is still disagreement on how to properly evaluate them \cite{borji2018pros}. 
In this section, we extend the training and validation split to the context of generative models. Then, we analyze the difference between image recovery using Eq.~\eqref{eq:recovery_L2}, between training and validation images.

\subsection{Training Protocols}
We summarize the details of each generative model below, in terms of their training procedure and purpose within this work.

\noindent\textbf{GAN}~ 
Generative Adversarial Networks (GAN) involve a stochastic training procedure which simultaneously trains a discriminator and a generator. 
The original GAN~\cite{goodfellow2014generative} optimization problem writes
\begin{equation}\label{eqn: GAN}\vspace*{-2mm}
\max_{D}\min_{G} \mathbb{E}_{z\sim \mathcal{N}(0,1), x \sim p_{data}}[\mathcal{L}_{adv}(D,G,z,x)]
\end{equation}
where $\mathcal{L}_{adv}(D,G,z,x) = \log(D(x)) + \log(1- D(G(z)))$.
We examine three prominent GANs in the literature. First is DCGAN~\cite{radford2015unsupervised}, as it is one of the most widely used GAN architectures and with still decent performance across a variety of datasets \cite{lucic2017gans}.
Then we study two state-of-the-art GANs for high resolution generation; progressive growing of GANs \cite{karras2017progressive}, which we refer to as \textsc{pggan} and the zero centered gradient penalty Resnet presented in \cite{mescheder2018training}, which we refer to as \textsc{mesch}. 
We train these three GANs on CelebA-HQ with a training split of the first 26k images and the first 70k images of LSUN bedroom and tower.
We chose these splits to preserve the quality of each method, as GAN quality significantly degrades with small dataset sizes.

\noindent \textbf{Generative Latent Optimization (GLO)}~
The recently introduced Generative Latent Optimization (GLO) creates a mapping from a fixed set of latent vectors to training images.
The GLO objective is as follows
\begin{align}\label{eqn: GLO}
\min_{G} \sum_{(z_{i},x_{i})} \mathcal{L}_{rec}(G(z_{i}),x_{i}) := \| G(z_{i}) - x_{i}\|^{2}_{2}
\end{align}
where $x_{i}\in \mathcal{D}$ refers to training images, $z_{i}\sim \mathcal{N}(0,1)$ samples a Gaussian distribution and the pairs $(z_{i},x_{i})$ are drawn once and for all before training begins%
\footnote{ Contrary to \cite{bojanowski2017optimizing}, we do not optimize the latent space and found no negative impact on the reconstruction capacity and generation quality.}.
Because we know the latent distribution is Gaussian, we can easily sample the network after it is trained.
\smallskip

\noindent \textbf{AutoEncoder GAN (AEGAN)}~ A huge variety of recent methods use a hybrid adversarial plus reconstruction loss in their framework, for example being applied to unpaired domain translation and image inpainting \cite{choi2018stargan,li2017generative}.
We study a simple autoencoder plus adervsarial loss \textsc{aegan} with the objective
\begin{align}\label{eqn: cyclegan}
\max_{D} \min_{G,E} \sum_{x_{i}\in \mathcal{D}} \mathcal{L}_{rec}(G(E(x_{i})),x_{i})+ \mathcal{L}_{adv}(G,E,D,x_{i})
\end{align}

\noindent \textbf{CycleGAN}~
We train a CycleGAN using the official code, which has the following objective:
\iftrue
\begin{align}\label{eqn: cyclegan}
\max_{D_{1},D_{2}} \min_{G_{1},G_{2}} & \sum_{x_{i}\in \mathcal{X},y_{i}\in \mathcal{Y}} \nonumber 
\\ \nonumber 
&\mathcal{L}_{rec}(G_{2}(G_{1}(x_{i})),x_{i})+ \mathcal{L}_{adv}(G_{1}(x_{i}),D_{2}) 
\\ \nonumber
&+ \mathcal{L}_{rec}(G_{1}(G_{2}(y_{i})),y_{i})+ \mathcal{L}_{adv}(G_{2}(y_{i}),D_{1}) 
\\
&+ \mathcal{L}_{id}(G_{1}(x_{i}),x_{i})+ \mathcal{L}_{id}(G_{2}(y_{i}),y_{i}) 
\end{align}
\fi
%
%
where $G_{1,2}, G_{2,1}$ are autoencoders translating across domains and $D_{1}, D_{2}$ are discriminators. We use the $\mathcal{L}_{id}$ loss as it is enabled by default in the CycleGAN code.

Concerning models trained with a reconstruction loss (\textsc{glo}, \textsc{cyclegan} and \textsc{aegan}), we selected these architectures for a theoretical perspective, as they offer interesting windows into how generators can memorize.
In particular, we will study the impact of the training set size $N$ on the overfitting inclination. 
For example, while we were unable to train a good quality GAN with a small set (say 256) images, \textsc{glo} converges extremely quickly on such a small set of images.
See the \textsc{glo}-256 network in Fig.~\ref{fig:recovery} (4\textsuperscript{th} row) where memorization is immediately apparent. As a result, we will refer respectively to \textsc{glo}-$N$, \textsc{cyclegan}-$N$ and \textsc{aegan}-$N$, to account for this size.
Additionally, for the autoencoder models \textsc{aegan} and \textsc{cyclegan}, we forgo optimization and simply use the reconstruction losses from the autoencoder. For \textsc{AEGAN}, this corresponds to taking $z^{*}_{i} = E(x_{i})$ w.r.t. Eq.~\ref{eq:recovery_L2} and for \textsc{cyclegan} this corresponds to $|G_{2}(G_{1}(x)) - x|_{2}^{2}$ (or vice-versa for the opposite domain $y$). For \textsc{cyclegan}, we split both domains into test and train sets and then compute translations between the two train sets.

In the next paragraphs, we will proceed to show that it is possible for generative networks to memorize in the sense that validation and training sets have significantly different recovery error distribution.

\begin{figure}[!tb]
\centering
\begin{subfigure}{.49\linewidth}
    \centering
    \includegraphics[width=1\linewidth]{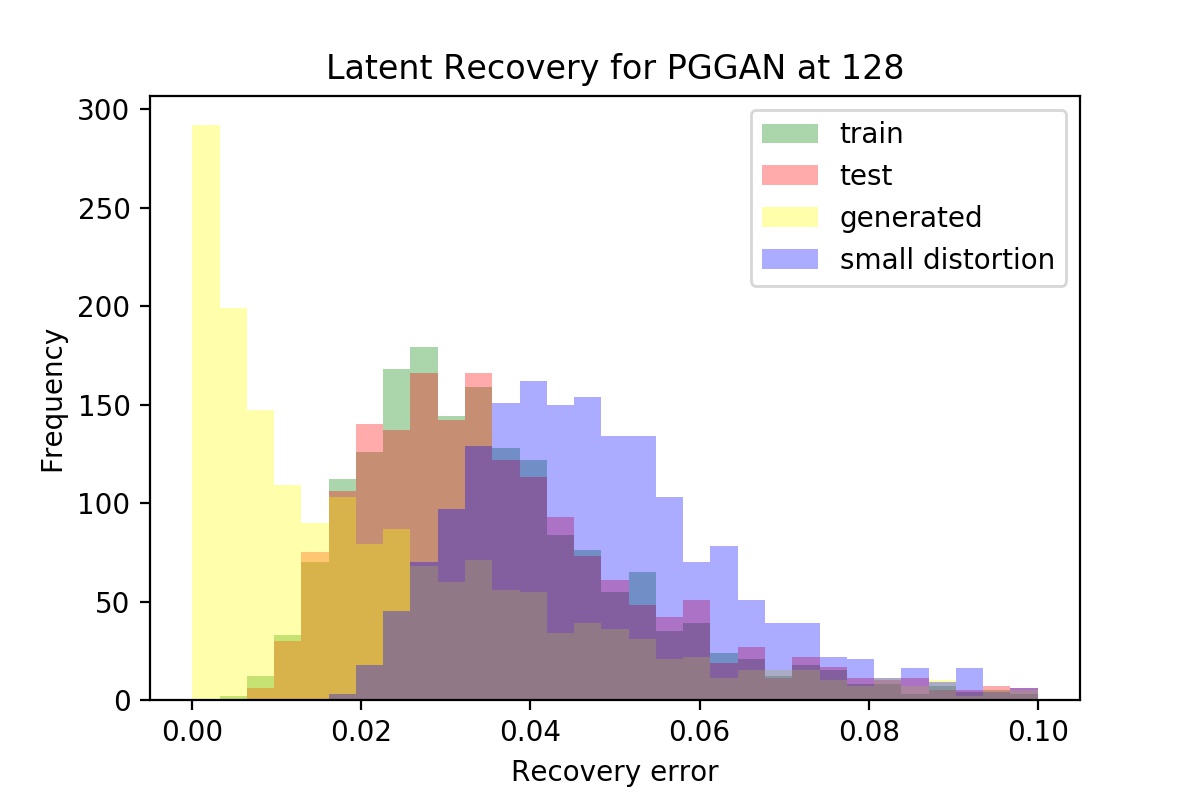}
	\caption{\textsc{pggan}}
\end{subfigure}
\begin{subfigure}{.49\linewidth}
    \centering
    \includegraphics[width=1\linewidth]{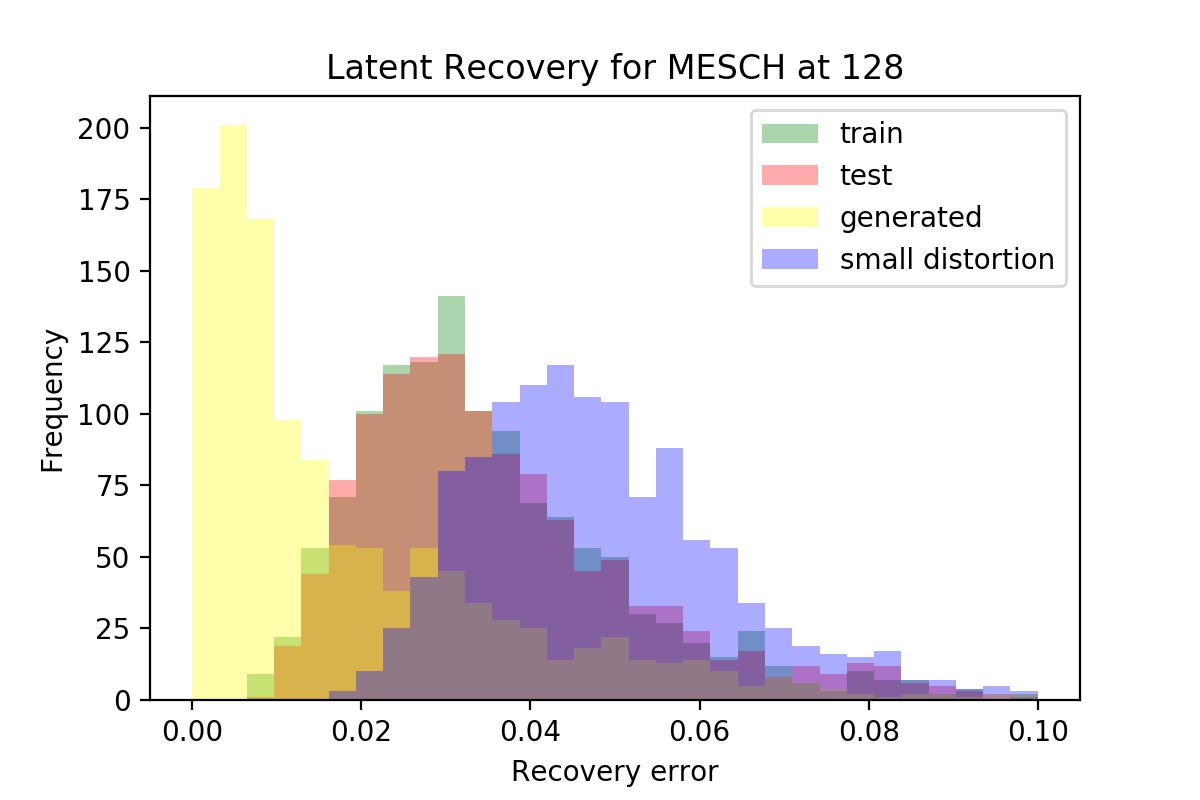}
	\caption{\textsc{mesch}}
\end{subfigure}

\centering
\begin{subfigure}{.49\linewidth}
    \centering
    \includegraphics[width=1\linewidth]{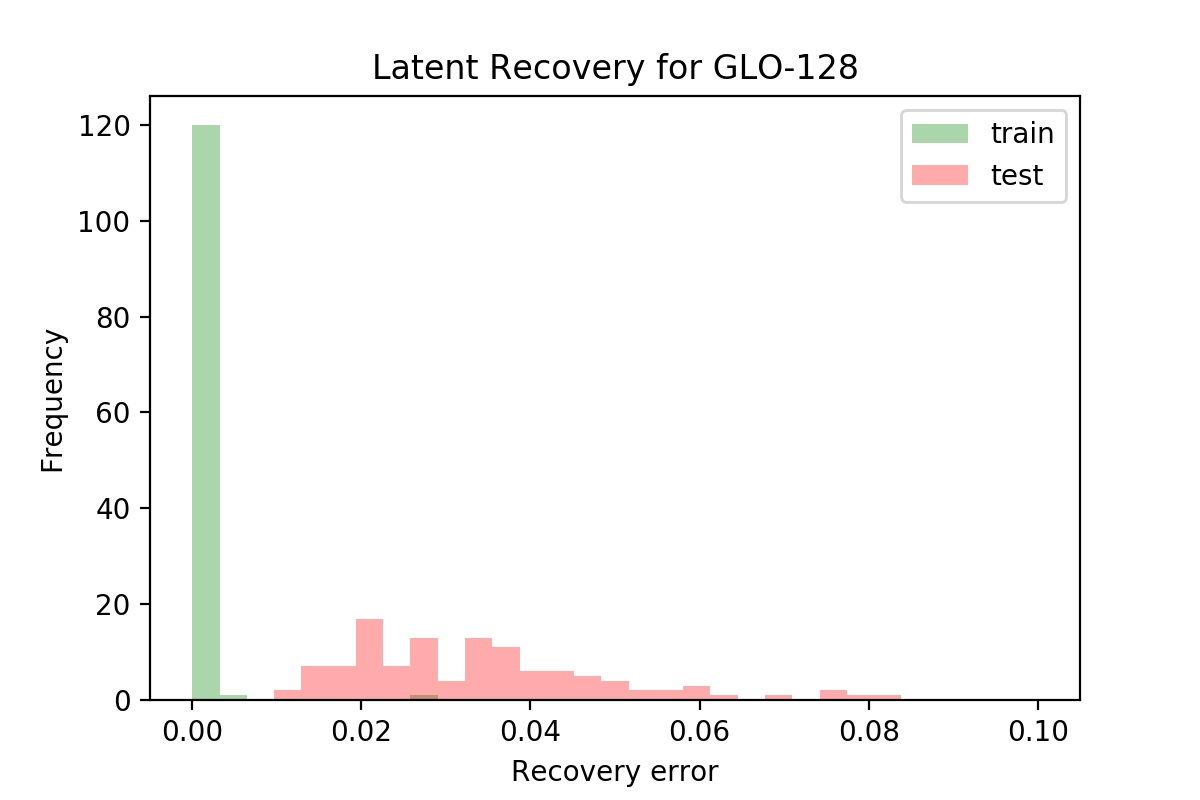}
	\caption{\textsc{glo}-128}
\end{subfigure}
\begin{subfigure}{.49\linewidth}
    \centering
    \includegraphics[width=1\linewidth]{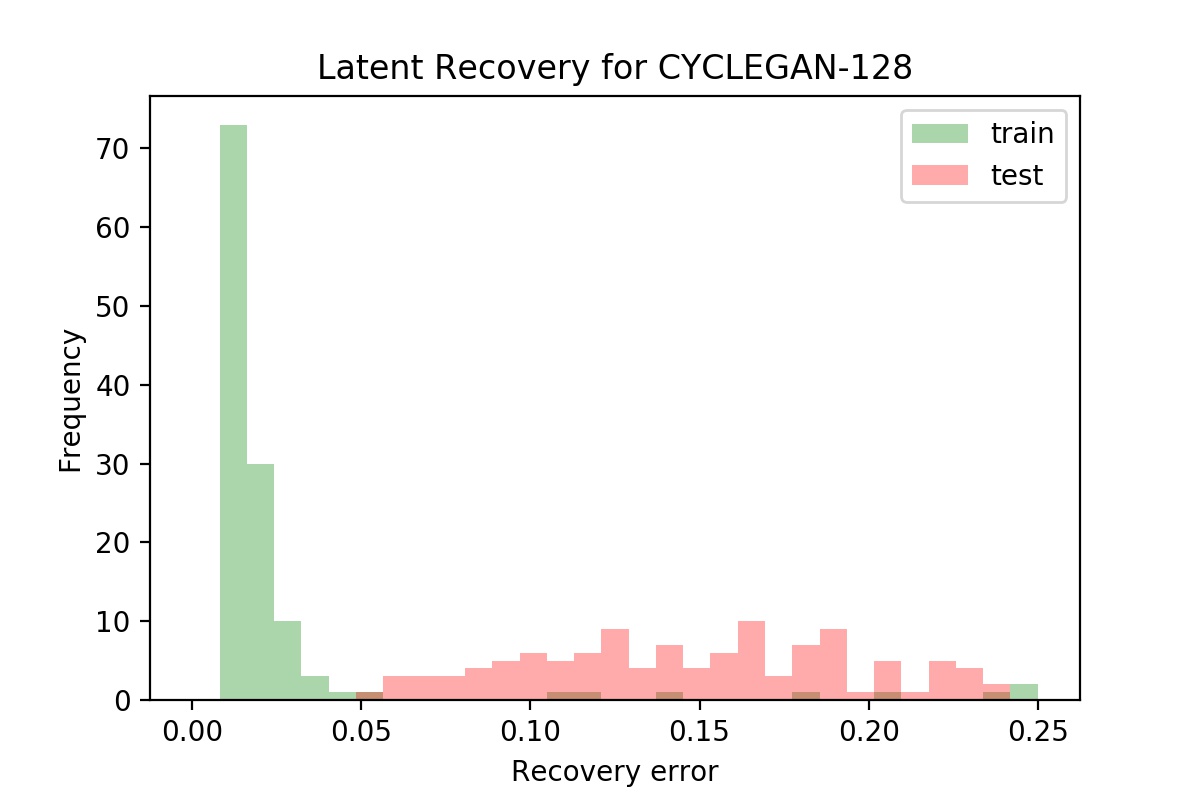}
	\caption{\textsc{aegan}-128}
\end{subfigure}

\centering
\begin{subfigure}{.49\linewidth}
    \centering
    \includegraphics[width=1\linewidth]{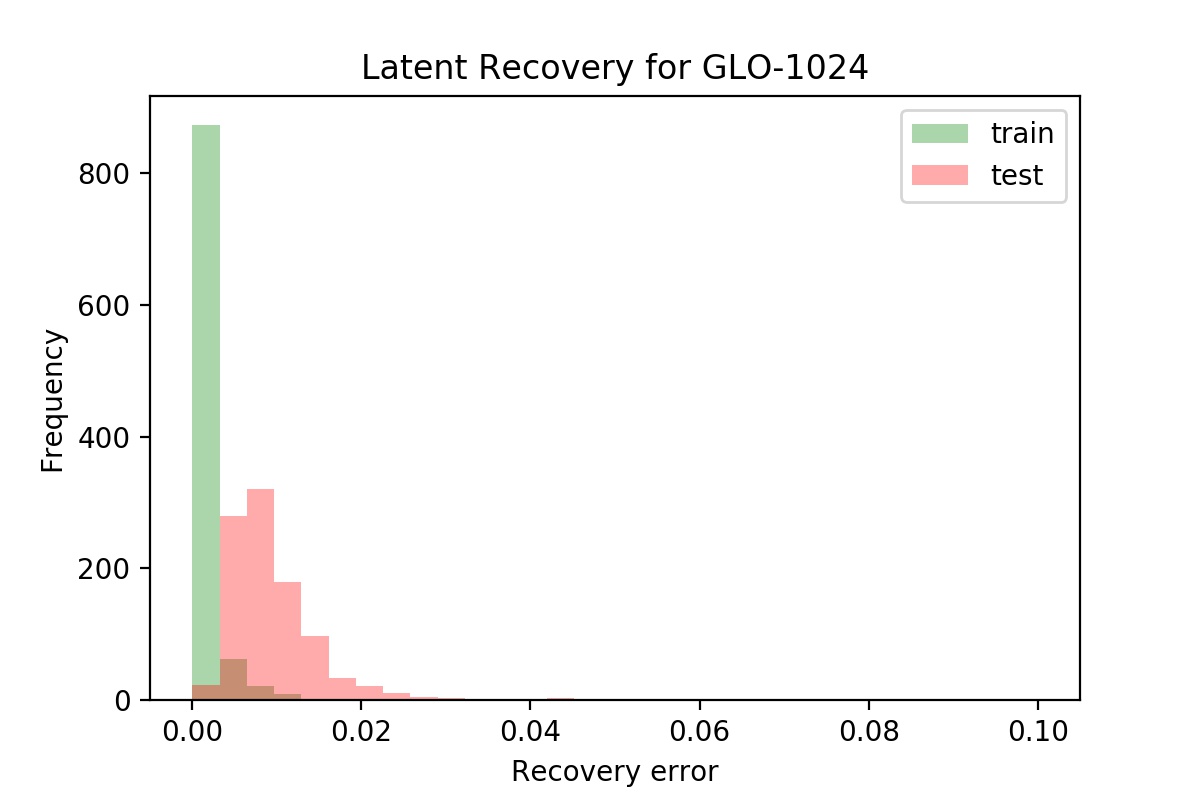}
	\caption{\textsc{glo}-1024}
\end{subfigure}
\begin{subfigure}{.49\linewidth}
    \centering
    \includegraphics[width=1\linewidth]{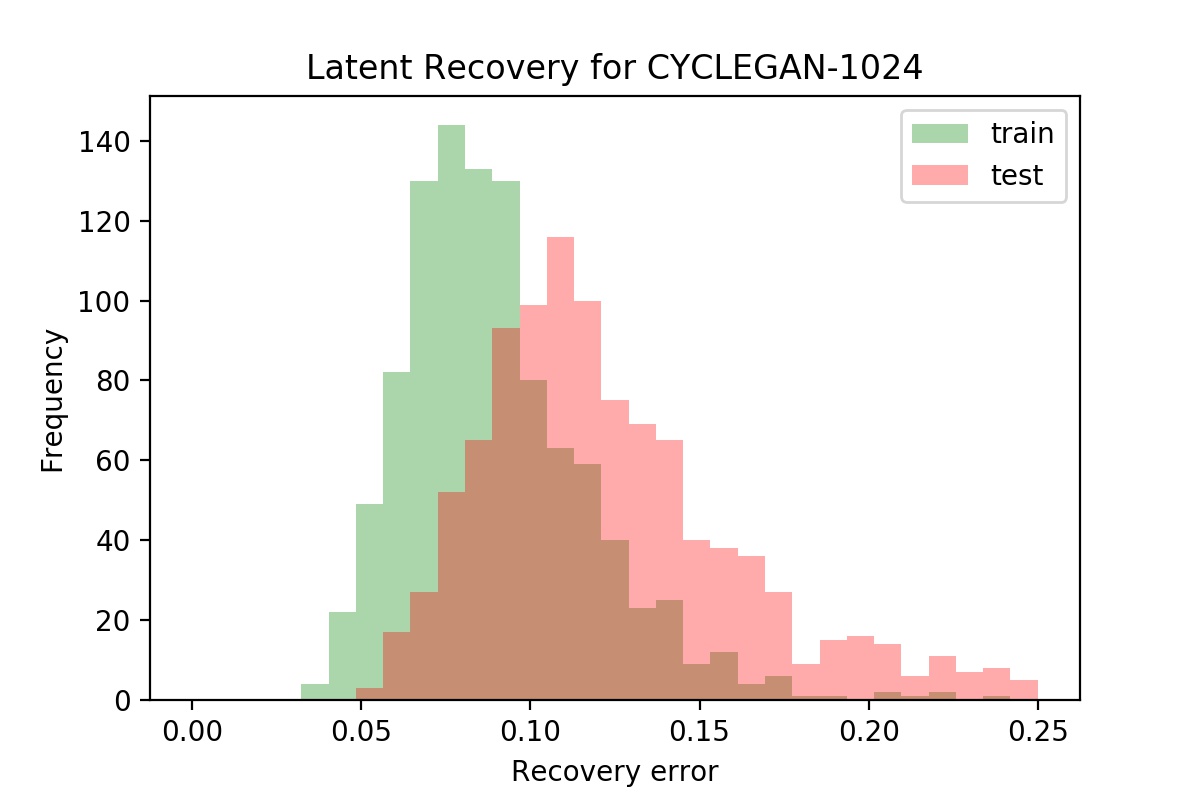}
	\caption{\textsc{aegan}-1024}
\end{subfigure}

\centering
\begin{subfigure}{.46\linewidth}
    \centering
    \includegraphics[width=1\linewidth]{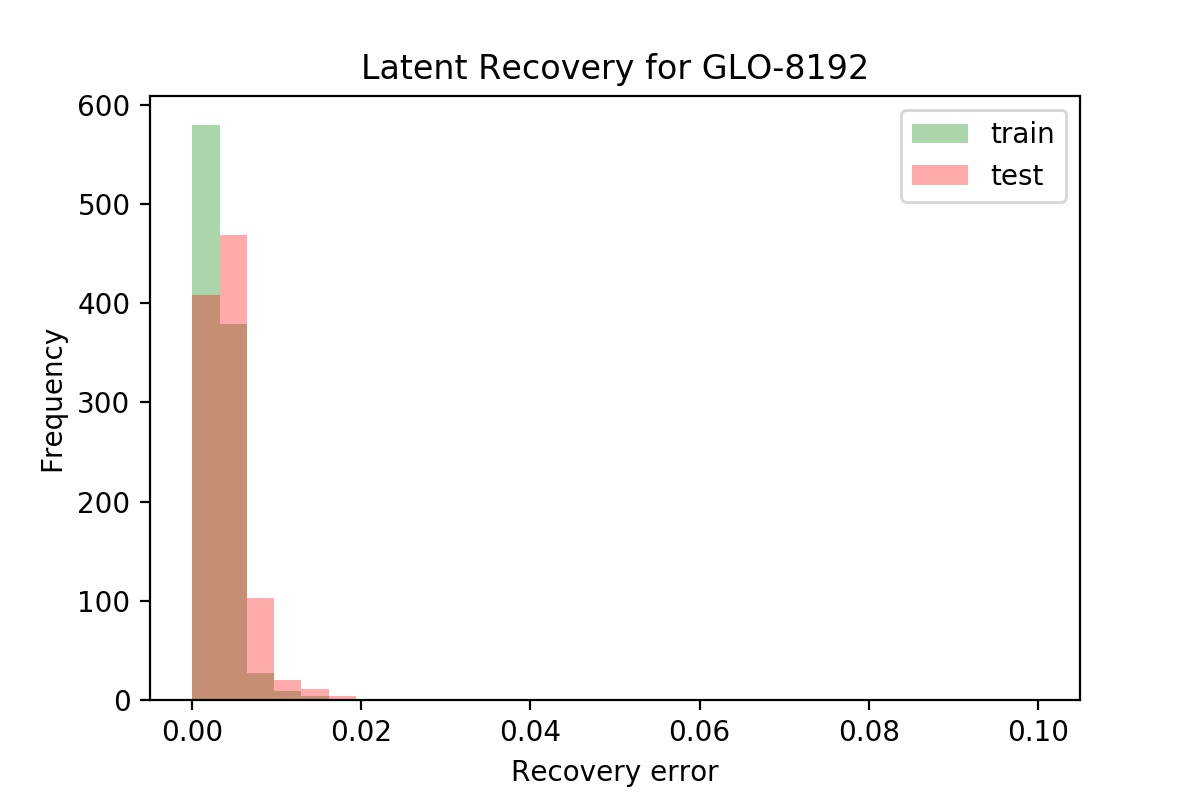}
	\caption{\textsc{glo}-8192}
\end{subfigure}
\begin{subfigure}{.46\linewidth}
    \centering
    \includegraphics[width=1\linewidth]{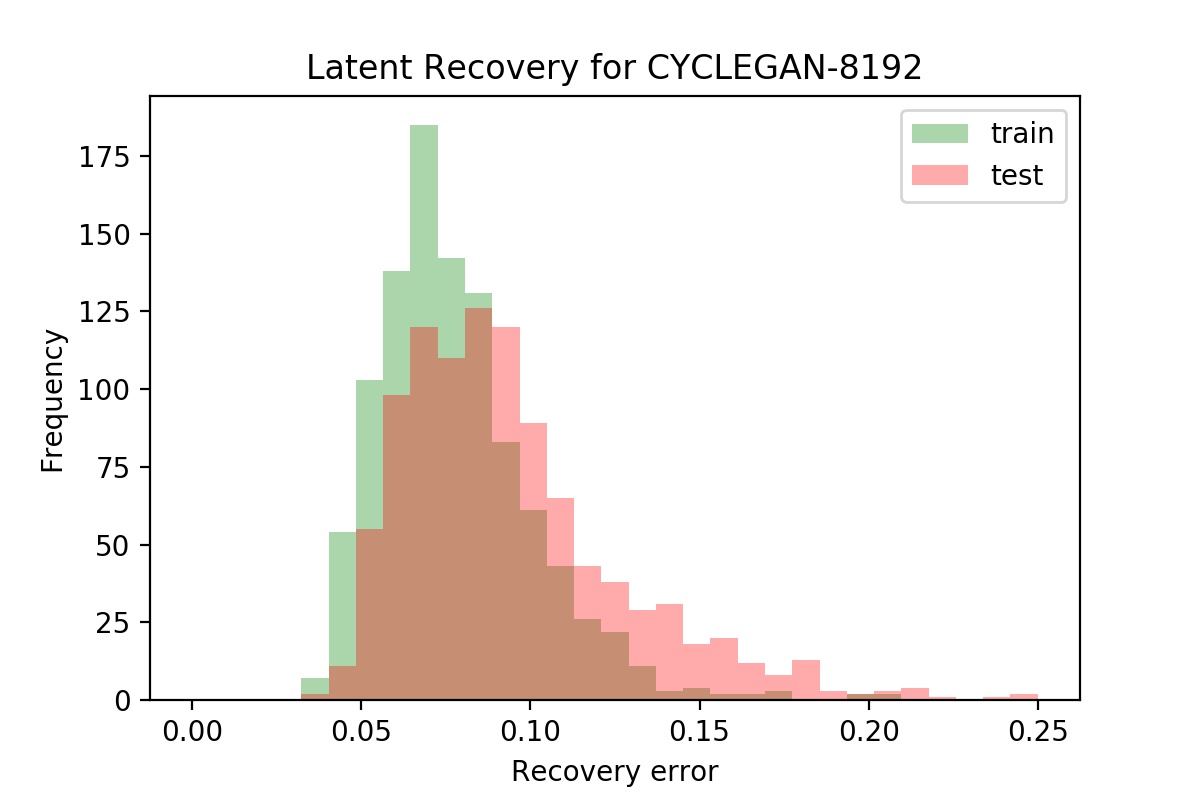}
	\caption{\textsc{aegan}-8192}
\end{subfigure}

\caption{\em Histograms or recovery errors on train $\cal D$ and validation $\cal T$ datasets from CelebA-HQ showing that overfitting is not happening for \textsc{pggan} and \textsc{mecsh} generators on the training dataset, but is for \textsc{glo-N} and \textsc{aegan-N} when training for a small dataset $N<8192$.
%
%
%
%
\vspace*{-3mm}
}
   \label{fig:glo_memorization}
\end{figure}

\subsection{Comparison of recovery errors}\label{sec:comparison_error}

Figure~\ref{fig:glo_memorization} shows the histograms of recovery errors on train ($\cal D$ in green) and validation ($\cal T$ in red) datasets from CelebA-HQ, for various generators.
For the sake of readability, the distribution of errors for  generated images (yellow) from $\cal G$ and distorted images (blue) are only displayed for \textsc{pggan} and \textsc{mesch}.
Confirming visual inspection from Fig.~\ref{fig:teaser}, observe that the recovery errors for generated images (in yellow) are quite low. 
Increasing the number of iterations and using several random initialization improve results, but have not been used to reduce computation costs. 

For both GAN generators \textsc{pggan} and \textsc{mesch} that are trained on the full $26$k dataset $\cal D$, the difference of recovery error distribution between the validation (green) and train (red) set is barely noticeable. 
Further statistical analysis in next paragraph shows indeed that such a small gap is very likely for two samples drawn from the same law and hence no overfitting is present. 

On the other hand, for \textsc{glo-N} and \textsc{aegan-N} generators with $N \in \{128, 1024, 8192\}$, the difference is striking, and is decreasing with $N$ the size of the dataset.
Recall that \textsc{glo-N} is a DCGAN architecture using GLO criterion~\cite{bojanowski2017optimizing} in Eq.~\eqref{eqn: GLO}, and
\textsc{aegan-N} is an auto encoder 
that is trained with a combination of adversarial loss and recovery loss (see Eq.~\eqref{eqn: cyclegan}).
Notice that it was not possible in practice to train pure GAN approaches (\textsc{pggan}, \textsc{dcgan} and \textsc{mesch}) with such small datasets.
On the other hand, generative networks trained with recovery loss create deterministic mappings between latent codes and output images, thus enforcing overfitting. 
However, it is surprising to note that even with a forced deterministic mapping, \textsc{glo} and \textsc{aegan} almost stop memorizing data at only 8192 images. 


In the following paragraph, we design a statistical test to assess automatically when such overfitting is appearing.

\subsection{Statistical Analysis}\label{sec:definition}


In light of the previous results illustrated in Fig.~\ref{fig:glo_memorization}, we propose two simple definitions to measure and detect overfitting without relying on histogram or image inspection. 

First, to resume the distribution of errors to a single value, we consider the \textbf{Median Recovery Error} (MRE), defined for a generator $G$ and a dataset $Y$ as
\begin{equation}\label{def:MRE}
	\text{MRE}_G(\mathcal Y) = \text{median} \left\{ \min_{z} \|y_i -  G(z)\|^2 \right\}_{y_i \in \mathcal Y}
\end{equation}
Table~\ref{table:p_values} reports such values for other deep generators (such as \textsc{AEGAN}) and on other datasets.

Then to measure the distance between two distributions, that is to estimate to which extent the generator overfits the training set, we simply compute the normalized \textbf{MRE-gap} between validation $\cal T$ and train $\cal D$ dataset, which writes 
\begin{equation}\label{def:MRE_gap}
	\hspace*{-3mm}
    \text{MRE-gap}_G = \left(\text{MRE}_G(\mathcal T) - \text{MRE}_G(\mathcal D) \right) 
    / \text{MRE}_G(\mathcal T)
\end{equation}

These values are reported\footnote{Notice that other metrics could have been used, such as the Wasserstein $W_p$ distance, that, however, requires either histogram quantization or that the two discrete distributions are computed on the same number of images.} in Table~\ref{table:p_values}.

Now, instead of using an empirical threshold to automatically assess if the amount of overfitting is significant regarding the size of the training set, we rely on a statistical test. We compute the $p$-value of the Kolmogorov-Smirnov test (KS) which measures the probability that two random samples drawn from the same distribution have a larger discrepancy (defined as the maximum absolute difference between cumulative empirical distributions) than the one observed.
Such $p$-values are displayed in Table~\ref{table:p_values}, and a threshold of $1\%$ is used to detect overfitting (values are highlighted).
To show the consistency between the two proposed metric, we also highlight the values of MRE-gap that are above $10\%$.

\definecolor{myblue}{rgb}{0.5,.8,1}
\definecolor{mygreen}{rgb}{0.5,1,.8}
\newcommand{\ColoR}{\cellcolor{myblue}}
\newcommand{\ColoG}{\cellcolor{mygreen}}
\begin{table*}[!ht] 
\caption{\em Kolmogorov-Smirnov (KS) p-values, normalized median error difference (MRE-gap) Eq.~\eqref{def:MRE_gap}, and Median recovery errors (MRE) Eq.~\eqref{def:MRE} for a variety of generators. 
Highlighted values indicate generators for which overfitting of the training set has been detected: (in blue) below $1\%$ threshold for p-value of the KS test, (in green) above $10\%$ threshold for MRE-gap.
}
\label{table:p_values}
\centering
\renewcommand{\arraystretch}{1.0}
\vspace*{-2mm}
\iftrue 
\begin{tabular}{|c|l| |c|c| |c|c|c|c|}
\cline{3-8}
\multicolumn{2}{c|}{\multirow{2}{*}{}}  		& KS p-value & MRE-gap & \multicolumn{4}{c|}{MRE}  \\ 
\cline{3-8}
\multicolumn{2}{c|}{}  		& \multicolumn{2}{c||}{train vs val} & train&val &generated& small distort  \\ 
\cline{3-8}
\hline
\multirow{14}{*}{CelebA-HQ} 
& \textsc{dcgan} &   9.43e-01 &  1.79e-02 &  4.95e-02 &  5.04e-02 &  3.68e-03 &  5.69e-02 \\ \cline{2-8} 
&\textsc{mesch} &  4.55e-01 &  6.96e-03 &  3.40e-02 &  3.43e-02 &  1.77e-02 &  4.63e-02 \\ \cline{2-8} 
&\textsc{pggan} &  2.22e-01 &  2.22e-02 &  3.31e-02 &  3.39e-02 &  1.78e-02 &  4.65e-02 \\ \cline{2-8} 
&\textsc{glo}-128 &  \ColoR \bf 0.00e+00 & \ColoG 9.70e-01 &  9.94e-04 &  3.30e-02 &  5.10e-05 &  9.32e-03 \\ \cline{2-8} 
&\textsc{glo}-1024 & \ColoR \bf 0.00e+00 & \ColoG 7.59e-01  &  1.95e-03 &  8.08e-03 &  1.29e-03 &  4.46e-03 \\ \cline{2-8} 
&\textsc{glo}-8192 & \ColoR \bf 2.25e-18 & \ColoG 1.75e-01  &  3.00e-03 &  3.64e-03 &  1.04e-03 &  3.20e-03 \\ \cline{2-8}
&\textsc{glo}-26000 &  2.12e-01 &   3.69e-02 &  4.27e-03 &  4.44e-03 &  4.08e-04 &  4.43e-03 \\ \cline{2-8} 
&\textsc{aegan}-128 & \ColoR \bf 0.00e+00 &\ColoG 9.02e-01 &  1.54e-02 &  1.57e-01 &  N/A &  2.82e-02 \\ \cline{2-8} 
&\textsc{aegan}-1024 & \ColoR \bf 0.00e+00 & \ColoG 2.68e-01&  8.52e-02 &  1.16e-01 &  N/A &  8.69e-02 \\ \cline{2-8} 
&\textsc{aegan}-8192 & \ColoR \bf 3.17e-27 & \ColoG 1.61e-01 &  7.42e-02 &  8.84e-02 &  N/A &  7.55e-02 \\ \cline{2-8} 
&\textsc{aegan}-26000 &   1.25e-01 &   1.85e-02 &  9.96e-02 &  1.01e-01 &  N/A &  1.00e-01 \\ \cline{2-8} 
&\textsc{cyclegan}-256 M2F &  \ColoR \bf 0.00e+00 & \ColoG  4.75e-01 &  9.03e-03 &   1.72e-02 &  N/A & - \\ \cline{2-8} 
&\textsc{cyclegan}-4096 M2F & \ColoR \bf 0.00e+00  &  \ColoG  2.62e-01 &   6.44e-03 &   8.73e-03 &  N/A & - \\ \cline{2-8} 
\hline
\multirow{4}{*}{LSUN} 
&\textsc{dcgan} (tower) &  7.02e-02 &  1.36e-02 &  7.96e-02 &  8.07e-02 &  1.49e-02 &  7.31e-02 \\ \cline{2-8} 
&\textsc{dcgan} (bedroom) & 3.65e-01 & 5.34e-03&  7.06e-02 &  7.10e-02 &  7.03e-02 &  7.09e-02 \\ \cline{2-8} 
&\textsc{glo}-8192 (bedroom) & \ColoR \bf 6.70e-06 & \ColoG 1.70e-01 &  5.45e-03 &  6.56e-03 &  5.37e-04 &  5.01e-03 \\ \cline{2-8} 
&\textsc{glo}-32768 (bedroom) & 2.62e-01 &  5.40e-02 &  6.58e-03 &  6.25e-03 &  8.40e-04 &  5.44e-03 \\ \hline
\hline
\multirow{2}{*}{Yosemite} 
&\textsc{cyclegan}-256 s2w & \ColoR \bf 1.60e-16 &  \ColoG 3.68e-01 &   1.67e-02 & 2.64e-02 &  N/A &  - \\ \cline{2-8}
&\textsc{cycelgan}-512 s2w & \ColoR \bf 6.10e-33 & \ColoG  3.78e-01 &  1.39e-02&   2.23e-02 & N/A & -\\ \hline
\hline
\multirow{4}{*}{MNIST} 
& \textsc{dcgan} &   2.41e-01 &  8.85e-02 &  3.00e-02 &  2.75e-02 &  6.89e-03 & -  \\ \cline{2-8} 
&\textsc{glo}-1024 & \ColoR \bf 0.00e+00 & \ColoG 6.78e-01 &  2.86e-04 &  8.88e-04 &  1.49e-03& -  \\ \cline{2-8} 
&\textsc{glo}-16384 &  3.48e-01 &  6.45e-03 &  8.72e-04 &  8.77e-04 &  1.41e-03& -  \\ \cline{2-8} 
&\textsc{aegan}-16384 &  \ColoR \bf  7.43e-02 &  2.29e-02 &  4.56e-02 &  4.67e-02 &  N/A & - \\ \hline

\hline
\multirow{2}{*}{CIFAR10} 
&\textsc{dcgan} &  5.40e-01 &  3.65e-03 &  2.29e-01 &  2.28e-01 &  1.30e-03 &- \\ \cline{2-8} 
&\textsc{glo}-1024 & \ColoR \bf 0.00e+00 & \ColoG 5.84e-01 &  2.77e-03 &  6.67e-03 &  8.53e-04&-  \\ \cline{2-8} 
&\textsc{glo}-16384 & 3.48e-01 &  6.45e-03 &  8.72e-04 &  8.77e-04 &  1.41e-03&-  \\ \hline

\end{tabular}
\fi
\end{table*}

Observe that the results are mostly confirming previous empirical evidences: memorization is strongly correlated to the number of images seen during training.
At $N=26000$ on CelebA-HQ and $N=32768$ on LSUN bedroom, overfitting is no longer detectable.

However, using the proposed statistics (p-value, normalized MRE-gap) is much more practical to detect overfitting than only inspecting histograms and easier to threshold than MRE itself.
It also illustrates that such statistical principle overrules empirical evaluation, 
as memorization is indeed sometimes quite hard to tell from simple visual inspection, such as for the \textsc{aegan} generator in Fig.~\ref{fig:recovery}.

\subsection{FID Does Not Detect Memorization}

\begin{figure}[!htb]
	\centering
	\includegraphics[width=1\linewidth]{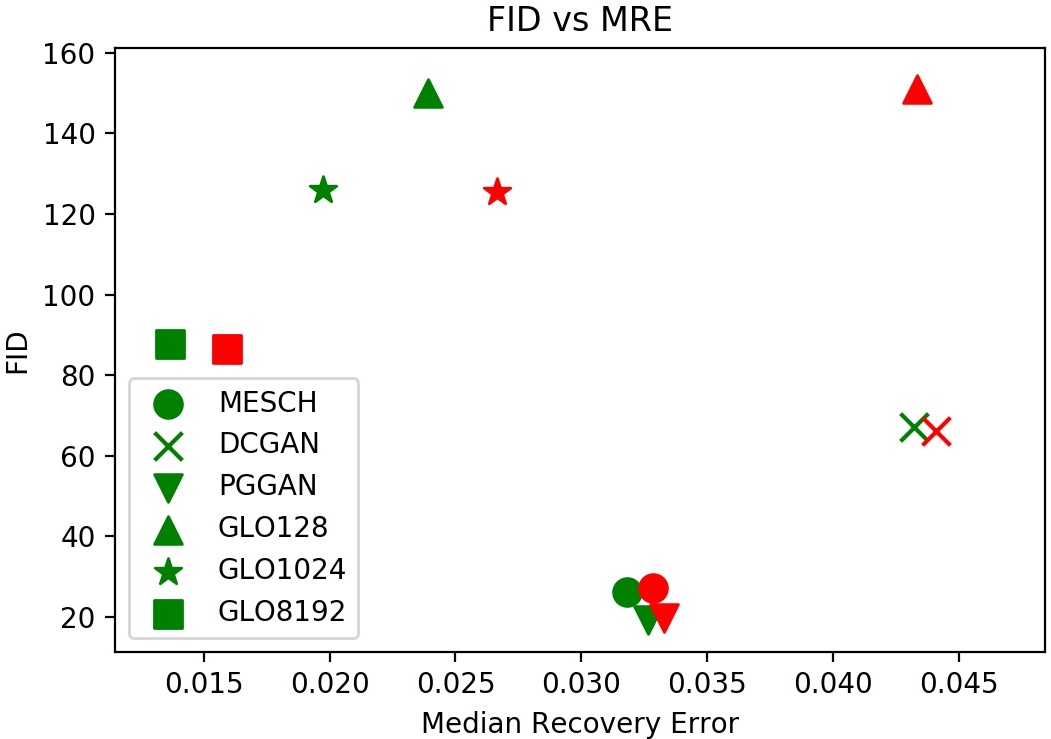}

\caption{\em Comparison of FID versus Median Recovery Error (MRE) for various models computed over training images (in green) and validation images (in red). FID does not detect memorization in GLO models.
\vspace*{-3mm}}
   \label{fig:fid_vs_mre}
\end{figure}

The FID is the standard GAN evaluation metric for images \cite{heusel2017gans}, so it is natural to ask whether this metric can be used to detect memorization. Figure \ref{fig:fid_vs_mre} displays FID scores computed between generated and training images (in green) and generated and test images (in red). While the median recovery error (MRE) is able to detect memorization in GLO models, the FID is not sensitive to this. We note that in the work  of \cite{sajjadi2018assessing}, they introduce notions of precision and recall for generated images. In a similar way, they note that the FID doesn't distinguish models which have dropped many modes to those which have poor average quality. Finally, we do not suggest replacing the FID, but rather using MRE to provide a more complete picture of generator performance.

\begin{figure*}[!h]
\vspace*{-4mm}
\begin{subfigure}{.05\linewidth}
  \begin{tabular}{c}
  {\rotatebox[origin=t]{90}{ \hspace{2mm} }} 
  \\
  {\rotatebox[origin=t]{90}{$\phi$ = mask}} 
  \\
  {\rotatebox[origin=t]{90}{ \hspace{35mm} }} 
  \\
  {\rotatebox[origin=t]{90}{$\phi$ = pooling}} 
  \\
  {\rotatebox[origin=t]{90}{ \hspace{1mm} }} 
  \\
  \end{tabular}
\end{subfigure}
\begin{subfigure}{.9\linewidth}
	\centering
	\begin{tabular*}{0.87\linewidth}{
    @{\extracolsep{\fill}}c@{\extracolsep{\fill}}
    @{\extracolsep{\fill}}c@{\extracolsep{\fill}}
    @{\extracolsep{\fill}}c@{\extracolsep{\fill}}
    @{\extracolsep{\fill}}c@{\extracolsep{\fill}}
    @{\extracolsep{\fill}}c@{\extracolsep{\fill}}
    @{\extracolsep{\fill}}c@{\extracolsep{\fill}}
    }
          $\phi(y)$ & $G(z^*(y))$ & $y$  &
          $\phi(y)$ & $G(z^*(y))$ & $y$  
    \end{tabular*}
        
    \centering
    \includegraphics[width=\linewidth]{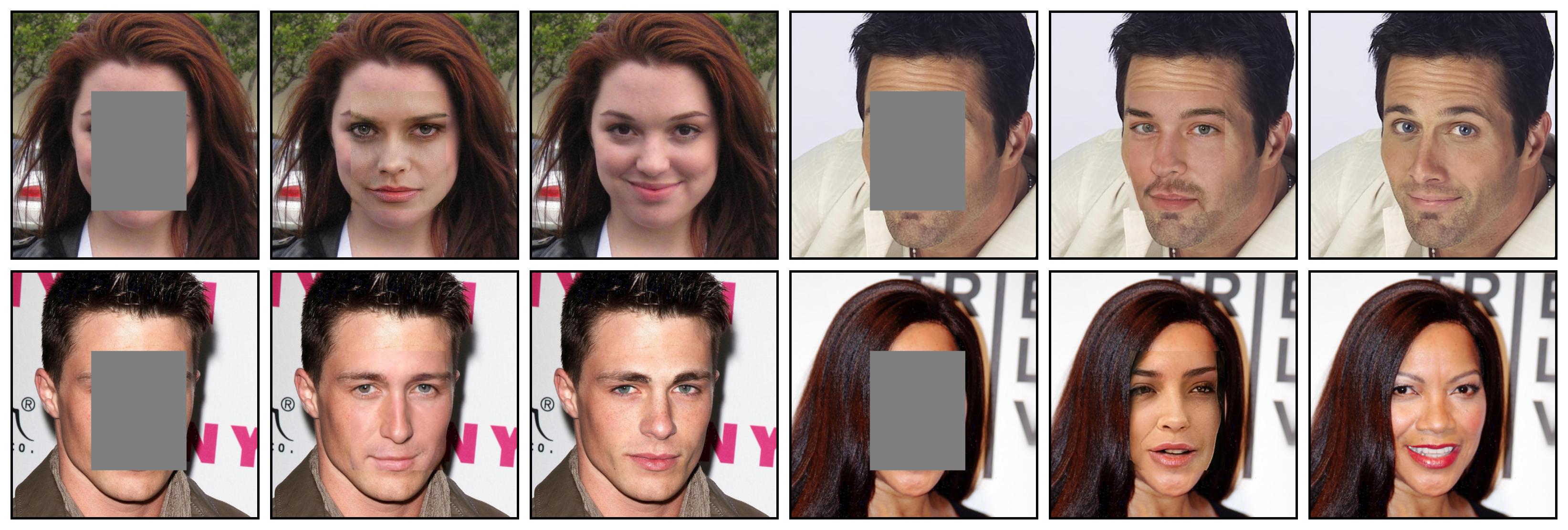}
    
    \centering
    \includegraphics[width=\linewidth]{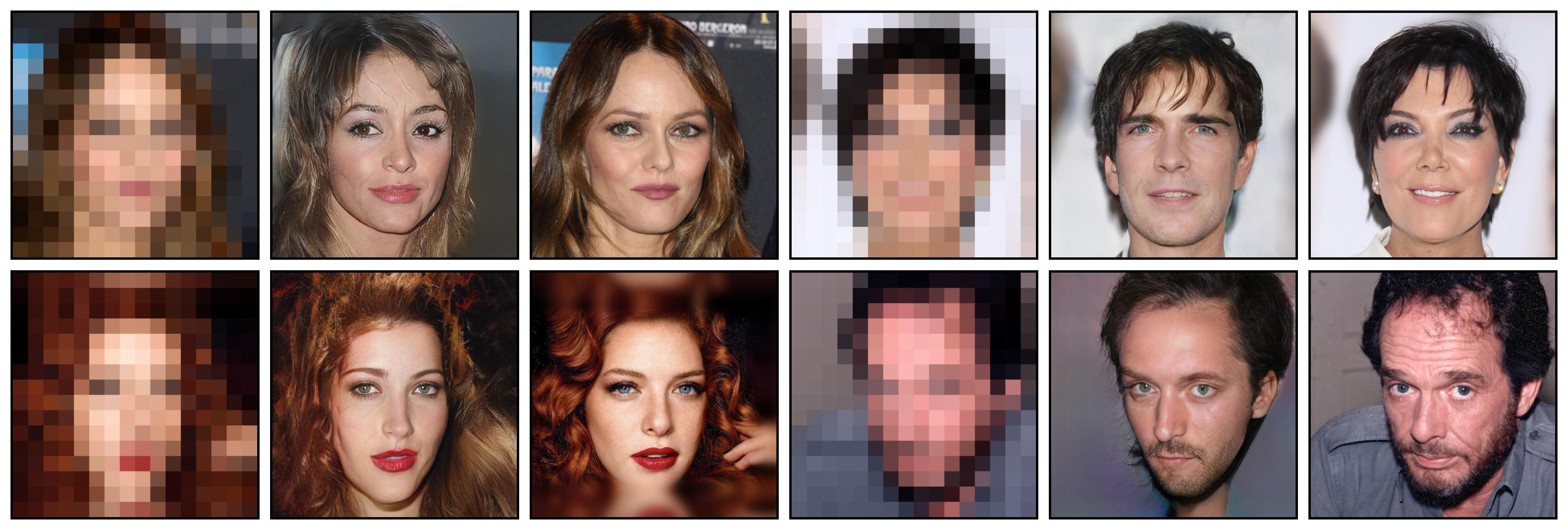}
\end{subfigure}
  \caption{\em Image recovery by solving Eq.~\eqref{eq:recovery_L2} with a 1024\emph{x}1024 generator $G$ by~\cite{karras2017progressive}. 
  From left to right: transformed image $\phi(y)$, recovered image $G(z^*(y))$ and ground truth image $y$. 
  The first two rows are image inpainting ($\phi$ is a mask) and next two are image super-resolution of images downsampled by a factor of 64 ($\phi$ is an average pooling).
\vspace*{-2mm}}
   \label{fig:inpainting}
\end{figure*}

\section{Applications to image editing}\label{sec:inpainting}

Recently, GANs have seen wide application to various face generation tasks, such as face attribute modification~\cite{choi2018stargan}, generative face completion~\cite{li2017generative} and face super-resolution~\cite{dahl2017pixel}.
Another approach orthogonal to ours can be found in \cite{ulyanov2018deep}, where recovery is performed by first fixing a random latent code, and optimizing over the parameters of a randomly initialized generator. Although their approach is very flexible and produces compelling results, it is by design bound to fail at inpainting large areas of semantic structures such as faces.

As a proof of concept, we show that a vanilla GAN not trained for any specific task can solve inverse problem decently well using latent recovery optimization~\eqref{eq:recovery_L2} with various operators $\phi$. 
Fig.~\ref{fig:recovery distortion} already displays some potential applications. For example, because deep generators are specific in the images they generate, they can be used to un-distort (Fig.~\ref{sub:warp}) images or for blind inpainting (Fig.~\ref{sub:hole}). 

Fig.~\ref{fig:inpainting} shows the progressive GAN generator \cite{karras2017progressive} applied to face inpainting ($\phi$ is a mask) and super-resolution ($\phi$ is a 64x pooling). 
While the face inpainting is artifacted, we note that the results are decent without any post processing and similar to those presented in~\cite{li2017generative} (while being non-feedforward). 
As for super-resolution, we obtain results at least on par with~\cite{dahl2017pixel}.
It is also interesting to note that while the pose of the face remains accurate for both face inpainting and super-resolution, the identity of the face changes. 
This happens despite the use of images that the PGGAN generator \cite{karras2017progressive} was trained on, which is in accordance with the observations of Sec.~\ref{sec:memorization}. 
In fact, this is a positive feature of the algorithm, as hybrid losses in deep inpainting methods, such as~\cite{iizuka2017globally}, could potentially overfit faces.

\section{Discussion and future work}\label{sec:discussion}

We demonstrated that using a recovery loss, like in GLO, makes it possible to overfit the training set, even when combined with an adversarial training.
However, it is noteworthy that GLO and CycleGAN can train with few data points, while still having a decent generation quality. 

We believe overfitting in the sense of Section~\ref{sec:memorization} does not tell the whole story. 
For instance, when the network in \cite{mescheder2018training} is trained on LSUN tower, images of the Eiffel Tower can be clearly recognized. 
While for faces, a GAN might perfectly generalize w.r.t. individual faces, on other datasets information about specific objects in the dataset might leak. 
In this way, a closer examination of privacy for generative methods is needed.

Finally, we note that the optimization technique in Eq.~\eqref{eq:recovery_L2} failed for a few generators trained on LSUN bedroom, in the sense that it could not verbatim recover generated images. This suggests the dataset itself may have some effect on the organization of the latent space and that the LSUN latent space is more complex with many local minima of Eq.~\eqref{eq:recovery_L2}. On the other hand, other networks were able to successfully verbatim recover generated images for LSUN tower and bedroom. We argue this phenomenon needs more investigation.

\paragraph{Conclusion}
In this work, we elucidated important properties of deep generators through latent recovery. 
We saw that a simple Euclidean loss was very effective at recovering latent codes and recovers plausible images even after significant image transformations. 
We used this fact to study whether a variety of deep generators memorize training examples. 
Our statistical analysis revealed that overfitting was undetectable for GANs, but detectable for hybrid adversarial methods like CycleGAN 
and non-adversarial methods like GLO, 
even for moderate training set sizes. Due to the ever-growing concerns on privacy and copyright of training data and the widespread application of hybrid adversarial losses, we provide methodology that is a step in the right direction towards analysis of overfitting in deep generators.

\paragraph{Acknowledgments}
This work was supported by Region Normandie, under grant \emph{RIN NormanD'eep}.

{
\bibliographystyle{ieee} 
\bibliography{bib}
}

\onecolumn
\appendix
\section{Apendix}

\subsection{Optimization Failures}\label{sec:optim_fail}
\begin{figure}[htb]
    \centering
    \begin{subfigure}{.45\linewidth}
      	\includegraphics[width=1\linewidth]{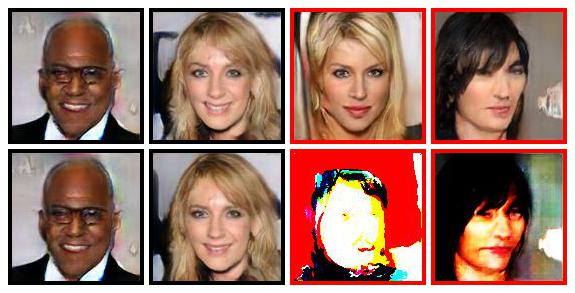}
        \caption{\em Generated recovery for MESCH on CelebA-HQ.}\label{sub:celeba_gen}
    \end{subfigure}
    \centering
    \begin{subfigure}{.45\linewidth}
  		\includegraphics[width=1\linewidth]{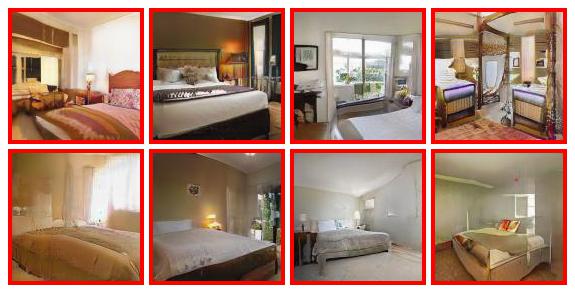}
        \caption{\em Generated recovery for PGGAN on LSUN Bedroom. }\label{sub:lsun_bed_gen}
    \end{subfigure}  
    \caption{Recovery failure detection with thresholding. First row generated images and second row is recoveries. The MESCH network in ~\ref{sub:celeba_gen} is inconsistent at image recovery, which can be alleviated by restarts. The PGGAN trained on LSUN bedroom in ~\ref{sub:lsun_bed_gen} did not verbatim recover any image. }
\end{figure}

We noted that most networks had the ability to exactly recover generated images. This is shown in Fig.~\ref{fig:gen_recov}, with failure cases highlighted in red. Interestingly, some networks were not able to recover their generated images at all, for example Fig.~\ref{sub:lsun_bed_gen} was a PGGAN trained on LSUN Bedroom, which did not verbatim recover any image. We think this may suggest a more complex latent space for some networks trained on LSUN, with many local minima to Eq.~\eqref{eq:recovery_L2}. Because we assert that we are finding the nearest neighbors in the space of generated images, we did not analyze networks which could never recover generated images. It should be noted that some LSUN networks were able to recover their generated images, for instance DCGAN \cite{radford2015unsupervised}.

\label{fig:gen_recov}

\subsection{Recovery Success Rate}
Disregarding networks which could not recover generated images, some networks had higher failure rates than others. To determine failure cases numerically, we chose a recovery error threshold of $\text{MSE}<.1$ to signify a plausible recovery for real images (for generated images a much smaller threshold of $\text{MSE}<.025$ can be used, which corresponds to verbatim recovery, e.g. in ~\ref{fig:gen_recov}). Table~\ref{tab:success_rate} summarizes recovery rates for a few networks. The MESCH resnets were notably less consistent than other architectures. To study if these failures were due to bad initialization, we tried simply restarting optimization 10 times per image, and saw the success rate go from 68\% to 98\% (shown in Table~\ref{tab:success_rate} as MESCH-10-RESTART). This shows that likely all training and generated images can be recovered decently well with enough restarts. 
\begin{table}[htb]
    \centering
    \caption{\em Recovery success rate for real and generated images. Percentages indicate rate of recoveries with a $\text{MSE}<.1$, which corresponds to a plausible synthesis. Failures are related to bad initialization, as simply restarting greatly increases success rate.}
    \begin{tabular}{ |c|c|c|c| }
    \hline
     & train & test & generated  \\ \hline
    MESCH &  68\% &  67\% &  67\% \\ \hline 
    MESCH-10-RESTART &  98\% &  99\% &  96\%  \\ \hline
    DC-CONV &  82\% &  82\%  &  100\%  \\ \hline 
    PGGAN & 97\% &  96\% &  95\%   \\ \hline 
    \end{tabular}
    \label{tab:success_rate}
\end{table}

\subsection{Local vs Global overfitting}\label{sec:local}

\begin{figure}[ht]
    \centering
    \includegraphics[width=\textwidth,height=\textheight,keepaspectratio]{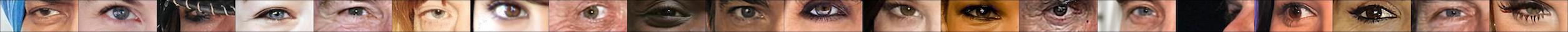}
    \small Target images (after cropping eye on training images)
    
    \centering
    \includegraphics[width=\textwidth,height=\textheight,keepaspectratio]{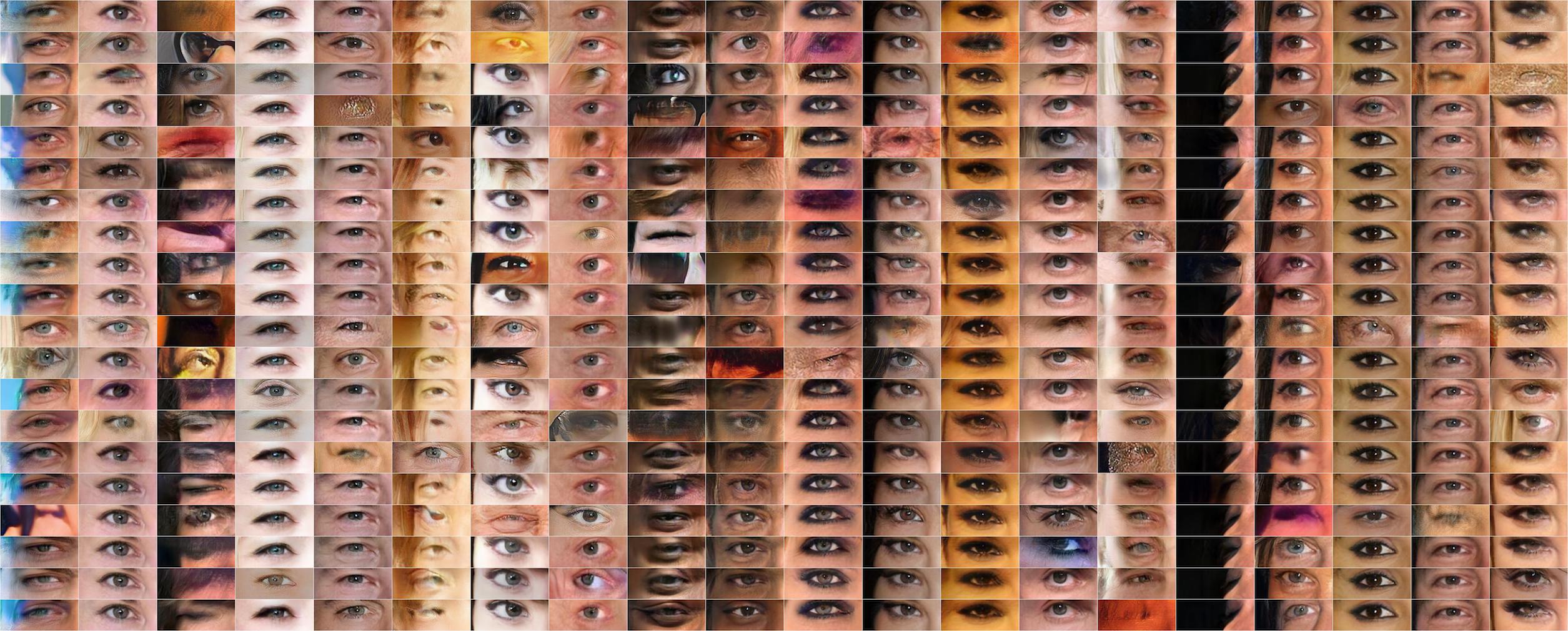}
    
    \small Recovered images (for different initialization)
    
    \caption{\em Visual results on training images recovery with a  \textbf{masking operator} on right eyes (using LBFGS and Euclidean objective loss, and PGGAN generator).
    First row: target (real) images $y_i$.
    First column: initialization. 
    Second column: optimization after 100 iterations. 
    Recovery is more consistent than global optimization.
    }
    \label{fig:LBFGS_visual_real_mouth}
\end{figure}

\begin{figure}[htb]
    \centering
    \begin{subfigure}{.35\linewidth}
      	\includegraphics[width=1\linewidth]{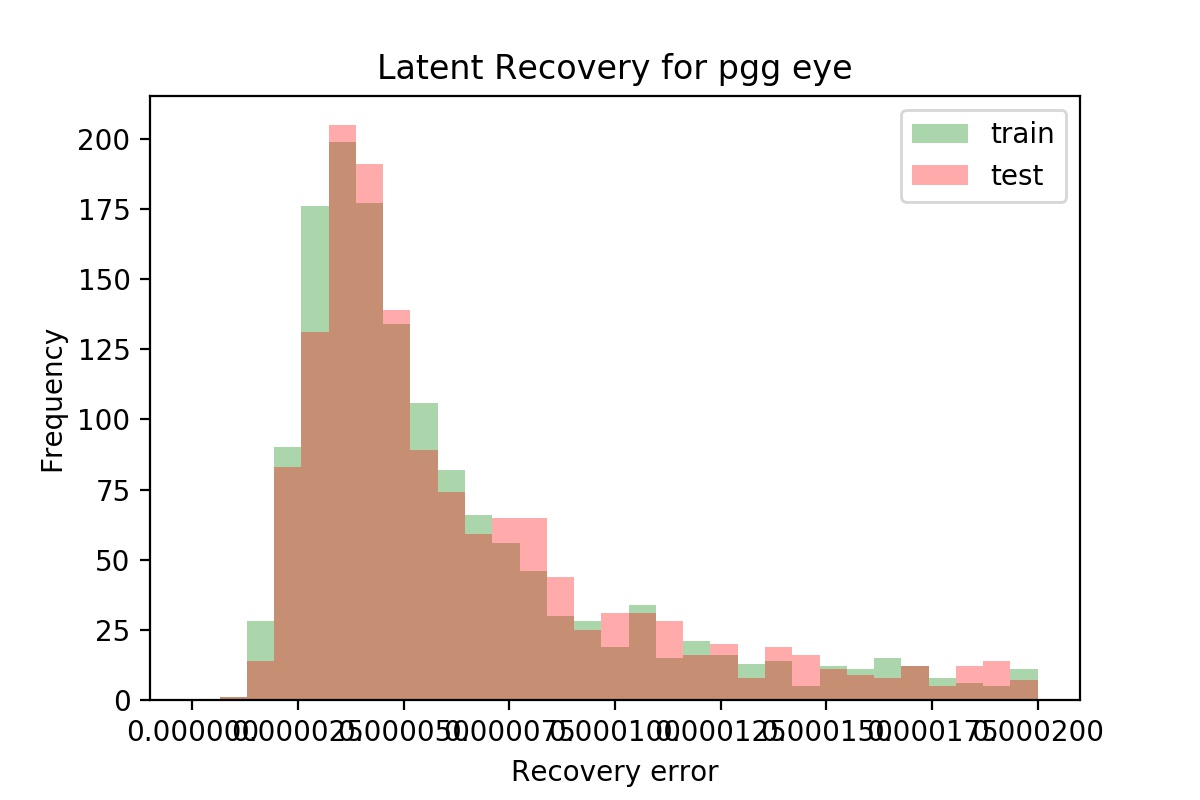}
        \caption{\em Eye patch recovery error.}\label{sub:eye patch}
    \end{subfigure}
    \centering
    \begin{subfigure}{.35\linewidth}
      	\includegraphics[width=1\linewidth]{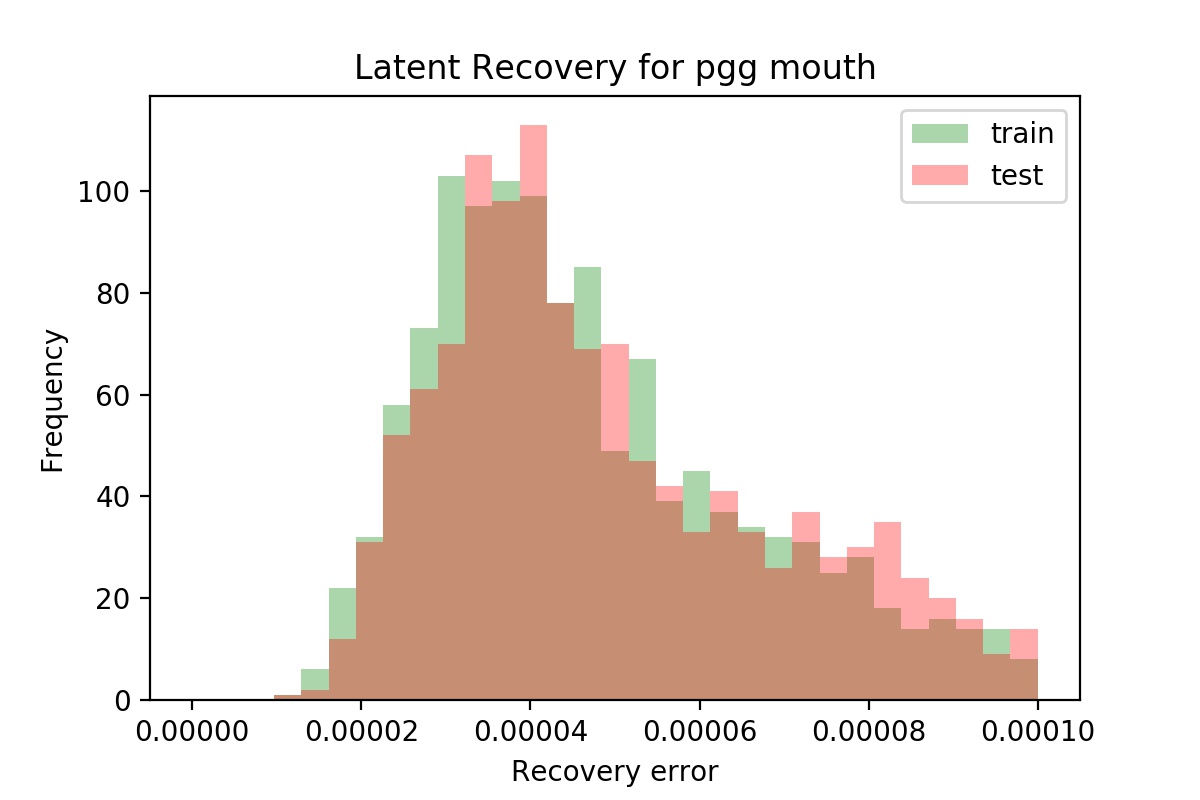}
        \caption{\em Mouth patch recovery error.}\label{sub:mouth patch}
    \end{subfigure} 
    \caption{\em Recover of eye patches (left) and mouth patches (right). The KS p-values for these two graphs are $p = 0.0545$ and $p=0.6918$ respectively from left to right. } 
    \label{fig:patch recov}
\end{figure}

While GANs generators appear to not overfit the training set on the entire image, one may wonder if they do however overfit training image patches. 
To investigate this, we take $\phi$ of Eq.~\eqref{eq:recovery_L2} to be a masking operator on eye or mouth regions of the image. 
To first verify this optimization is stable (see Section~\ref{sec:conv}, for more information of stability of optimization), we recover eyes \textsc{pggan} for a number of random initializations in Fig.~\ref{fig:LBFGS_visual_real_mouth}. Recovery is consistent for most initializations, which is also the case for global image recovery (see Sec.~\ref{sec:conv}). 
Finally, we observe the recovery histograms and KS p-values for patches in Fig.~\ref{fig:patch recov}.
As with global recovery, no significant local overfitting is detected for eye or mouth areas: the p-values for the Kolmogorov-Smirnov test (used to measure the similarity of the training and validation recovery error distributions) are $p = 0.0545$ and $p=0.6918$ respectively, above the $1\%$ threshold we used for global overfitting detection.

\clearpage
\section{Comparison with Other Loss Functions}\label{sec:loss_visual}
\begin{figure*}[!htb]
  \centering
  \begin{subfigure}{.05\linewidth}
  \begin{tabular}{c}
  {\rotatebox[origin=t]{90}{Targets}} 
  \\[14mm]
  {\rotatebox[origin=t]{90}{$L_2$}} 
  \\[14mm]
  {\rotatebox[origin=t]{90}{pool}} 
  \\[14mm]
  {\rotatebox[origin=t]{90}{VGG-19}} 
  \\[14mm]
  {\rotatebox[origin=t]{90}{$L_2$}} 
  \\[14mm]
  {\rotatebox[origin=t]{90}{ pool}} 
  \\[14mm]
  {\rotatebox[origin=t]{90}{VGG-19}} 
  \end{tabular}
  \end{subfigure}
  \begin{subfigure}{.85\linewidth}
  		\includegraphics[width=1\linewidth]{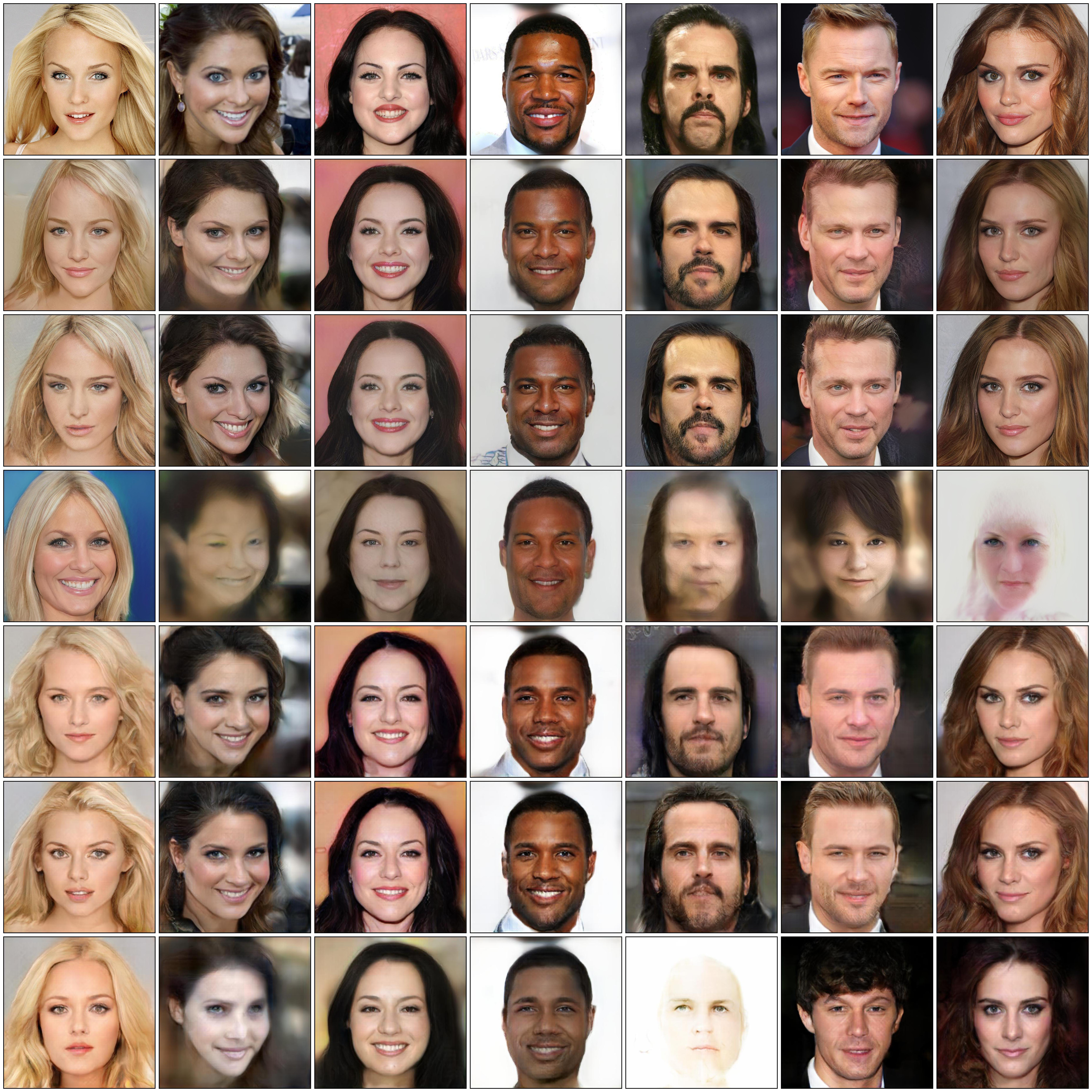}
  \end{subfigure}      
        
  \caption{\em Using other loss functions for image recovery. The first row is target images from CelebA-HQ, the next three rows are recovery from PGGAN network and the final three are from MESCH generator. Pooling seems to help slightly with textural details without hindering recovery of facial pose. Surprisingly, a VGG-19 loss hinders recovery for the PGGAN generator.}
   \label{fig:other_losses}
\end{figure*}

We visually compare in Fig.~\ref{fig:other_losses} 
the simple Euclidean loss used in this paper for analyzing overfitting (\emph{i.e.} $\phi = \text{Id}$ in Eq.~\eqref{eq:recovery_L2}) with other operators: 
\begin{itemize}
    \item $\phi = $ pooling by a factor of 32 (as used in applications for super-resolution);
    \item $\phi = $ various convolutional layers of the VGG-19 (\emph{i.e.} the \emph{perceptual loss} previously mentioned in the paper).
\end{itemize}
While the perceptual loss has been shown to be effective for many synthesis tasks, it appears to hinder optimization in the case when interacting with a high quality generator $G$. Observing the recovered images in Fig.~\ref{fig:other_losses}, the pooling operator seems to help with recovered textures as it relaxes the loss, while still recovering a highly similar face to the naive loss $\phi = \text{Id}$.

\clearpage
\section{Convergence analysis of latent recovery}\label{sec:conv}
In general, optimization was successful and converges nicely for most random initializations. We provide numerical and visual evidence in this section supporting fast and consistent convergence of LBFGS compared to other optimization techniques like SGD or Adam.

\subsection{Protocol}

To demonstrate that the proposed optimization of the latent recovery is stable enough to detect overfitting, the same protocol is repeated in the following experiments. We used the same 20 random latent codes $z_i^*$ to generate images as target for recovery: $y_i = G(z_i*)$ .
We also used 20 real images as targets 
the same as in Section 2 for local recovery.
We also initialized the various optimization algorithms with the same 20 random latent codes $z_i$.
We plot the median recovery error (MRE) for 100 iterations.
This curve (in red) is the median of all MSE curves (whatever the objective function is) and is compared to the 25th and 75th percentile (in blue) of those 400 curves.

\subsection{Comparison of optimization algorithm}

\begin{figure}[htb]
    \centering
    \begin{tabular}{cc}
        & Target images (generated with PGGAN)
        \\
        
        & \includegraphics[width=.8\linewidth]{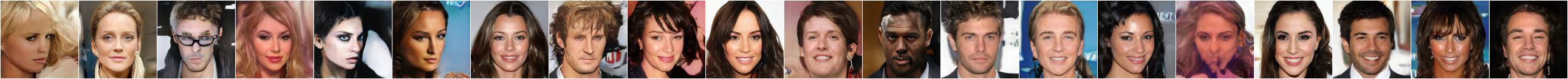}
        \\
        \raisebox{.2\linewidth}{\rotatebox[origin=c]{90}{Initialization (LBFGS)}}  
        
        \includegraphics[height=.395\linewidth,trim={0 0 0 46cm},clip]{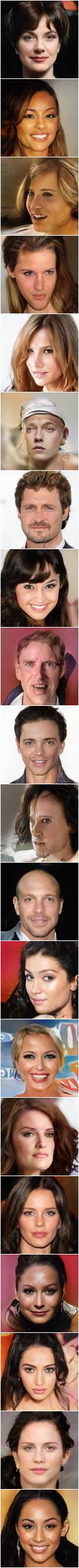}
        &
        \includegraphics[width=.8\linewidth,trim={0 0 0 46cm},clip]{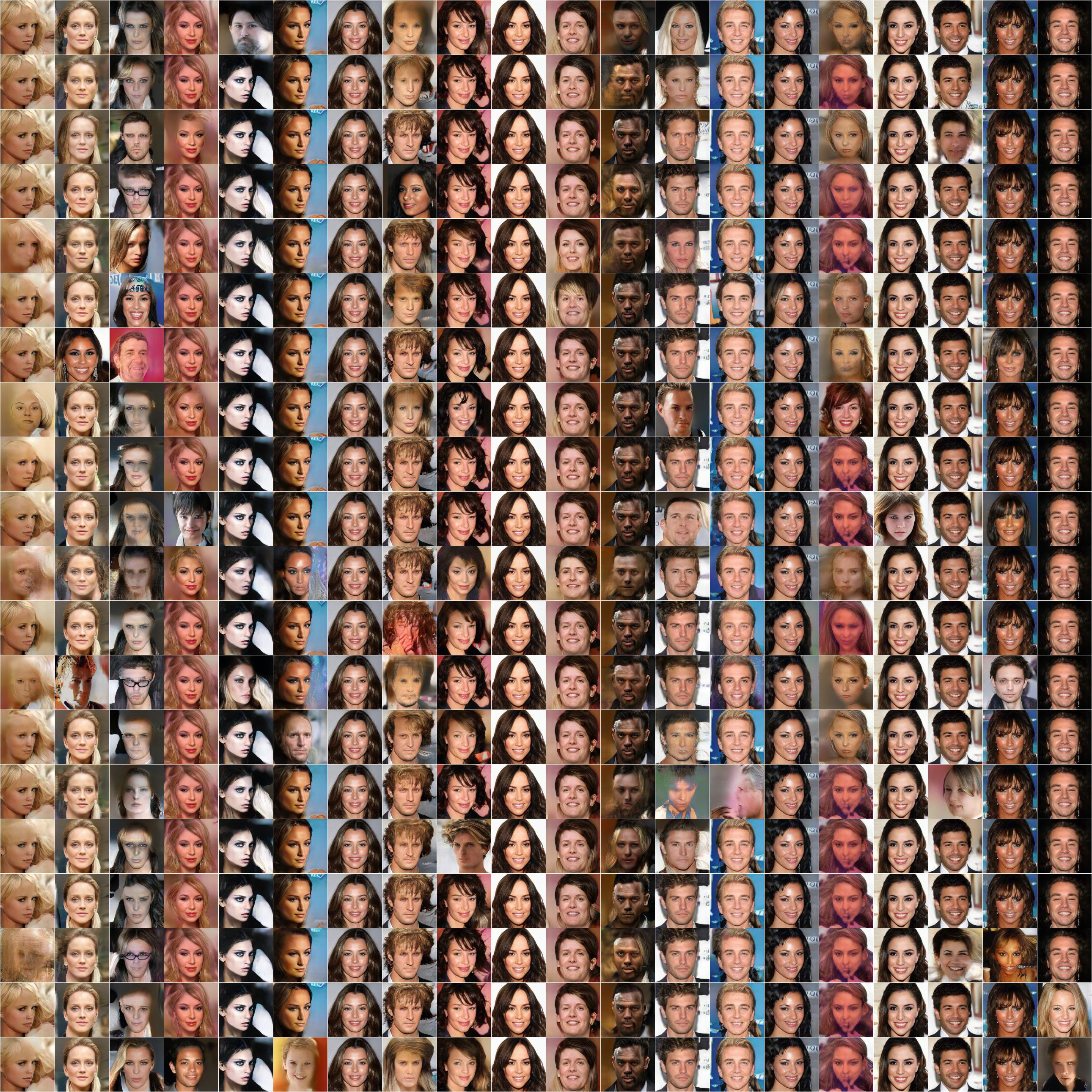}
        \\
        
        \\
        \raisebox{.2\linewidth}{\rotatebox[origin=c]{90}{Initialization (SGD)}}  
        
        \includegraphics[height=.395\linewidth,trim={0 0 0 46cm},clip]{supp/col_input_optim.jpg}
        &
        \includegraphics[width=.8\linewidth,trim={0 0 0 46cm},clip]{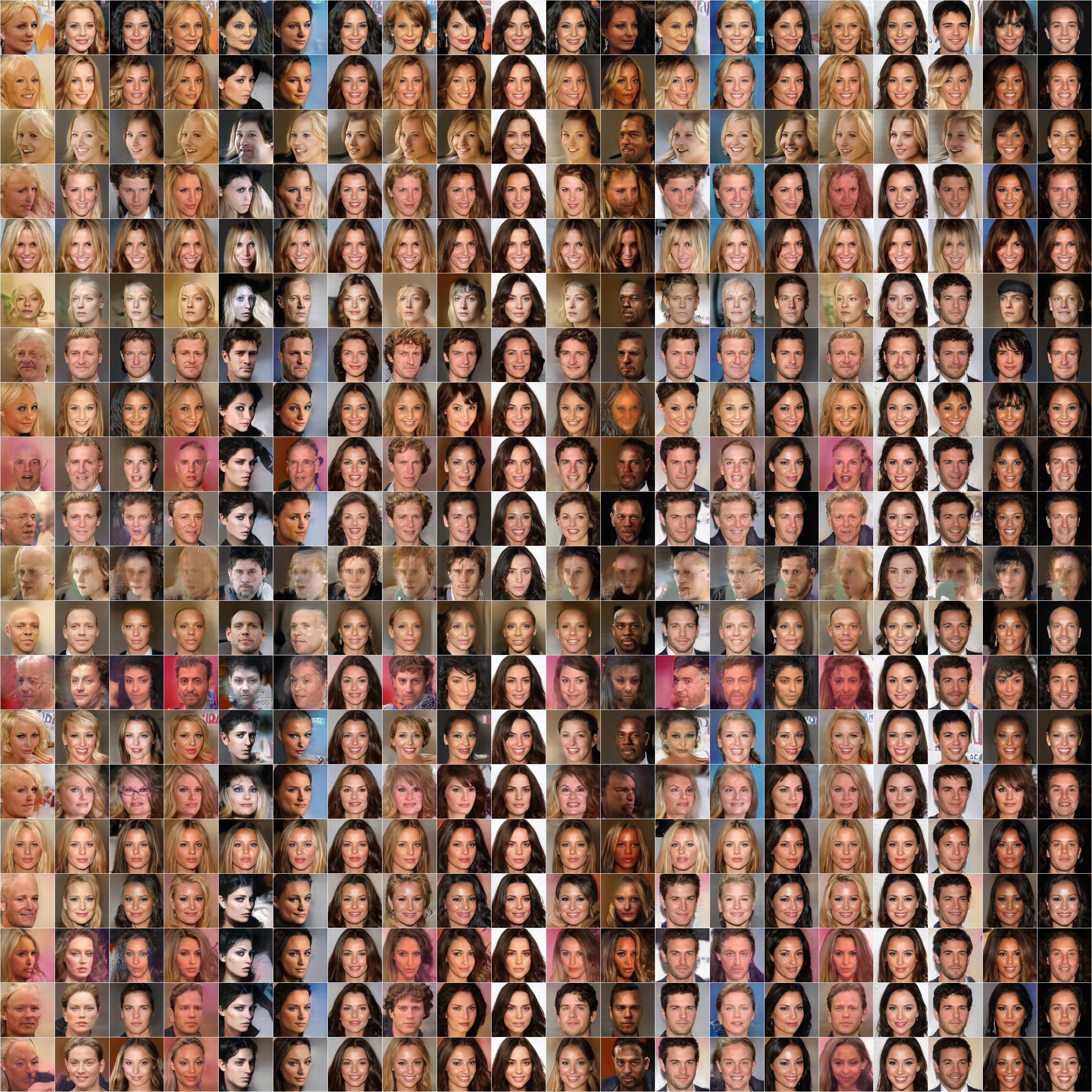}
    \end{tabular}
    
    \caption{\em Visual comparison of recovery with \textbf{LBFGS} (top) and  \textbf{SGD} for the Euclidean loss (see (NNG) optimization problem in the main paper.).
    First row: target (generated) images $y_i=G(z_i^*)$.
    First column: initialization ($G(z_i^{(0)})$).
    Second column: optimization after 100 iterations ($G(z_i^{(100)})$).
    LBFGS gives much better results than SGD that is much slower to converge, but still needs sometimes some restart (here shown without restarting).
    }
    \label{fig:LBFGS_vs_SGD_visual}
\end{figure}

\begin{figure}[p]
    \centering
    \begin{tabular}{ccc}
        \includegraphics[width=.3\linewidth]{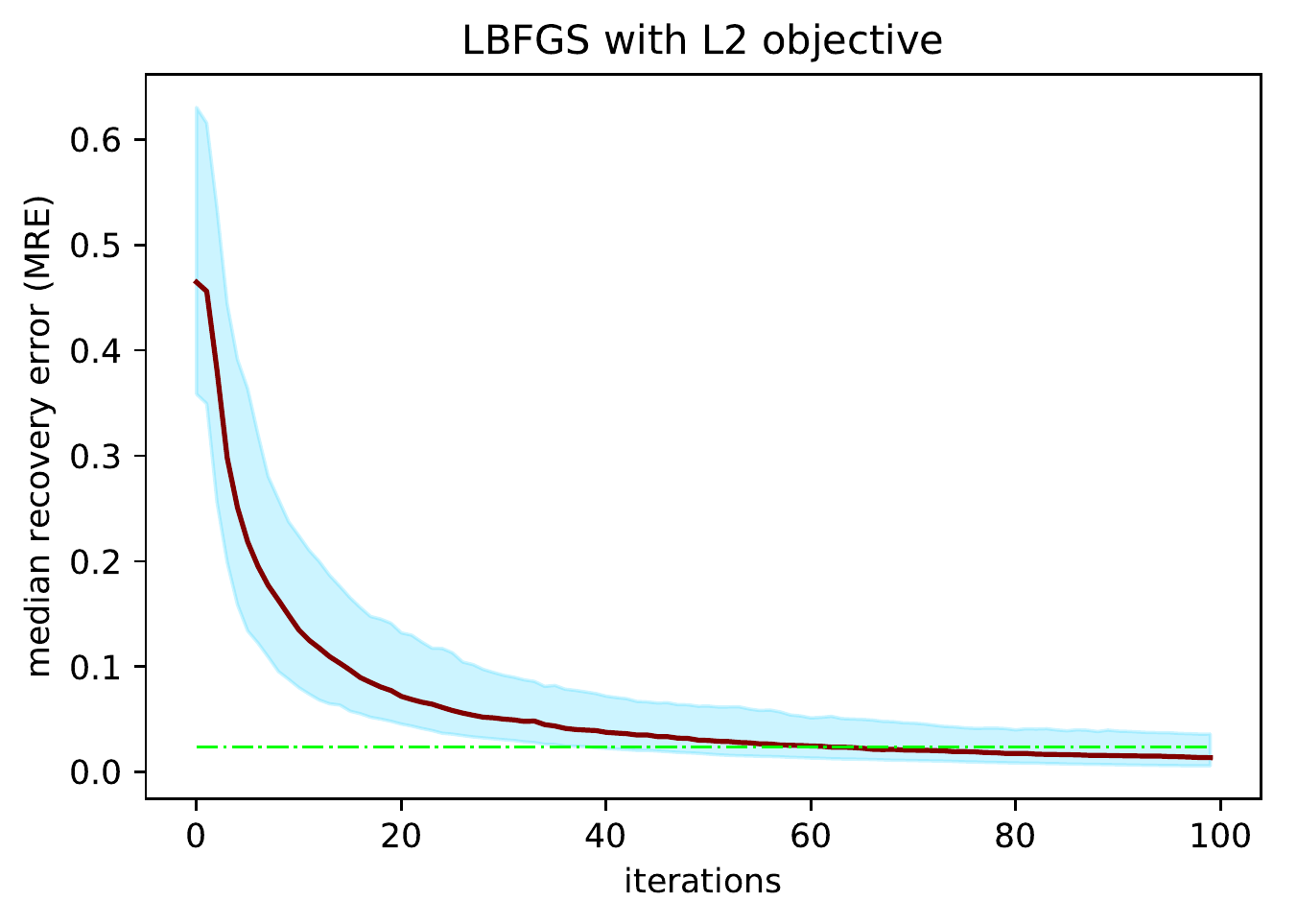} 
        & \includegraphics[width=.3\linewidth]{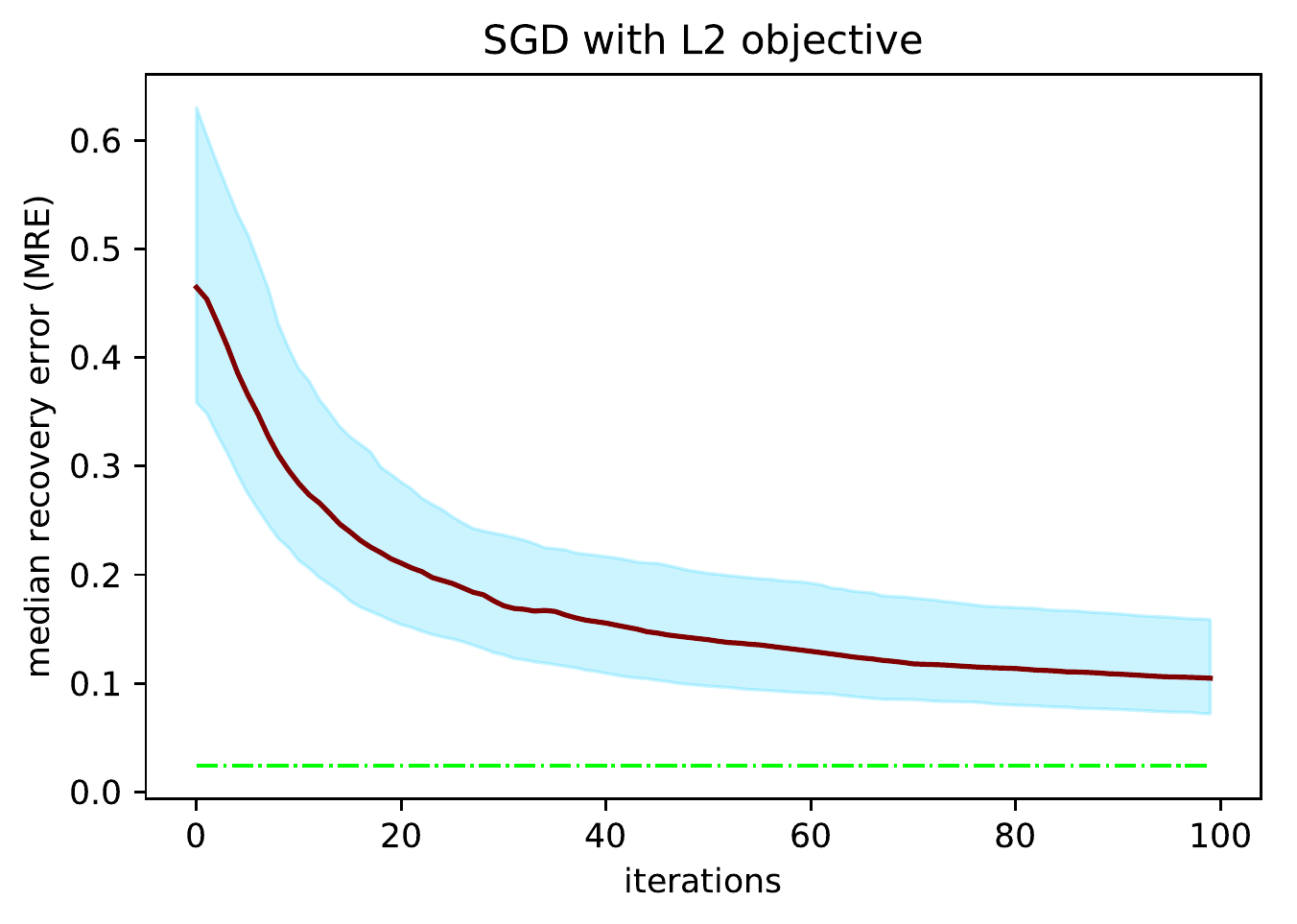}
        & \includegraphics[width=.3\linewidth,height=.21\linewidth]{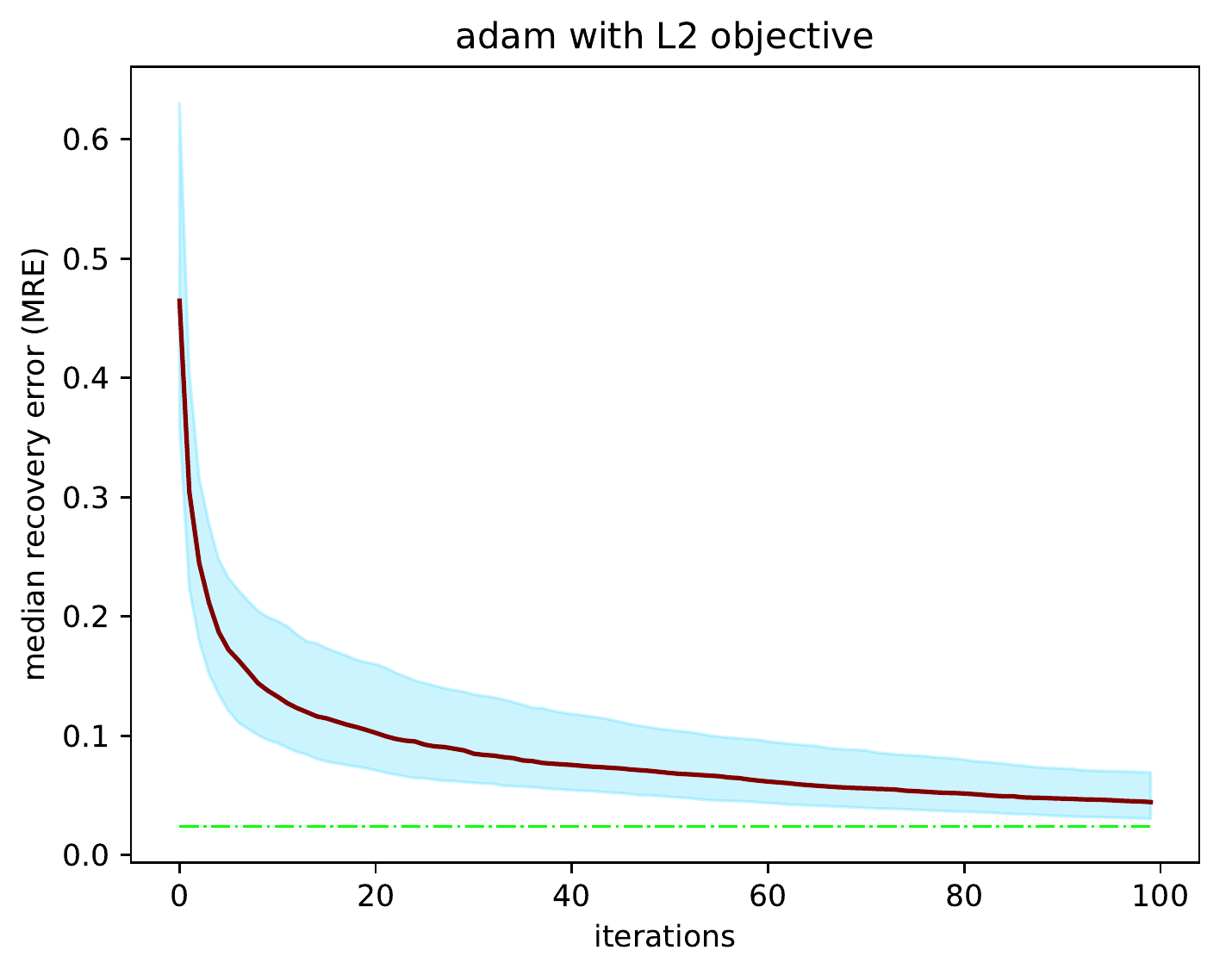}
        \\
        
        LBFGS & SGD & Adam
    \end{tabular}
    
    \caption{\em First row: median recovery error (MRE) curve.
    Second row: 400 superimposed recovery error curves for 20 images with 20 random initialization.
    LBFGS (first column) converges faster than SGD or Adam (second and third column respectively).
    }
    \label{fig:LBFGS_vs_SGD}
\end{figure}

We first show the average behavior in Fig.~\ref{fig:LBFGS_vs_SGD} the chosen optimization algorithm (LBFGS) to demonstrate that it convergences much faster than SGD and Adam.
A green dashed line shows the threshold used to detect if the actual nearest neighbor is well enough recovered ($\text{MRE} = 0.024$).
One can see that only 50 iterations are required in half the case to recover the target image.
Figure~\ref{fig:LBFGS_vs_SGD_visual} compares recovery images for PGGAN obtained with LBFGS and SGD optimization algorithms, demonstrating that LBFGS gives most of the time better results.

\subsection{Comparison of objective loss functions}

\begin{figure}[!htb]
    \centering
    \begin{tabular}{ccc}
        \includegraphics[width=.3\linewidth]{supp/median_LBFGS_with_L2_at_128.pdf} 
        & \includegraphics[width=.3\linewidth,,height=.21\linewidth]{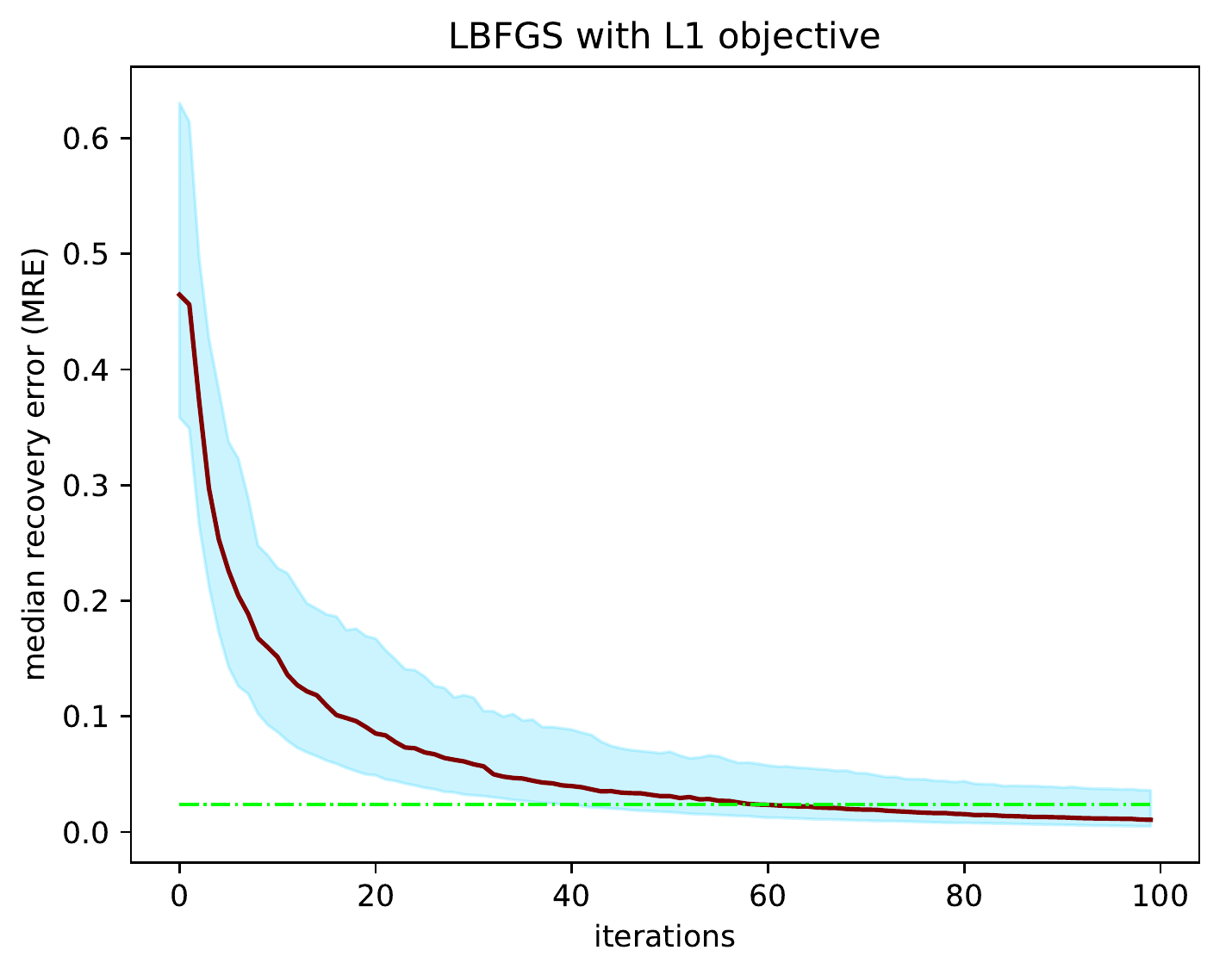}
        & \includegraphics[width=.3\linewidth]{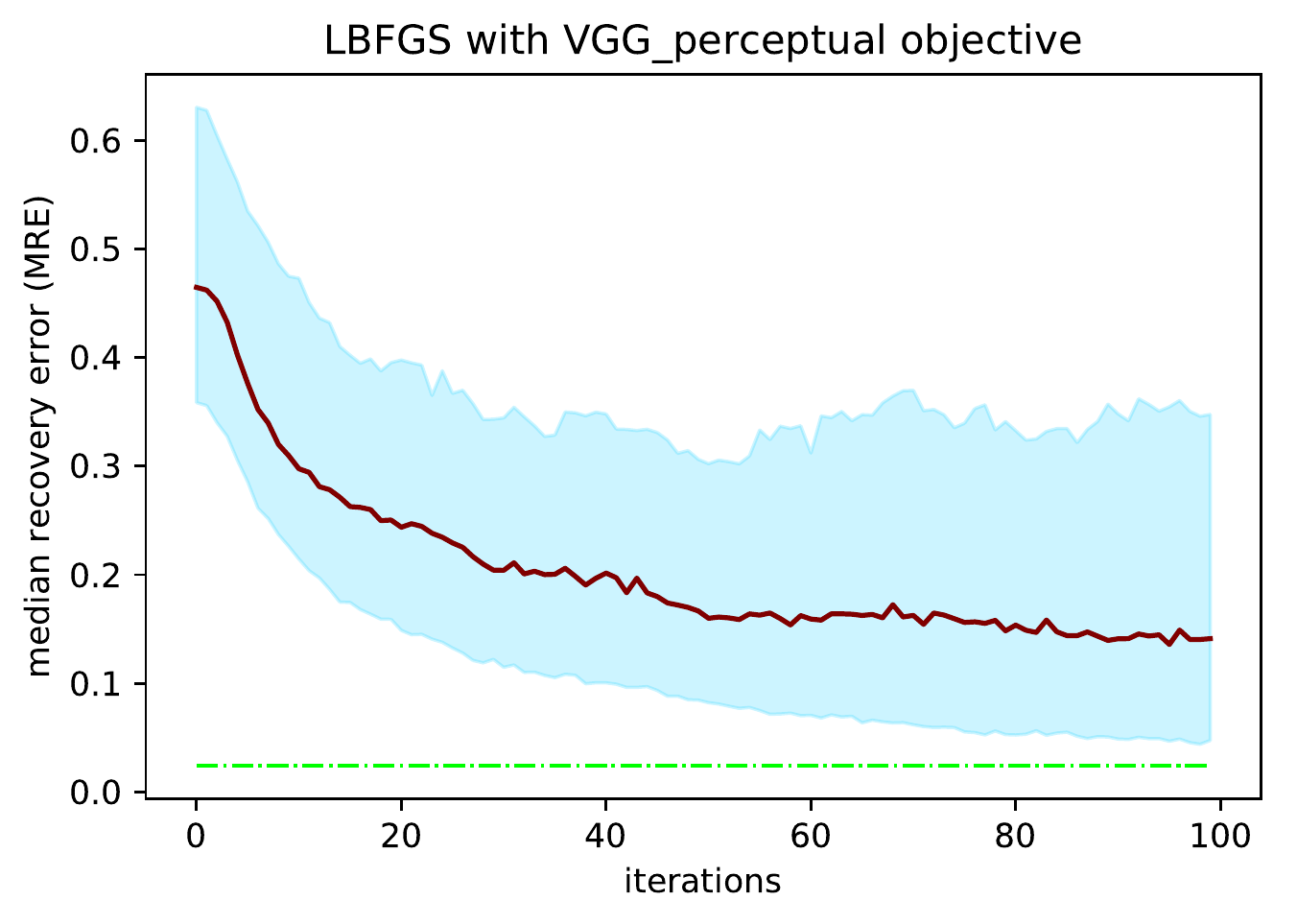}
        \\[2mm]
        
        Euclidean distance ($L_2$)
        & Manhattan distance ($L_1$) 
        & Perceptual loss (VGG-19)
    \end{tabular}
    
    \caption{LBFGS with various objectives for latent recovery for PGGAN.}
    \label{fig:Loss_comp}
\end{figure}

In Figure \ref{fig:Loss_comp} are plotted the MRE (median recovery error) when optimizing various objective functions: 
\begin{itemize}
    \item Euclidean distance ($L_2$) as used throughout the paper,
    \item Manhattan distance ($L_1$), which is often used as an alternative to the Euclidean distance that is more robust to outliers,
    \item VGG-based~\emph{perceptual loss}.
\end{itemize} 

\subsection{Convergence with operator $\phi$}

\begin{figure}[!htb]
    \centering
    \begin{tabular}{ccc}
        \includegraphics[width=.3\linewidth]{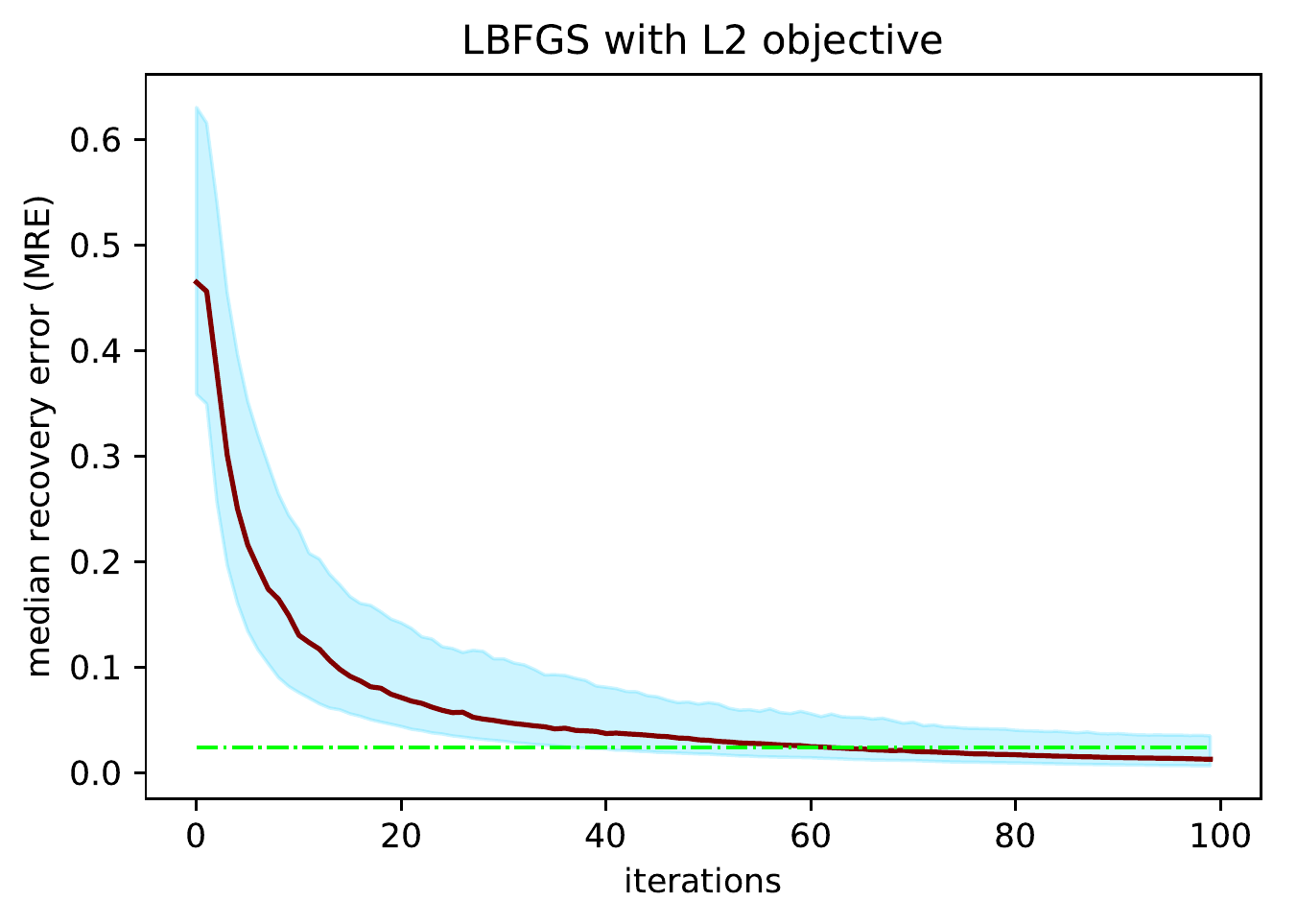} 
        & \includegraphics[width=.3\linewidth]{supp/median_LBFGS_with_L2_at_128.pdf}
        & \includegraphics[width=.3\linewidth]{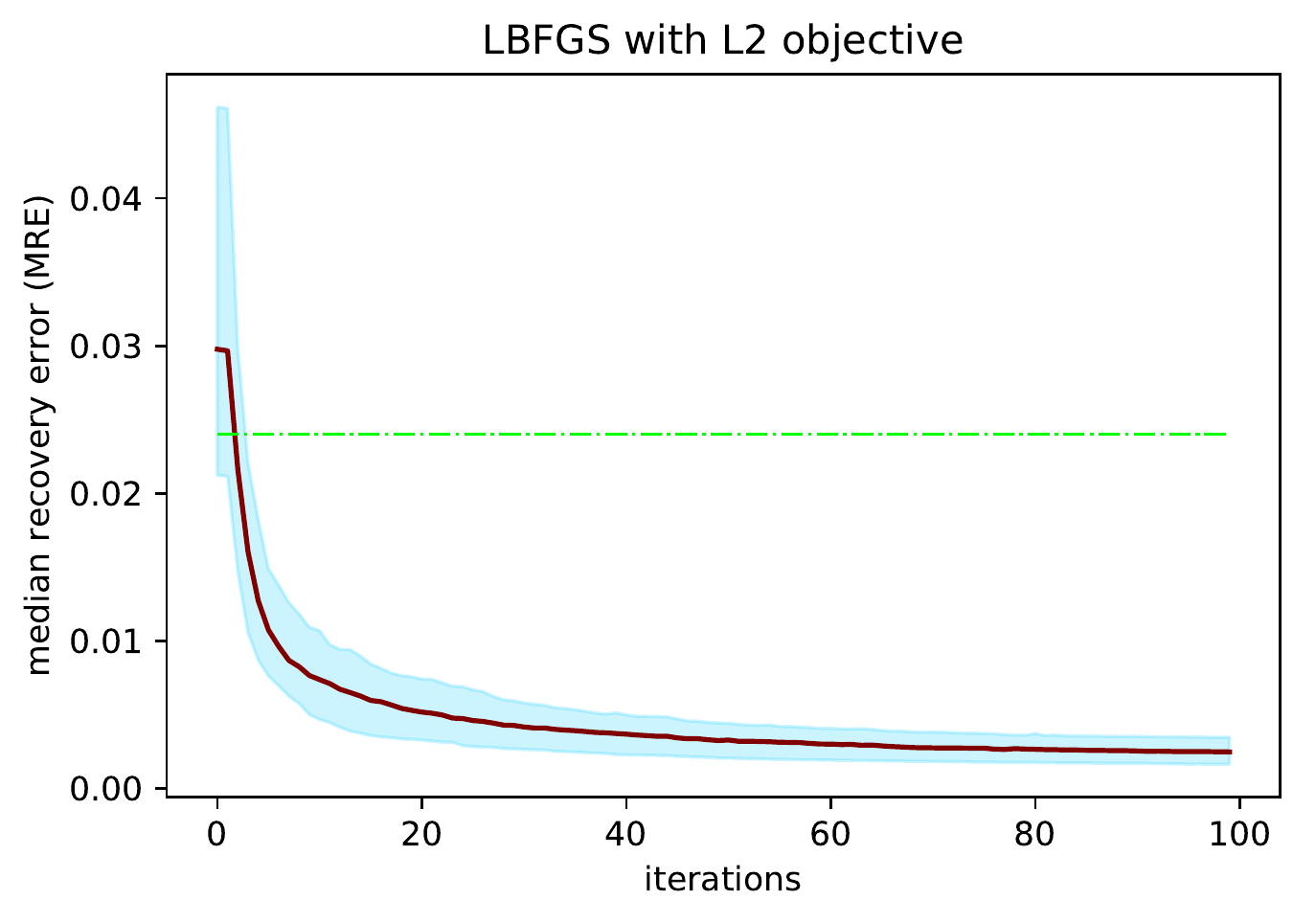}
        \\
        
        $\phi = \text{identity}$ (resolution of 1024) 
        & $\phi = \text{pooling}$ (resolution of 128) 
        & $\phi = \text{mask}$ (cropping around mouth area) 
    \end{tabular}
    
    \caption{Using various $\phi$ operator for applications (super-resolution, inpainting) has no effect on convergence.}
    \label{fig:pool}
\end{figure}

Figure \ref{fig:pool} demonstrates convergence under various operators $\phi$.

\subsection{Recovery with other generators}

\begin{figure}[!htb]
    \centering
    \begin{tabular}{cccc}
        
        \includegraphics[width=.2\linewidth]{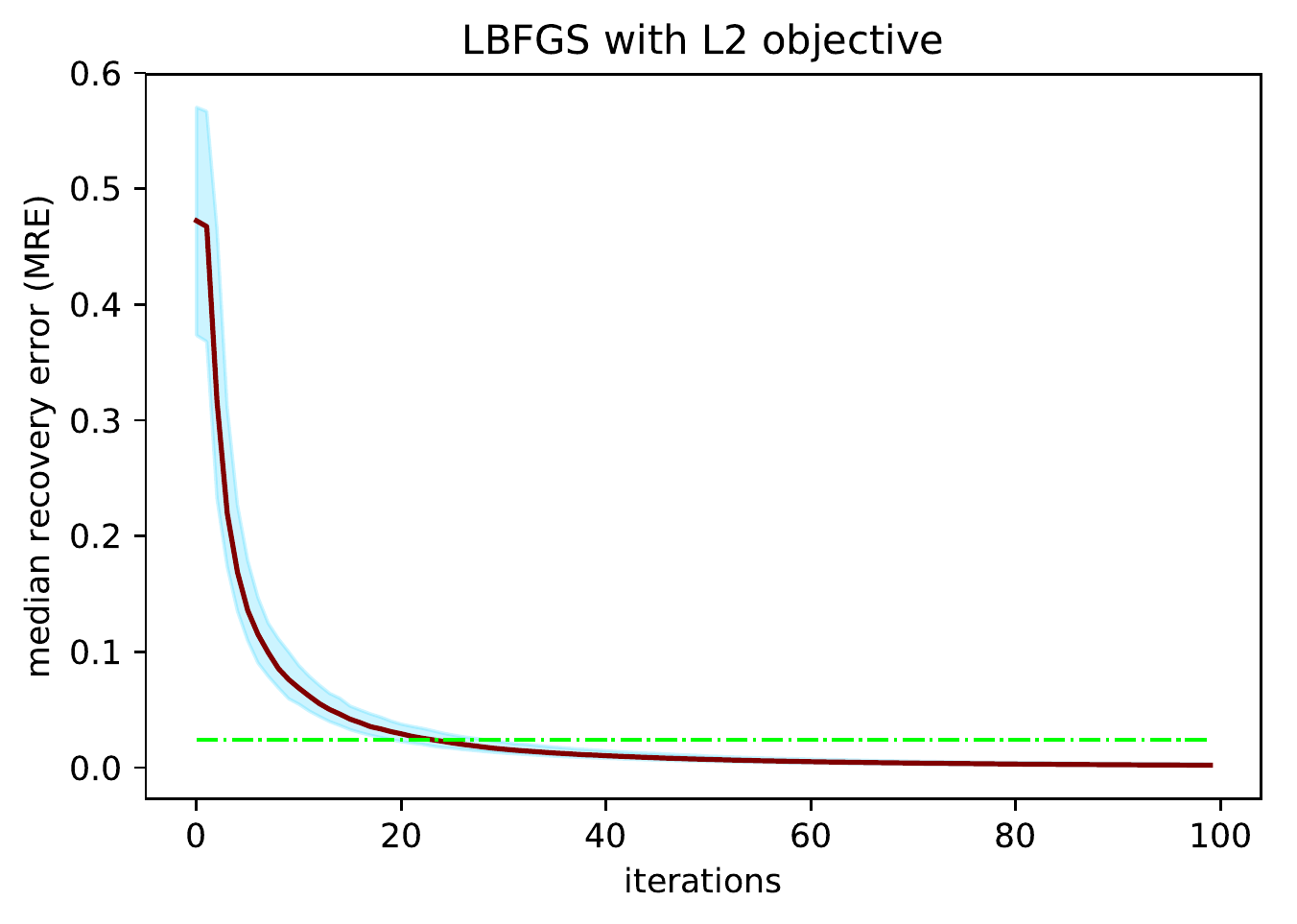}
        & \includegraphics[width=.2\linewidth]{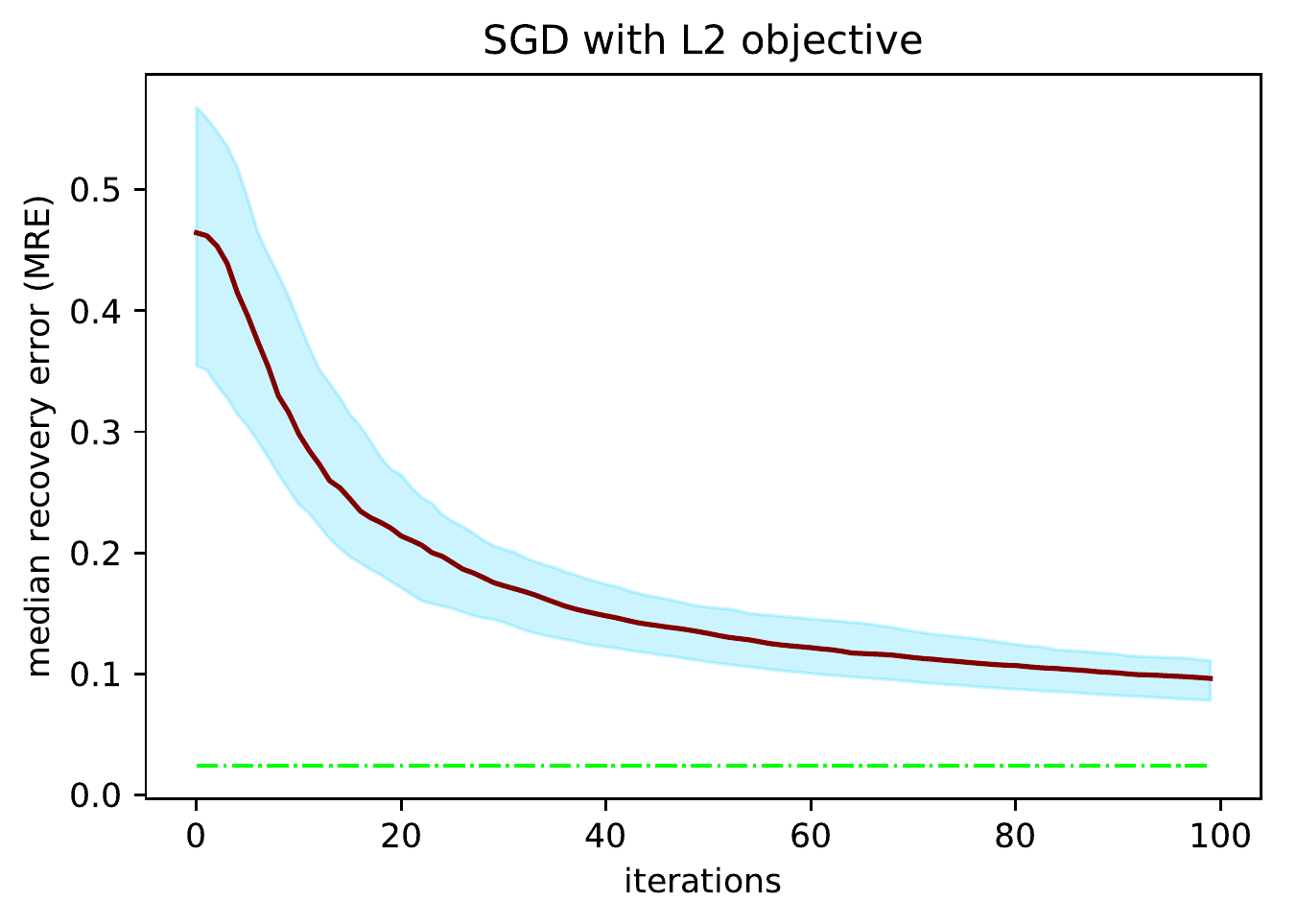}
        & \includegraphics[width=.2\linewidth]{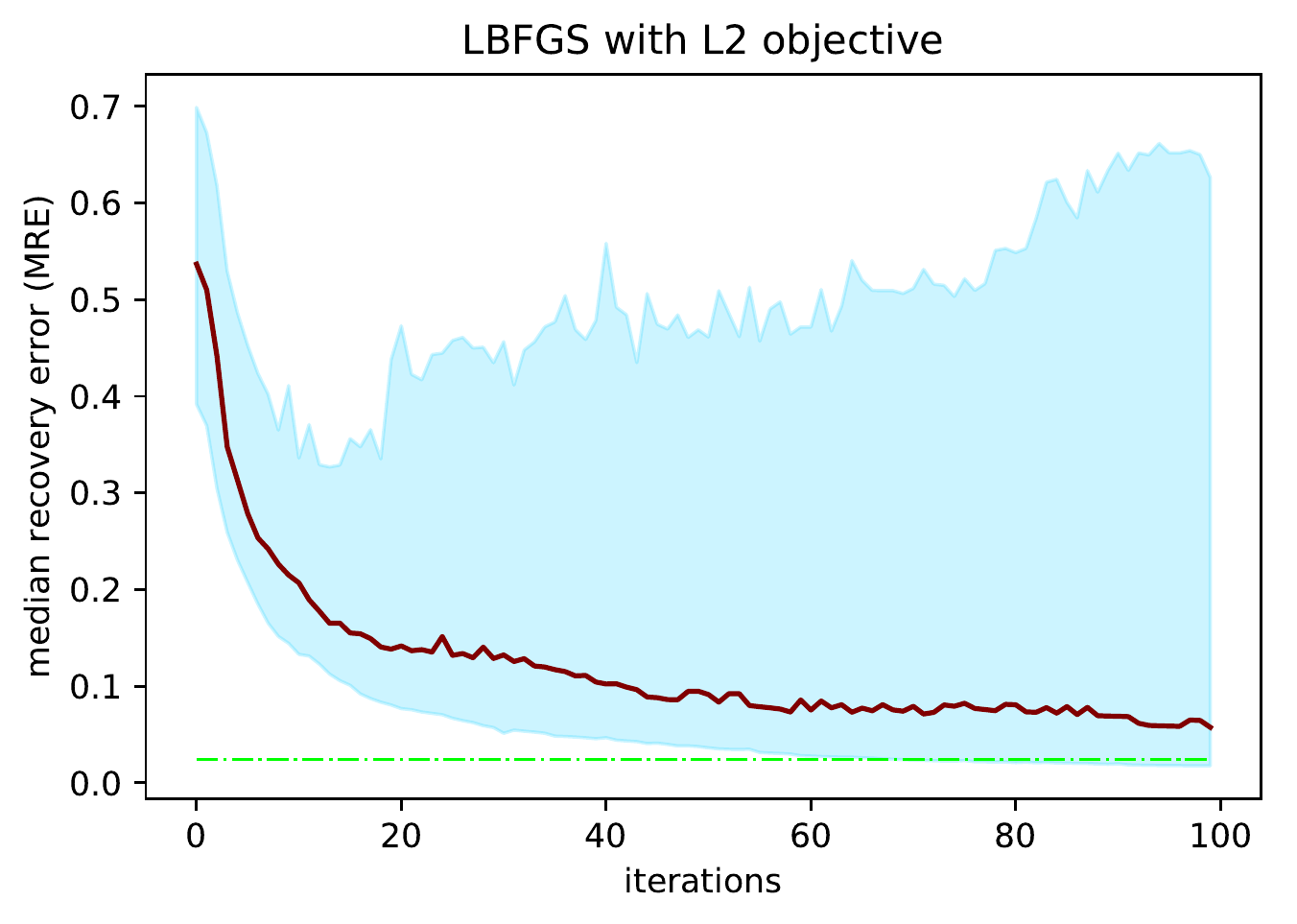}
        & \includegraphics[width=.2\linewidth]{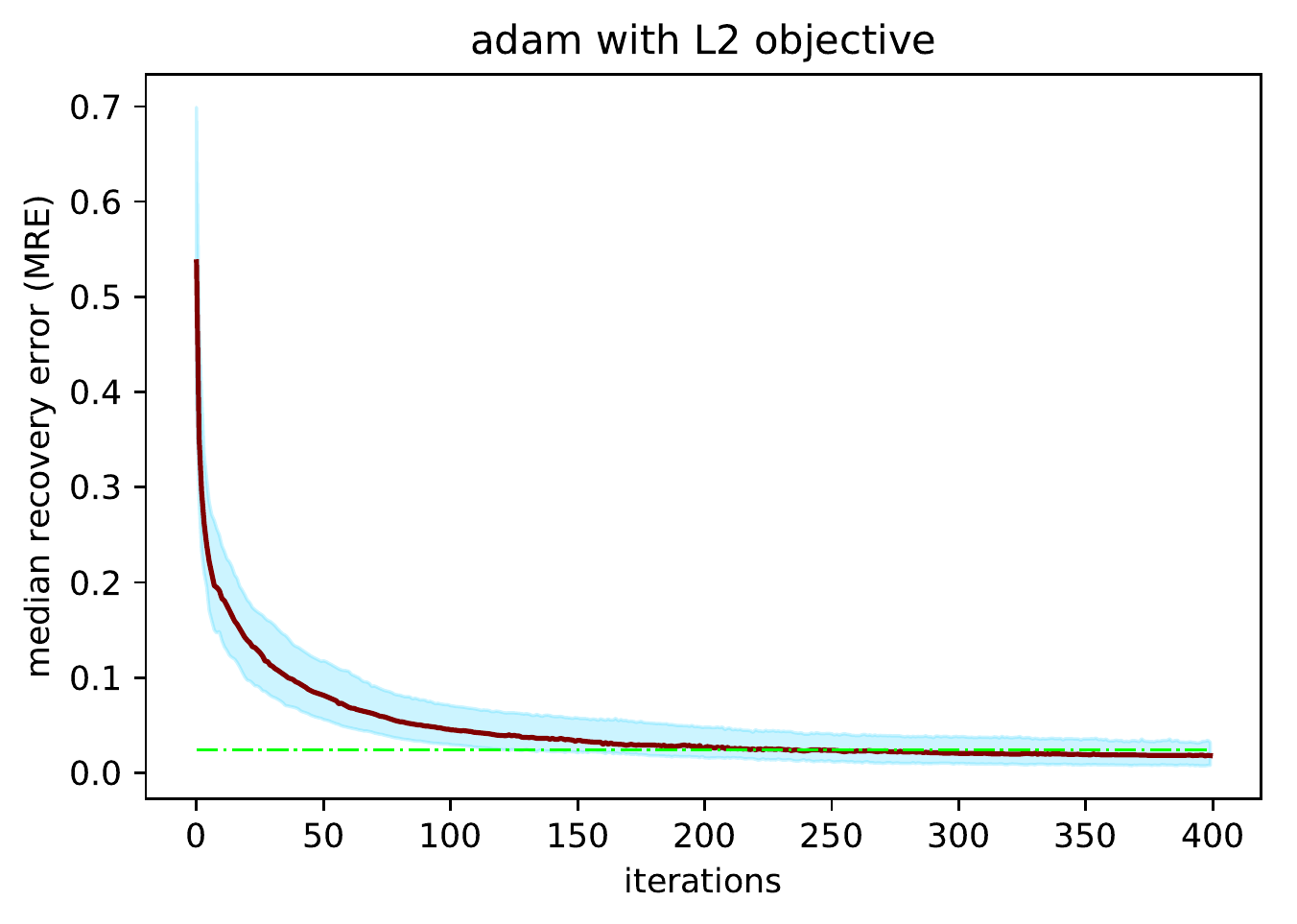}
        \\
        
        DCGAN with LBFGS 
        & DCGAN with SGD
        & MESCH with LBFGS 
        & MESCH with Adam (400 iterations)
    \end{tabular}
    
    \caption{Comparing LBFGS with SGD and Adam algorithms for generated image recovery with a DCGAN and a MESCH network}
    \label{fig:DCGAN_MESCH}
\end{figure}

\begin{figure}[p]
    \centering
    \begin{tabular}{cc}
        & Target images (generated with DCGAN)
        \\
        & \includegraphics[width=.89\linewidth]{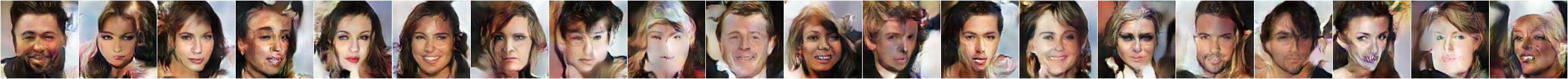}
        \\
        \raisebox{.2\linewidth}{\rotatebox[origin=c]{90}{Initialization}}  
        
        \includegraphics[height=.4\linewidth,trim={0 50cm 0 0},clip]{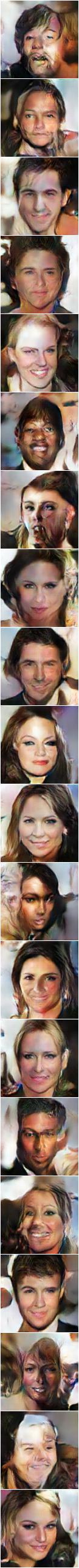}
        &
        \includegraphics[height=.4\linewidth,trim={0 50cm 0 0},clip]{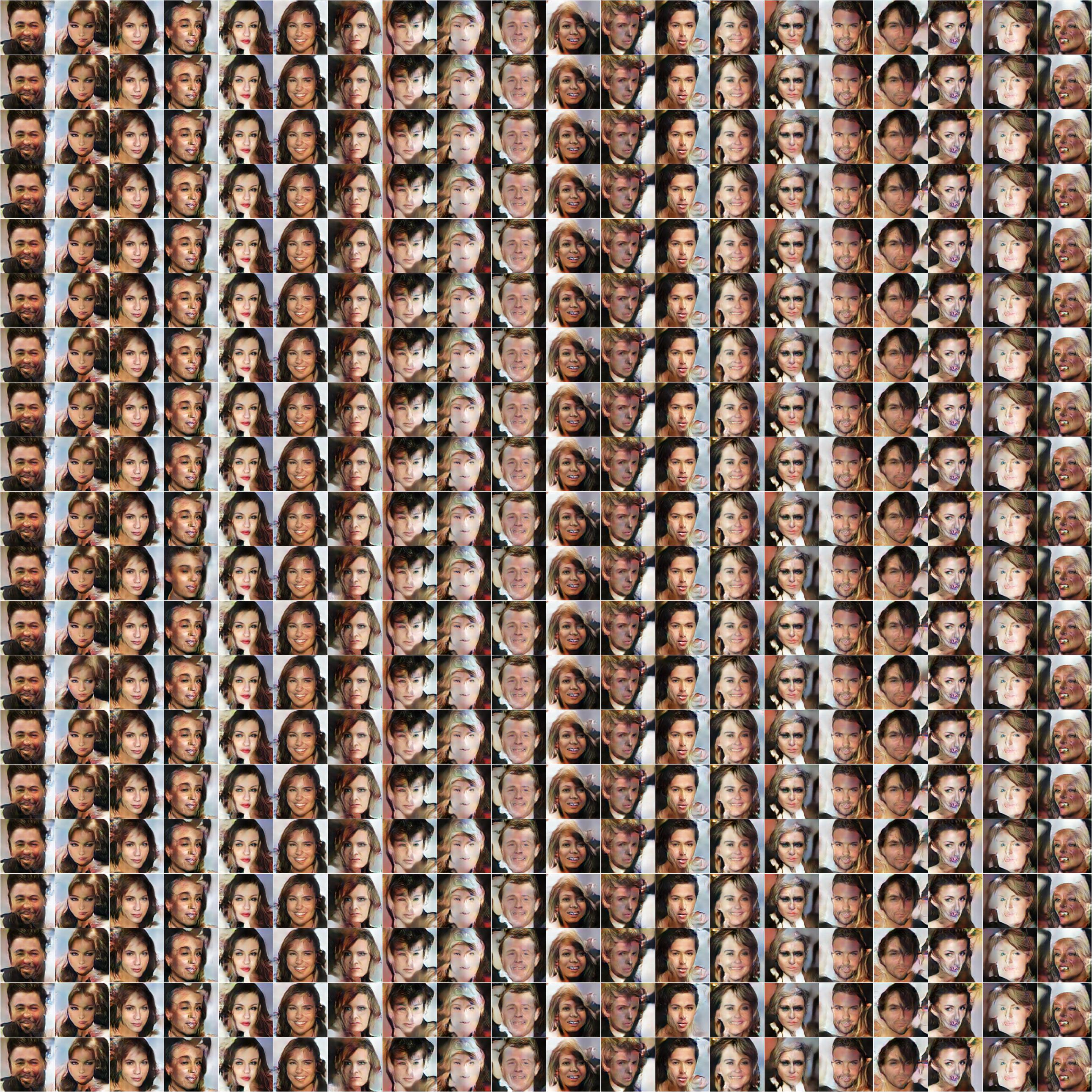}
    \end{tabular}
    
    \caption{\em Visual results on recovery of generated images of a \textbf{DCGAN} network.
    First row: target (generated) images $y_i=G(z_i^*)$.
    First column: initialization ($G(z_i^{(0)})$).
    Second column: optimization  with {LBFGS} and Euclidean loss after 100 iterations ($G(z_i^{(100)})$).
    }
    \label{fig:DCGAN_visual}
\end{figure}

\begin{figure}[p]
    \centering
    \begin{tabular}{cc}
        & Target images (generated with MESCH)
        \\
        & \includegraphics[width=.8\linewidth]{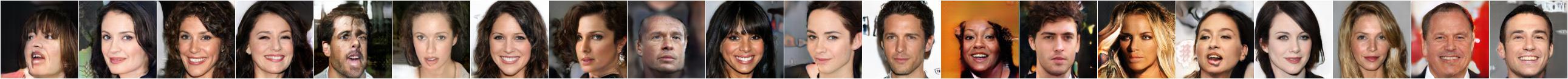}
        \\
        \raisebox{.2\linewidth}{\rotatebox[origin=c]{90}{Initialization}}  
        
        \includegraphics[height=.4\linewidth]{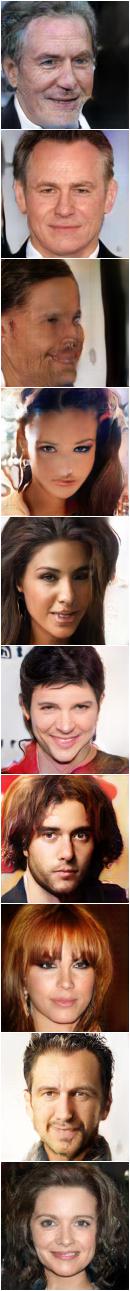}
        &
        \includegraphics[height=.4\linewidth]{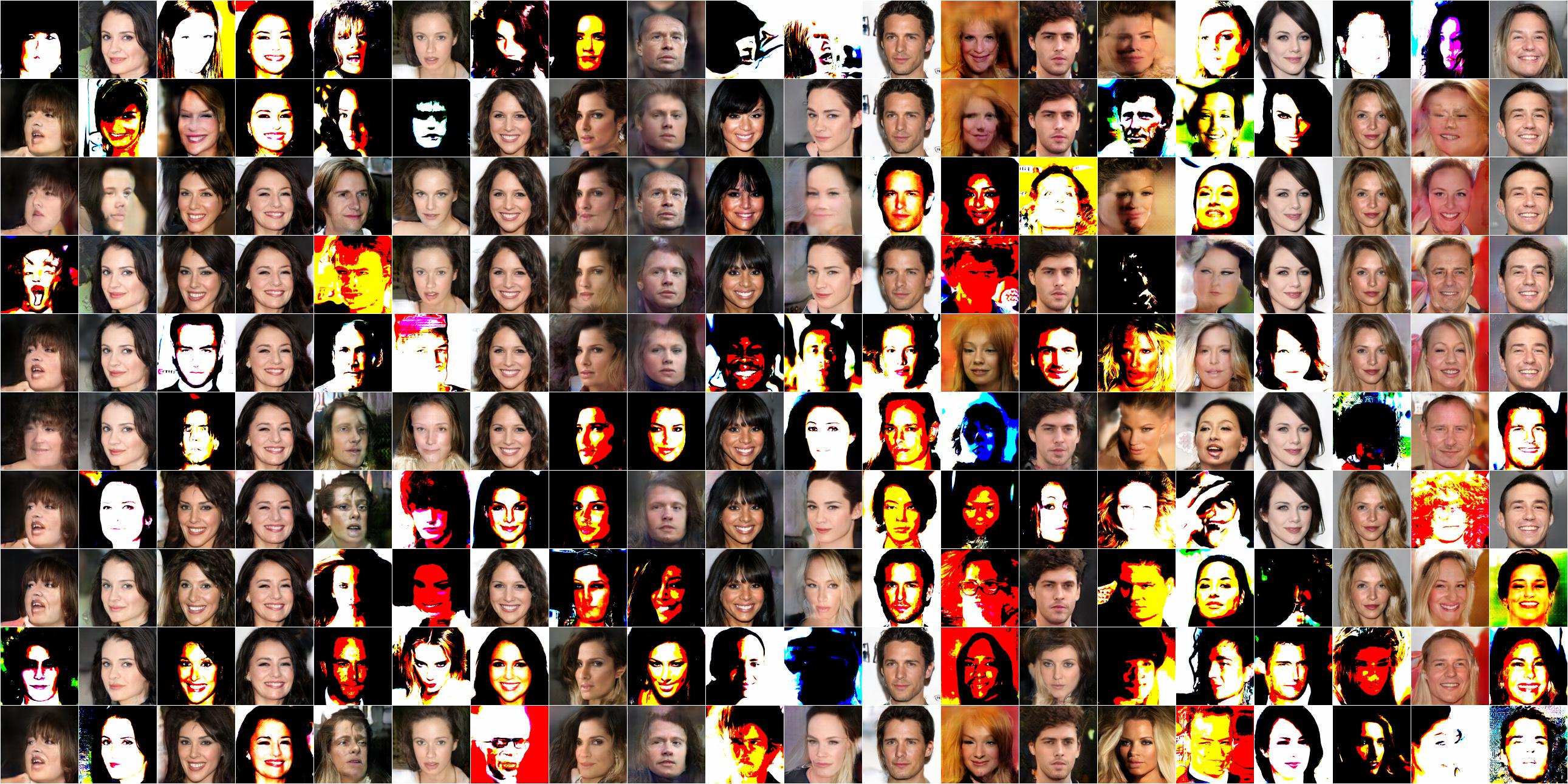}
    \end{tabular}
    
    \caption{\em Visual results on recovery of generated images of a \textbf{MESCH} network.
    First row: target (generated) images $y_i=G(z_i^*)$.
    First column: initialization ($G(z_i^{(0)})$).
    Second column: optimization with {LBFGS} and Euclidean loss after 100 iterations ($G(z_i^{(100)})$). 
    As previously reported in Section 4, success rate of optimization for MESCH generator is only around 67\%, but it can be circumvented easily by restarting with new random initialization or by using Adam optimization.
    }
    \label{fig:MESCH_visual}
\end{figure}

Figure~\ref{fig:DCGAN_MESCH} displays median recovery error (MRE) when optimizing with LBFGS and SGD for DCGAN and MESCH generators.
Visual results are given for LBFGS in Figures~\ref{fig:DCGAN_visual} and~\ref{fig:MESCH_visual}.
The MESCH network is more inconsistent, but using 10 random initialization is enough to ensure the recovery of a generated (or real) image with 96\% chance.

\subsection{Convergence on real images}

\begin{figure}[htb]
    \centering
    \includegraphics[width=.45\linewidth]{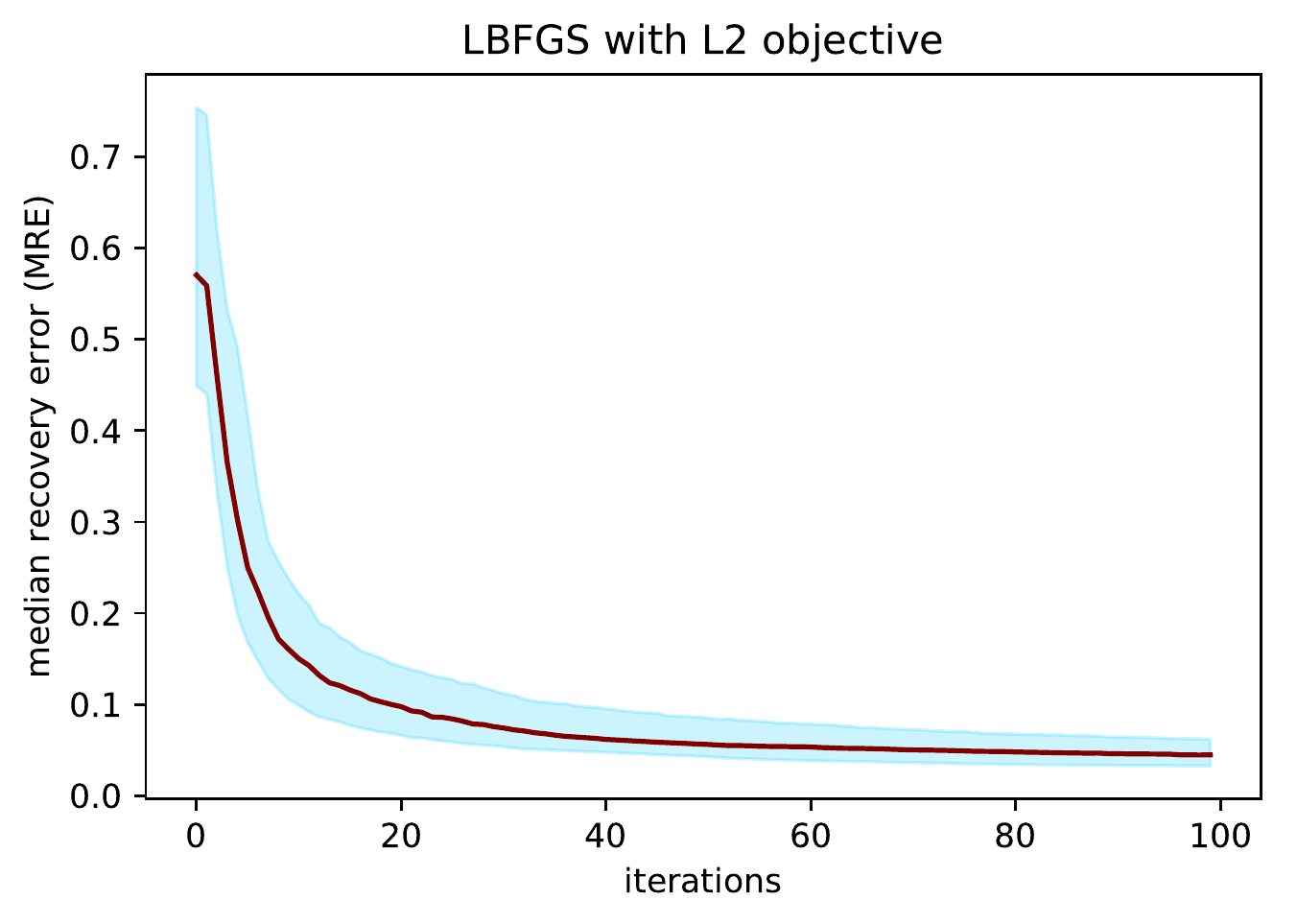}
        
    \centering
    \begin{tabular}{cc}
        & Target images (from the training set)
        \\
        & \includegraphics[width=.8\linewidth]{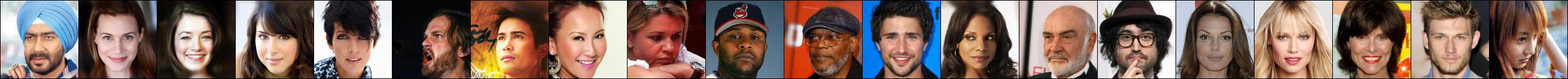}
        \\
        \raisebox{.4\linewidth}{\rotatebox[origin=c]{90}{Initialization}}  
        
        \includegraphics[height=.8\linewidth]{supp/col_input_optim.jpg}
        &
        \includegraphics[height=.8\linewidth]{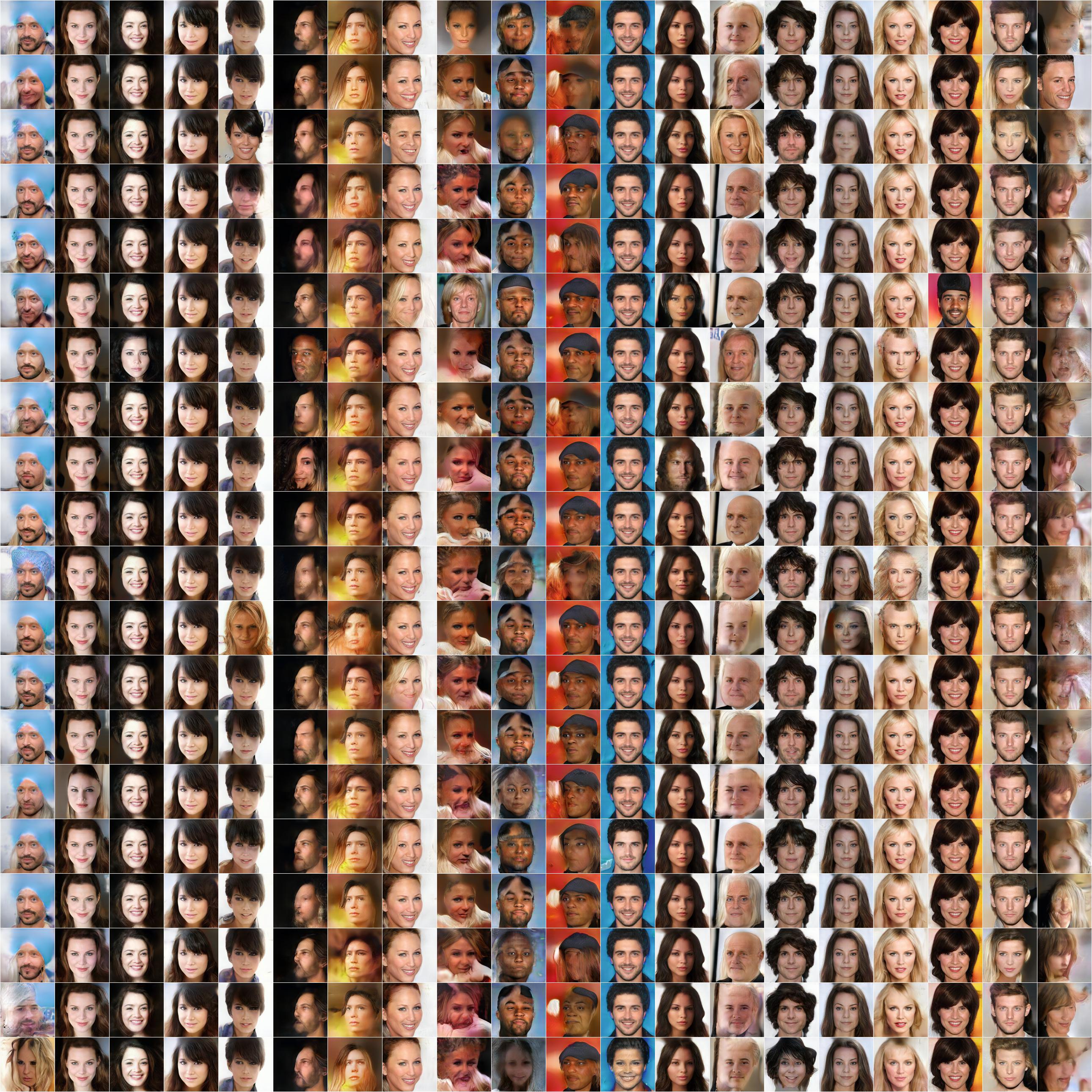}
    \end{tabular}
    
    \caption{\em \textbf{Top:} average convergence curve of LBFGS for recovery of training images. 
    \textbf{Bottom:} Visual results on \textbf{real} images recovery (training set from celeba-HQ) with {LBFGS} and Euclidean objective loss, and PGGAN generator.
    First row: target (real) images $y_i$.
    First column: initialization ($G(z_i^{(0)})$).
    Second column: optimization after 100 iterations ($G(z_i^{(100)})$).
    }
    \label{fig:LBFGS_visual_real}
\end{figure}

Figure~\ref{fig:LBFGS_visual_real} shows highly consistent recovery on real images for the PGGAN network.

\end{document}